\documentclass{article}

\usepackage[preprint]{neurips_2025}

\usepackage[utf8]{inputenc} 
\usepackage[T1]{fontenc}    
\usepackage{hyperref}       
\usepackage{url}            
\usepackage{booktabs}       
\usepackage{amsfonts}       
\usepackage{nicefrac}       
\usepackage{microtype}      
\usepackage{xcolor}         
\usepackage{comment}        
\usepackage{amsthm}
\newtheorem{proposition}{Proposition}
\usepackage{amsmath}
\usepackage{amssymb}

\newtheorem{theorem}{Theorem}

\usepackage{enumitem}
\usepackage[most]{tcolorbox}
\usepackage{algorithm}
\usepackage{algpseudocode}
\usepackage{caption}  
\usepackage{cleveref}
\usepackage{csquotes}
\usepackage{tikz}
\usepackage{tabularx}

\usepackage{graphicx}
\usepackage{wrapfig}

\title{Online Feedback Efficient Active Target Discovery \\
           in Partially Observable Environments}

\author{
    Anindya Sarkar\thanks{Equal contribution} ,~~~~~~Binglin Ji\footnotemark[1] ,~~~~~~Yevgeniy Vorobeychik \\
    \texttt{\{anindya,~binglin.j,~yvorobeychik\}@wustl.edu,}\\Department of Computer Science and Engineering\\ Washington University in St. Louis, USA\\}

\begin{document}

\maketitle

\begin{abstract}
In various scientific and engineering domains, where data acquisition is costly—such as in medical imaging, environmental monitoring, or remote sensing—strategic sampling from unobserved regions, guided by prior observations, is essential to maximize target discovery within a limited sampling budget.
In this work, we introduce Diffusion-guided Active Target Discovery (\emph{DiffATD}), a novel method that leverages diffusion dynamics for active target discovery. \emph{DiffATD} maintains a belief distribution over each unobserved state in the environment, using this distribution to dynamically balance exploration-exploitation. Exploration reduces uncertainty by sampling regions with the highest expected entropy, while exploitation targets areas with the highest likelihood of discovering the target, indicated by the belief distribution and an incrementally trained reward model designed to learn the characteristics of the target. 
\emph{DiffATD} enables efficient target discovery in a partially observable environment within a fixed sampling budget, all without relying on any prior supervised training. Furthermore,  \emph{DiffATD} offers interpretability, unlike existing black-box policies that require extensive supervised training. 
Through extensive experiments and ablation studies across diverse domains, including medical imaging, species discovery and remote sensing, we show that \emph{DiffATD} performs significantly better than baselines and competitively with supervised methods that operate under full environmental observability.
\end{abstract}

\section{Introduction}\label{sec:int}
A key challenge in many scientific discovery applications is the high cost associated with sampling and acquiring feedback from the ground-truth reward function, which often requires extensive resources, time, and cost. 
For instance, in MRI, doctors aim to minimize radiation exposure by strategically selecting brain regions to identify potential tumors or other conditions. However, obtaining detailed scans for accurate diagnosis is resource-intensive and time-consuming, involving advanced equipment, skilled technicians, and substantial patient time. Furthermore, gathering reliable feedback to guide treatment decisions often requires expert judgment, adding another layer of complexity and cost.
Similarly, in a search and rescue mission, personnel must strategically decide where to sample next based on prior observations to locate a target of interest (e.g., a missing person), taking into account limitations such as restricted field of view and high acquisition costs. In this context, obtaining feedback from the ground truth reward function could involve sending a team for on-site verification or having a human analyst review the collected imagery, both of which require significant resources and expert judgment.
This challenge extends to various scientific and engineering fields, such as material and drug discovery, where the objective may vary but the core problem—optimizing sampling to identify a target under constraints—remains the same. The challenge is further exacerbated by the fact that obtaining labeled data for certain rare target categories, such as rare tumors, is often not feasible. Naturally, the question emerges: \emph{\textbf{Is it possible to design an algorithm capable of identifying the target of interest while minimizing reliance on expensive ground truth feedback, and without requiring any task-specific supervised training?}}

The challenge is twofold: (i) the agent must judiciously sample the most informative region from a partially observed scene to maximize its understanding of the search space, and (ii) it must concurrently ensure that this selection contributes to the goal of identifying regions likely to include the target. One might wonder why the agent cannot simply learn to choose regions that directly unveil target regions, bypassing the need to acquire knowledge about the underlying environment. The crux of the issue lies in the inherent complexity of reasoning within an unknown, partially observable domain. Consequently, the agent must deftly balance exploration—identifying regions that yield the greatest informational gain about the search space—with exploitation—focusing on areas more likely to contain the target based on the evolving understanding of the environment. 
Prior approaches typically rely on off-the-shelf reinforcement learning algorithms to develop a search policy that can efficiently explore a search space and identify target regions within a partially observable environment by training on large-scale pre-labeled datasets from similar tasks. However, obtaining such supervised datasets is often impractical in practice, restricting the models' applicability in real-world scenarios. Suppose the task is identifying a rare tumor in an MRI image. In that case, it is unrealistic to expect large-scale supervised data for such a rare condition, with positive labels indicating the target.

To address this challenge, we introduce \emph{DiffATD}, an approach capable of uncovering a target in a partially observable environment without the need for training on pre-labeled datasets, setting it apart from previous methods. Conditioned on the observations gathered so far, \emph{DiffATD} leverages Tweedie’s formula to construct a belief distribution over each unobserved region of the search space. 
\begin{figure}
    \centering
    \includegraphics[width=0.98\textwidth]{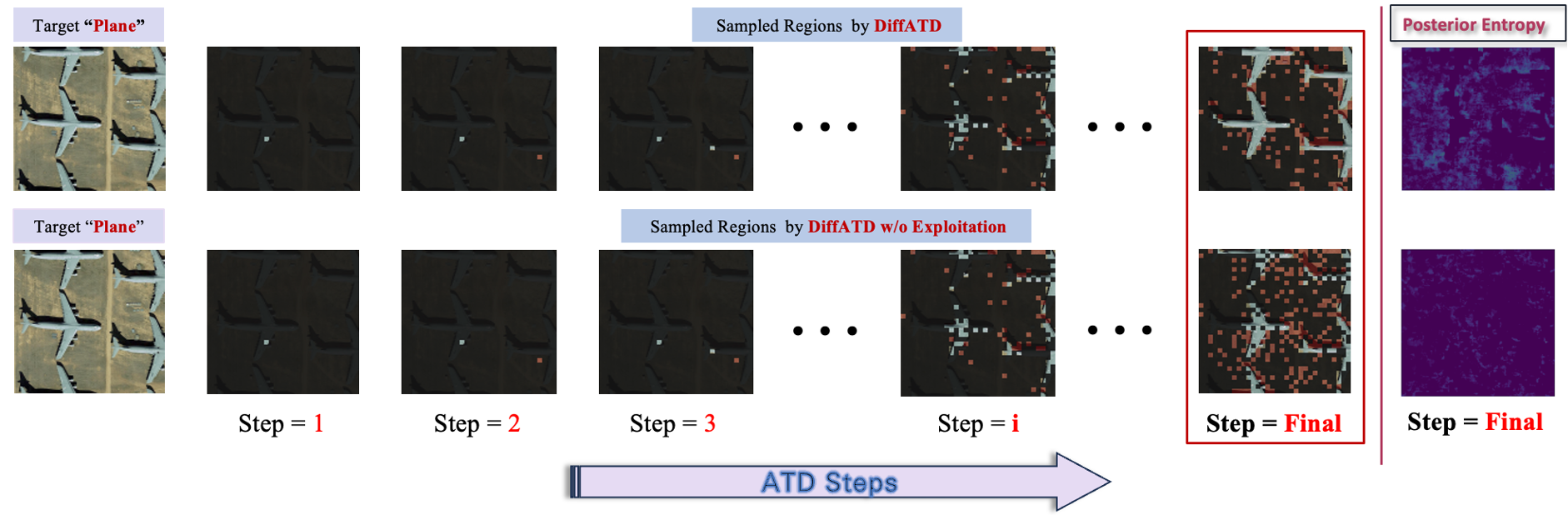}
    \vspace{-6pt}
    \caption{\emph{DiffATD} uncovers more target pixels (e.g., plane) while exhibiting higher uncertainty in the search space. In contrast, \emph{DiffATD without exploitation} explores the environment more effectively, resulting in lower entropy but identifies fewer target pixels.}
    \label{fig:intro}
    \vspace{-16pt}
\end{figure}
Exploration is guided by selecting the region with the highest expected entropy of the corresponding belief distribution, while exploitation directs attention to the region with the highest likelihood score measured by the lowest expected entropy, modulated by the reward value, predicted by an online-trained reward model designed to predict the \enquote{target-ness} of a region. Figure 1 illustrates how exploitation enables \emph{DiffATD} to uncover more target pixels (e.g., plane) while exhibiting higher uncertainty about the search space. Moreover, \emph{DiffATD} offers interpretability, providing enhanced transparency compared to existing methods that rely on opaque black-box policies, all while being firmly grounded in the theoretical framework of Bayesian experiment design. We summarize our contributions as follows:
\begin{itemize}[noitemsep,topsep=0pt, leftmargin=*]
\begin{tcolorbox}[colback=blue!5!white, colframe=blue!40!black, boxrule=0.5pt, arc=2mm]
    \item We propose \emph{DiffATD}, a novel algorithm that uncovers the target of interest within a feedback/query budget in a partially observable environment, minimizing reliance on costly ground truth feedback and avoiding the need for task-specific training with annotated data. 
    \item \emph{DiffATD} features a white-box policy for sample selection based on Bayesian experiment design, offering greater interpretability and transparency compared to existing black-box methods.
    \item We demonstrate the significance of each component in our proposed \emph{DiffATD} method through extensive quantitative and qualitative ablation studies across a range of datasets, including medical imaging, species distribution modeling, and remote sensing.
    \end{tcolorbox}
\end{itemize}



\section{Related Work}
\noindent
\textbf{Active Target/Scene Reconstruction.} There is extensive prior work on active reconstruction of scenes and/or objects~\citep{jayaraman2016look, jayaraman2018learning, xiong2018snap}. Recently, deep learning methods have been developed to design subsampling strategies that aim to minimize reconstruction error. Fixed strategies, such as those proposed by~\citep{huijben2020deep, bahadir2020deep}, involve designing a single sampling mask for a specific domain, which is then applied uniformly to all samples during inference. While effective in certain scenarios, these methods face limitations when the optimal sampling pattern varies across individual samples. To overcome this, sample-adaptive approaches~\citep{van2021active,bakker2020experimental,yin2021end,stevens2022accelerated} have been introduced, dynamically tailoring sampling strategies to each sample during inference. 
~\cite{van2021active} trained a neural network to iteratively construct an acquisition mask of M k-space lines, adding lines adaptively based on the current reconstruction and prior context. Similarly, \cite{bakker2020experimental} used an RL agent for this process. However, these methods rely on black-box policies, making failure cases hard to interpret. Generative approaches, like~\citep{sanchez2020closed,nolan2024active}, address this with transparent sampling policies, such as maximum-variance sampling for measurement selection.
However, all these prior methods typically focus solely on optimizing for reconstruction, while our ultimate goal is identifying target-rich regions. Success for our task hinges on balancing \emph{exploration} (obtaining useful information about the scene) and \emph{exploitation} (uncovering targets).

\noindent
\textbf{Learning Based approaches for Active Target Discovery.}
Several RL-based approaches, such as~\citep{uzkent2020learning,sarkar2024visual,sarkar2023partially,nguyen2024amortized}, have been proposed for active target discovery, but these methods rely on full observability of the search space and large-scale pre-labeled datasets for learning an efficient subsampling strategy. Training-free methods inspired by Bayesian decision theory offer a promising approach to active target discovery~\citep{garnett2012bayesian,jiang2017efficient,jiang2019cost}. However, their reliance on full observability of the search space remains a significant limitation, rendering them ineffective in scenarios with partial observability. More Recently, a series of methods, such as~\citep{rangrej2022consistency,pirinen2022aerial,sarkar2024gomaa}, have been proposed to tackle active discovery in partially observable environments. Despite their potential, these approaches face a significant hurdle—the reliance on extensive pre-labeled datasets to fully realize their capabilities. In this work, we present a novel algorithm, \emph{DiffATD}, designed to overcome the limitations of prior approaches by enabling efficient target discovery in partially observable environments. \emph{DiffATD} eliminates the need for pre-labeled training data, significantly enhancing its adaptability and practicality across diverse applications, from medical imaging to remote sensing.

\section{Preliminaries}
Denoising diffusion models are trained to reverse a Stochastic Differential Equation (SDE) that gradually perturbs samples $x$ into a standard normal distribution~\citep{song2020score}. The SDE governing the noising process is given by:
\vspace{-1pt}
\[
dx = -\frac{\beta(\tau)}{2}x \, d\tau + \sqrt{\beta(\tau)} \, dw
\]
\vspace{-1pt}
where \( x_0 \in \mathbb{R}^d \) is an initial clean sample, \( \tau \in [0, T] \), \( \beta(\tau) \) is the noise schedule, and \( w \) is a standard Wiener process, with \( x(T) \sim \mathcal{N}(0, I) \). According to~\citep{anderson1982reverse}, this SDE can be reversed once the score function \( \nabla_x \log p_\tau(x) \) is known, where \( \bar{w} \) is a standard Wiener process running backwards: 
\[
dx = \Bigg[-\frac{\beta(\tau)}{2}x - \beta(\tau) \nabla_x \log p_\tau(x) \,\Bigg] d\tau + \sqrt{\beta(\tau)} \, d\bar{w}
\]
Following~\citep{ho2020denoising,chung2022diffusion}, the discrete formulation of the SDE is defined as \( x_\tau = x(\tau T / N) \), \( \beta_\tau = \beta(\tau T / N) \), \( \alpha_\tau = 1 - \beta_\tau \), and \( \bar{\alpha}_\tau = \prod_{s=1}^\tau \alpha_s \), where \( N \) represents the number of discretized segments. The diffusion model reverses the SDE by learning the score function using a neural network parameterized by \( \theta \), where \( s_\theta(x_\tau, \tau) \approx \nabla_{x_\tau} \log p_\tau(x_\tau) \). The reverse diffusion process can be conditioned on observations gathered so far \( y_t \) to generate samples from the posterior \( p(x \mid y_t) \). This is achieved by substituting \( \nabla_{x_\tau} \log p_\tau(x_\tau) \) with the conditional score function \( \nabla_{x_\tau} \log p_\tau(x_\tau \mid y_t) \) in the above Equation. The difficulty in handling \( \nabla_{x_\tau} \log p_\tau (y_t \mid x_\tau) \), stemming from the application of Bayes’ rule to refactor the posterior, has inspired the creation of several approximate guidance techniques~\citep{song2023pseudoinverse,rout2024solving} to compute these gradients with respect to a partially noised sample \( x_\tau \). Most of these methods utilize \emph{Tweedie's formula}, which serves as a one-step denoising operation from $\tau \to 0$, represented as $\mathcal{T}_\tau(\cdot)$. Using a trained diffusion model, the fully denoised sample $x_0$ can be estimated from its noisy version \( x_\tau \) as follows:
\[
\small{\hat{x}_0 = \mathcal{T}_\tau(x_\tau) = \frac{1}{\sqrt{\bar{\alpha}_\tau}} \left( x_\tau + (1 - \bar{\alpha}_\tau) s_\theta(x_\tau, \tau) \right)}
\]

\section{Problem Formulation}
In this section, we present the details of our proposed Active Target Discovery (ATD) task setup. 
  ATD involves actively uncovering one or more targets within a search area, represented as an region $x$ divided into $N$ grid cells, such that $x = (x^{(1)}, x^{(2)}, ..., x^{(N)})$. ATD operates under a query budget $\mathcal{B}$, representing the maximum number of measurements allowed. Each grid cell represents a sub-region and serves as a potential measurement location. A measurement reveals the content of a specific sub-region $x^{(i)}$ for the $i$-th grid cell, yielding an outcome $y^{(i)}$ $\in$ $[0, 1]$, where $y^{(i)}$ represents the ratio of pixels in the grid cell $x^{(i)}$ that belongs to the target of interest.
In each task configuration, the target's content is initially unknown and is revealed incrementally through observations from measurements. The goal is to identify as many grid cells belonging to the target as possible by strategically exploring the grid within the given budget $\mathcal{B}$.
Denoting a query performed in step $t$ as $q_t$, \emph{the overall task optimization objective is}:
\begin{tcolorbox}[colback=blue!5!white, colframe=blue!40!black, boxrule=0.5pt, arc=2mm]
\begin{equation}\label{eq:objective}
    \small{U(x^{\{(q_t)\}};\{q_t\}) \equiv 
    \max_{\{q_t\}} \sum_{t} y^{(q_t)} \quad\\
    \text{ subject to } \> t \le \mathcal{B}}
\end{equation}
\end{tcolorbox}
With objective~\ref{eq:objective} in focus, we aim to develop a search policy that efficiently explores the search area ($x_\text{test} \sim X_\text{test}$) to identify as many target regions as possible within a measurement budget $\mathcal{B}$, and to achieve this by utilizing a pre-trained diffusion model, trained in an unsupervised manner on samples $x_\text{train}$ from the training set, $X_\text{train}$.


\section{Solution Approach}\label{sec:method}
Crafting a sampling strategy that efficiently balances exploration and exploitation is the key to success in solving ATD. Before delving into exploitation, we first explain how \emph{DiffATD} tackles exploration of the search space using the measurements gathered so far. To this end, \emph{DiffATD} follows a maximum-entropy strategy, by selecting measurement $q^{\text{exp}}_t$, 
\begin{equation}
\label{eq:entropy}
q^{\text{exp}}_t = \arg\max_{q_t} \left[ I(\hat{x}_t; x \mid Q_t, \tilde{x}_{t-1}) \right]
\end{equation}
where \(I\) denotes the mutual information between the full search space \(x\) and the predicted search space \(\hat{x}_t\), conditioned on prior observations \((\tilde{x}_{t-1})\) and the measurement locations \(Q_t = \{q_0, q_1, \ldots, q_t\}\) selected up to time \(t\), where \(q_t\) is the measurement location at time \(t\). 
We now present a proposition that facilitates the simplification of Equation~\ref{eq:entropy}.
\begin{tcolorbox}[colback=blue!5!white, colframe=blue!40!black, boxrule=0.2pt, arc=1mm]
\begin{proposition}\label{eq:mi}
Maximizing $I(\hat{x}_t; x \mid Q_t, \tilde{x}_{t-1})$ w.r.t a measurement location $q_t$ is equivalent to maximizing the marginal entropy of the estimated search space $\hat{x}_t$, i.e.,
\[
\arg\max_{q_t} \left[ I(\hat{x}_t; x \mid Q_t, \tilde{x}_{t-1}) \right] = \arg \max_{q_t} 
\left[ 
H(\hat{x}_t | Q_t, \tilde{x}_{t-1})\right]
\]
\end{proposition}
\end{tcolorbox}
Where $H$ denotes entropy. We derive Prop.~\ref{eq:mi} in the Appendix. 
The marginal entropy $H(\hat{x}_t | Q_t, \tilde{x}_{t-1})$ can be expressed as $-\mathbb{E}_{\hat{x}_t}[\log p(\hat{x}_t | Q_t, \tilde{x}_{t-1})]$, representing the expected log-likelihood of the estimated search space \(\hat{x}_t\) based on the observations collected so far and the set of measurement locations $Q_t$. We refer to this distribution as the \emph{belief distribution}, essential for ranking potential measurement locations from an exploration perspective. Next, we discuss our approach to estimating the \emph{belief distribution}. To this end, \emph{DiffATD} functions by executing a reverse diffusion process for a batch  $\{x^{(i)}_{\tau}\}$ of size $N_B$, where $i \in 0, \ldots, N_B$, 
\begin{algorithm}
\small
\caption{Diffusion Dynamics Guided by Measurements}
\label{alg:guided_diffusion}
\begin{algorithmic}[1]
\Require $T, N_B, s_{\theta}, \zeta, \{\alpha_{\tau}\}_{\tau=0}^T, \{\tilde{\sigma}_{\tau}\}_{\tau=0}^T, M, Q_0 = \emptyset$
\State $t = 0; \>\>\>\> \{x^{(i)}_T \sim \mathcal{N}(0, I)\}_{i=0}^{N_B}; z \sim \mathcal{N}(0, I), \tilde{x}_0=0$
\For{$\tau = T$ to $1$}
    \For{$i = 0$ to $N_B$}
        \State $\hat{x}_{\tau}^{(i)} \gets \mathcal{T}_{\tau}(x_{\tau}^{(i)}) = \frac{1}{\sqrt{\bar{\alpha}_{\tau}}} \left(x_{\tau}^{(i)} + (1 - \bar{\alpha}_{\tau})\hat{s}^{(i)}\right)$; \text{Where} $\> \hat{s}^{(i)} \gets s_\theta(x_{\tau}^{(i)}, \tau)$.
        \State $x_{\tau-1}^{(i)'} \gets \frac{\sqrt{\alpha_{\tau}} (1 - \bar{\alpha}_{\tau-1})}{1 - \bar{\alpha}_{\tau}} x_{\tau}^{(i)} + \frac{\sqrt{\bar{\alpha}_{\tau-1}}\beta_{\tau}}{1 - \bar{\alpha}_{\tau}} \hat{x}_{\tau}^{(i)} + \tilde{\sigma}_{\tau} z$
        \State $x_{\tau-1}^{(i)} \gets x_{\tau-1}^{(i)'} - \zeta \underbrace{\nabla_{x_{\tau}^{(i)}} ||[x]_{Q_t} - [\hat{x}_{\tau}^{(i)}]_{Q_t}||^2}_\textit{Measurement Guidance}$; $[.]_{Q_t}$ \text{select elements of} $[.]$ \text{indexed by the set} $Q_t$.
    \EndFor
    \If{$\tau \in M$}
        \State $t \gets t + 1$
        \State Acquire new measurement at location $q_t$ and update: $Q_t \gets Q_{t-1} \cup q_t$, $\tilde{x}_t \gets \tilde{x}_{t - 1} \cup [x]_{q_t} $
    \EndIf
\EndFor
\end{algorithmic}
\end{algorithm}
guided by an incremental set of observations $\{\tilde{x}_t\}_{t=0}^{\mathcal{B}}$, as detailed in Algorithm~\ref{alg:guided_diffusion}. 
These observations are obtained through measurements collected at reverse diffusion steps $\tau$, that satisfy \( \tau \in M \), where $M$ is a measurement schedule. 
 An element of this batch, $x^{(i)}_{\tau}$, is referred to as \emph{particle}. These \emph{particles} implicitly form a \emph{belief distribution} over the entire search space $x$ throughout reverse diffusion and are used to estimate uncertainty about $x$. Concretely, we model the belief distribution simply as mixtures of $N_B$ isotropic Gaussians, with variance $\sigma^2_x$I, and mean $\hat{x}^{(i)}_t$ computed via Tweedie's operator $\mathcal{T}_{\tau}(\{x^{(i)}_{\tau}\})$ as follows:
\begin{equation}\label{eq:plan}
\hat{x}^{(i)}_t = \mathbb{E}[\hat{x}^{(i)}_{\tau} \mid x^{(i)}_{\tau}] = \frac{1}{\sqrt{\bar{\alpha}_{\tau}}} \left( x^{(i)}_{\tau} + (1 - \bar{\alpha}_{\tau}) s_\theta(x^{(i)}_{\tau}, \tau) \right)
\end{equation}
We assume that the reverse diffusion step $\tau$ corresponds to the measurement step $t$. Utilizing the expression~\ref{eq:plan}, we can express the \emph{belief distribution} as follows:
\begin{equation}
\small{p(\hat{x}_t | Q_t, \tilde{x}_{t-1}) = \sum_{i=0}^{N_B} \alpha_i \mathcal{N}(\hat{x}^i_t, \sigma^2_xI )}
\end{equation}
According to~\citep{hershey2007approximating}, the marginal entropy of this \emph{belief distribution} can be estimated as follows:
\begin{equation}\label{eq:mar_ent}
  H(\hat{x}_t | Q_t, \tilde{x}_{t-1}) \propto \sum^{N_B}_{i=0} \alpha_i \text{log} \sum^{N_B}_{j=0} \alpha_j \text{exp} \left\{ {\frac{||\hat{x}^{(i)}_t - \hat{x}^{(j)}_t||^2_2}{2\sigma^2_x}} \right\}
\end{equation}
As per Eqn.~\ref{eq:mar_ent}, uncovering the optimal measurement location ($q^{\text{exp}}_t$) requires computing a separate set of $\hat{x}_t$ for each possible choice of $q_t$.
Next, we present a theorem that empowers us to select the next measurement as the region with the highest total variance across $\{ \hat{x}^{(i)}_t\}$ estimated from each particle in the batch conditioned on ($Q_t, \tilde{x}_{t-1}$), effectively bypassing the need to compute a separate set of predicted search space $\hat{x}_t$ for each potential measurement location $q_t$.
\begin{tcolorbox}[colback=blue!5!white, colframe=blue!40!black, boxrule=0.2pt, arc=1mm]
\begin{theorem}\label{eq:argmax_error}
Assuming \( k \) represents the set of possible measurement locations at step $t$, and that all particles have equal weights (\( \alpha_i = \alpha_j, \, \forall i, j \)), then $q^{\text{exp}}_t$ is given by:
\[
\arg\max_{q_t} \left[ \sum_{i=0}^{N_B} \log \sum_{j=0}^{N_B} \exp \left( \frac{ \sum_{q_t \in k}^{ }([\hat{x}^{(i)}_t]_{q_t} - [\hat{x}^{(j)}_t]_{q_t})^2 }{2\sigma_x^2} \right) \right],
\]
where $[.]_{q_t}$ selects element of $[.]$ indexed by $q_t$. 
\end{theorem}
\end{tcolorbox}
We prove this result in Appendix.
Its consequence is that the optimal choice for the next measurement location ($q^{\text{exp}}_t$) lies in the region of the measurement space where the predicted search space ($\hat{x}^{(i)}_t$) exhibits the greatest disagreement across the \emph{particles} ($i \in 0, \ldots, N_B$), quantified by a metric, denoted as $\mathrm{expl}^{\mathrm{score}}(q_t)$, which enables ranking potential measurement locations to optimize exploration efficiency. 
In cases where the measurement space consists of pixels, $\hat{x}^{(i)}_t$ represents predicted estimates of the full image based on observed pixels $\tilde{x}_{t-1}$ so far. 
\begin{figure*}[htbp]
    \centering
    \includegraphics[width=0.92\textwidth]{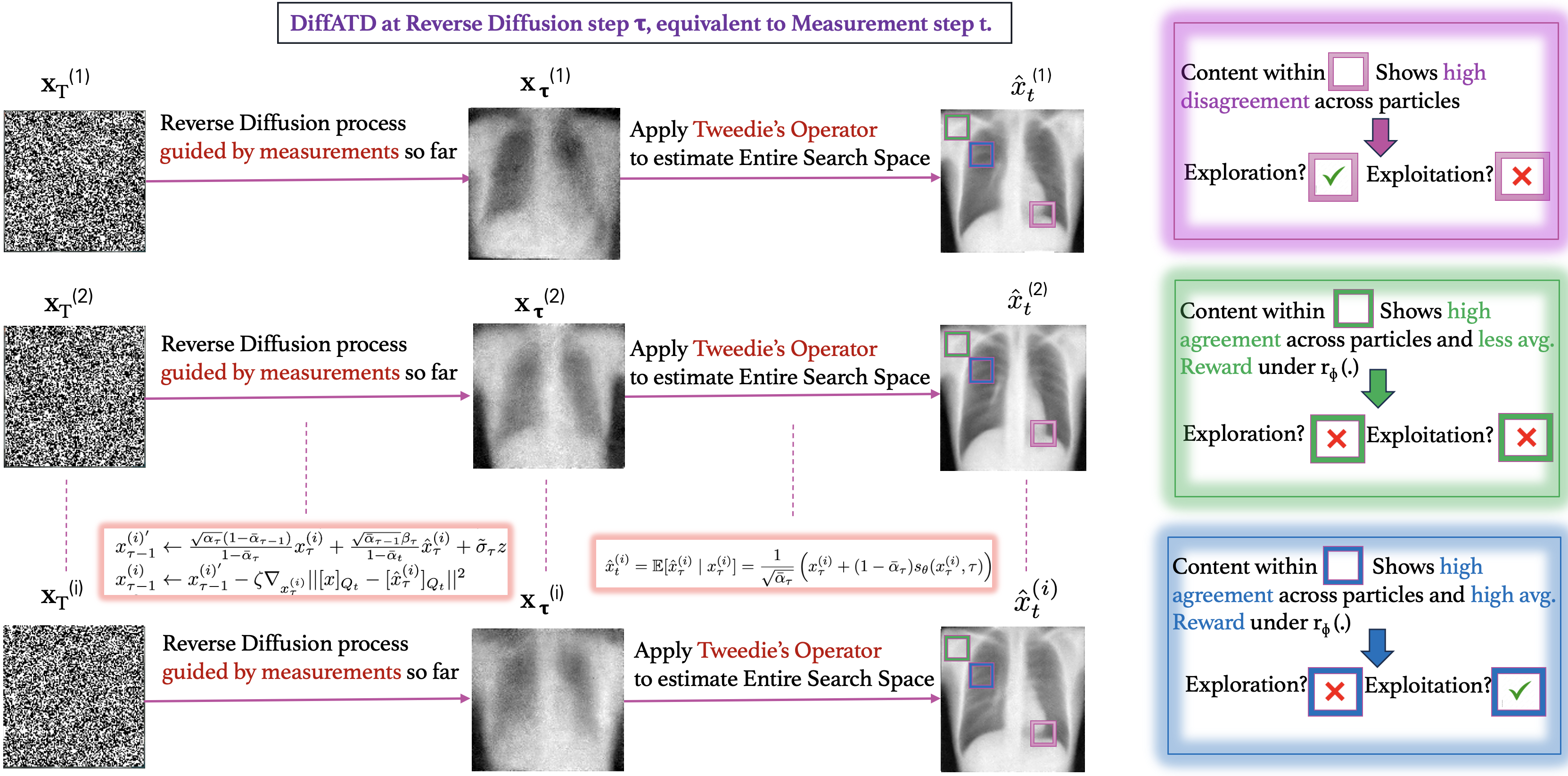}
    \caption{An Overview of \emph{DiffATD}.}
    \label{fig:diffatd}
\end{figure*}
\begin{equation}\label{eq:exp-ecore}
\mathrm{expl}^{\mathrm{score}}(q_t) = \left[ \sum_{i=0}^{N_B} \sum_{j=0}^{N_B} \frac{([\hat{x}^{(i)}_t]_{q_t} - [\hat{x}^{(j)}_t]_{q_t})^2}{2\sigma_x^2} \right]
\end{equation}
We now delve into how \emph{DiffATD} leverages the observations collected so far to address exploitation effectively. To this end, \emph{DiffATD}: $(i)$ employs a \emph{reward model} ($r_{\phi}$) parameterized by $\phi$, which is incrementally trained on supervised data obtained from measurements, enabling it to predict whether a specific measurement location belongs to the target of interest; $(ii)$ estimates $\mathbb{E}_{\hat{x}_t}[ \log p(\hat{x}_t | Q_t, \tilde{x}_{t-1})]_{q_t}$, representing the expected log-likelihood score at a measurement location $q_t$, denoted as $\mathrm{likeli}^{\mathrm{score}}(q_t)$,
as evaluating it is essential for prioritizing potential measurement locations from an exploitation standpoint. The following theorem establishes an expression for computing $\mathrm{likeli}^{\mathrm{score}}(q_t)$.
\begin{tcolorbox}[colback=blue!5!white, colframe=blue!40!black, boxrule=0.2pt, arc=1mm]
\begin{theorem}\label{eq:like-ecore}
The expected log-likelihood score at a measurement location $q_t$ can be expressed:
\[
\mathrm{likeli}^{\mathrm{score}}(q_t) =  \sum_{i=0}^{N_B} \sum_{j=0}^{N_B} \exp \left\{ - \frac{([\hat{x}^{(i)}_t]_{q_t} - [\hat{x}^{(j)}_t]_{q_t})^2}{2\sigma_x^2} \right\}
\]
\end{theorem}
\end{tcolorbox}
We present the derivation of Theorem~\ref{eq:like-ecore} in the Appendix. \( \mathrm{likeli}^{\mathrm{score}}(q_t) \) serves as a key factor in calculating the exploitation score at measurement location $q_t$, where the exploitation score, \( \mathrm{exploit}^{\mathrm{score}}(q_t) \), is defined as \emph{the reward-weighted expected log-likelihood}, as shown below:
\begin{equation}\label{eq:exploit-score}
\mathrm{exploit}^{\mathrm{score}}(q_t) = \underbrace{\mathrm{likeli}^{\mathrm{score}}(q_t)}_\textit{Expected log-likelihood} \times \underbrace{\sum^{N_B}_{i=0} r_{\phi}([\hat{x}^{(i)}_t]_{q_t})}_\textit{reward}
\end{equation}
According to Eqn.~\ref{eq:exploit-score}, a measurement location $q_t$ is preferred for exploitation if it satisfies two conditions: (1) the predicted content at $q_t$, across the \emph{particles}, is highly likely to correspond to the target of interest, as determined by the reward model ($r_{\phi}$) that predicts whether a specific measurement location belongs to the target; (2) the predicted content at $q_t$, denoted as $[\hat{x}^{(i)}_t]_{q_t}$, shows the highest degree of consensus among the \emph{particles}, indicating that the content at $q_t$ is highly predictable. Additionally, we demonstrate that maximizing the exploitation score is synonymous with maximizing the expected reward at the current time step, highlighting that the exploitation score operates as a purely greedy strategy. We establish the relationship between \( \mathrm{exploit}^{\mathrm{score}}(q_t) \) and a pure greedy strategy and present the proof in the Appendix.
\begin{tcolorbox}[colback=blue!5!white, colframe=blue!40!black, boxrule=0.2pt, arc=1mm]
\begin{theorem}
Maximizing $\mathrm{exploit}^{\mathrm{score}}(q_t)$ w.r.t a measurement location $q_t$ is equivalent to maximizing expected reward as stated below:
\( \arg\max_{q_t} \left[ \mathrm{exploit}^{\mathrm{score}}(q_t) \right] \equiv \arg\max_{q_t} \mathbb{E}_{\hat{x}_t \sim p(\hat{x}_t | Q_t, \tilde{x}_{t-1})} [r_{\phi}[\hat{x}_t]_{q_t}] \).
\end{theorem}
\end{tcolorbox}

It is important to note that we randomly initialize the reward model parameters ($\phi$) and update them progressively using \emph{binary cross-entropy loss} as new supervised data is acquired after each measurement. The training objective of reward model $r_{\phi}$ at measurement step $t$ is defined as below:
\begin{equation}\label{eq:reward}
\min_{\phi} [ \sum_{i=1}^{t} -(y^{(q_i)} \log (r_{\phi}([x]_{q_t})) + (1-y^{(q_i)}) \log (1-r_{\phi}([x]_{q_t}))
\end{equation}
Initially, the reward model is imperfect, making the exploitation score unreliable. However, this is not a concern, as \emph{DiffATD} prioritizes exploration in the early stages of the discovery process. Next, we integrate the components to derive the proposed \emph{DiffATD} sampling strategy for efficient active target discovery. This involves ranking each potential measurement location ($q_t$) at time $t$ by considering both exploration and exploitation scores, as defined below:
\begin{equation}\label{eq:final_score}
    \mathrm{Score}(q_t) = \kappa(\mathcal{B}) \cdot \mathrm{expl}^{\mathrm{score}}(q_t) + (1 - \kappa(\mathcal{B})) \cdot \mathrm{likeli}^{\mathrm{score}}(q_t)
\end{equation}
Here, $\kappa(\mathcal{B})$ is a function of the measurement budget that enables \emph{DiffATD} to achieve a budget-aware balance between exploration and exploitation.
After computing the scores (as defined in~\ref{eq:final_score}) for each measurement location $q_t$, \emph{DiffATD} selects the location with the highest score for sampling. The functional form of $\kappa(\mathcal{B})$ is a design choice that depends on the specific problem. For instance, a simple approach is to define $\kappa(\mathcal{B})$ as a linear function of the budget, such as $\kappa(\mathcal{B}) = \frac{\mathcal{B} - t}{\mathcal{B} + t}$. We empirically demonstrate that this simple choice of $\kappa(\mathcal{B})$ is highly effective across a wide range of application domains. We detail the \emph{DiffATD} algorithm in~\ref{alg:DiffATD} and an overview in Figure~\ref{fig:diffatd}.
\begin{algorithm}
\small
\caption{DiffATD Algorithm}
\label{alg:DiffATD}
\begin{algorithmic}[1]
\Require $T, N_B, s_{\theta}, \zeta, \{\alpha_{\tau}\}_{\tau=0}^T, \{\tilde{\sigma}_{\tau}\}_{\tau=0}^T, M, Q_0 = \emptyset, \mathcal{B}$
\State $t = 0; \>\>\>\> \{x^{(i)}_T \sim \mathcal{N}(0, I)\}_{i=0}^{N_B}; z \sim \mathcal{N}(0, I), \tilde{x}_0=0$, $\mathcal{D}_{t} = \emptyset, \{k\} = \text{Set of All measurements}, R=0$
\For{$\tau = T$ to $1$}
    \For{$i = 0$ to $N_B$}
        \State $\hat{x}_{\tau}^{(i)} \gets \mathcal{T}_{\tau}(x_{\tau}^{(i)}) = \frac{1}{\sqrt{\bar{\alpha}_{\tau}}} \left(x_{\tau}^{(i)} + (1 - \bar{\alpha}_{\tau})\hat{s}^{(i)}\right)$; \text{Where} $\>$ $\hat{s}^{(i)} \gets s_\theta(x_{\tau}^{(i)}, \tau)$
        \State $x_{\tau-1}^{(i)'} \gets \frac{\sqrt{\alpha_{\tau}} (1 - \bar{\alpha}_{\tau-1})}{1 - \bar{\alpha}_{\tau}} x_{\tau}^{(i)} + \frac{\sqrt{\bar{\alpha}_{\tau-1}}\beta_{\tau}}{1 - \bar{\alpha}_t} \hat{x}_{\tau}^{(i)} + \tilde{\sigma}_{\tau} z$
        \State $x_{\tau-1}^{(i)} \gets x_{\tau-1}^{(i)'} - \zeta \nabla_{x_{\tau}^{(i)}} ||[x]_{Q_t} - [\hat{x}_{\tau}^{(i)}]_{Q_t}||^2$
    \EndFor
    \If{$\tau \in M$ and $\mathcal{B} > 0$ }
        \State Compute $\mathrm{expl}^{\mathrm{score}}(q_{t})$ and $\mathrm{exploit}^{\mathrm{score}}(q_t)$ using Eqn.~\ref{eq:exp-ecore} and~\ref{eq:exploit-score} respectively for each $q_t \in k$.
        \State Compute $\mathrm{score}(q_t)$ using Eqn.~\ref{eq:final_score} for each $q_t \in k$ and sample a location $q_t$ with the highest $\mathrm{score}$.
        \State $t \gets t + 1$, $\mathcal{B} \gets \mathcal{B} - 1$, $\{ k \} \gets \{ k \} \setminus q_t$
        \State Update: $Q_{t} \gets Q_{t-1} \cup q_{t}$, $\tilde{x}_t \gets \tilde{x}_{t - 1} \cup [x]_{q_t}.$; Update: $\mathcal{D}_{t} \gets \mathcal{D}_{t-1} \cup \{[x]_{q_{t}}, y^{(q_{t})}\}, R$ += $y^{(q_t)} $
        \State Train $r_{\phi}$ with updated $\mathcal{D}_{t}$ and optimize $\phi$ with~\ref{eq:reward}.
    \EndIf
\EndFor
\State \textbf{Return} R.
\end{algorithmic}
\end{algorithm}
\vspace{-18pt}




\section{Experiments}\label{sec:experiments}
\vspace{-3pt}
\textbf{Evaluation metrics.}
Since ATD seeks to maximize the identification of measurement locations containing the target of interest, we assess performance using the \emph{\textbf{success rate (SR) of selecting measurement locations that belong to the target}} during exploration in partially observable environments. Therefore, SR is defined as:
\begin{tcolorbox}[colback=blue!5!white, colframe=blue!40!black, boxrule=0.5pt, arc=2mm]
\begin{equation}
\small{\mathrm{SR} = \frac{1}{L}\sum_{i=1}^{L}\frac{1}{\min{\{\mathcal{B}, U_i}\}}\sum_{t=1}^{\mathcal{B}} y_{i}^{(q_t)}; L = \text{number of tasks.}}
\end{equation}
\end{tcolorbox}
Here, $U_i$ denotes the maximum number of measurement locations containing the target in the $i$-th search task. We evaluate \emph{DiffATD} and the baselines across different measurement budgets $\mathcal{B} \in \{ 150, 200, 250, 300\}$ for various target categories and application domains, spanning from medical imaging to remote sensing. Our code and models are publicly available at this \href{https://github.com/KevinG396/DiffATD}{\textcolor{blue}{link}}.
\\ \\
\noindent
\textbf{Baselines.} 
We compare our proposed \emph{DiffATD} policy to the following baselines:
\begin{itemize}[noitemsep,topsep=0pt,leftmargin=*]
    \item \emph{\textbf{Random Search} (RS)}, in which unexplored measurement locations are selected uniformly at random.
    \item \emph{\textbf{Max-Ent}} approach ranks a set of unexplored measurement locations based on $\mathrm{expl}^{\mathrm{score}}(q_t)$ and then selects $q_t$ corresponding to the maximum value of $\mathrm{expl}^{\mathrm{score}}(q_t)$.  
    \item \emph{\textbf{Greedy-Adaptive} (GA)} approach ranks a set of unexplored measurement locations based on $\mathrm{exploit}^{\mathrm{score}}(q_t)$ and then selects $q_t$ corresponding to the maximum value of $\mathrm{exploit}^{\mathrm{score}}(q_t)$. Like \emph{DiffATD}, GA approach updates ($r_\phi$) using Equation~\ref{eq:reward} after each observation.
\end{itemize}
\noindent
\vspace{-7pt}
\paragraph{Active Discovery of Objects from Overhead Imagery} 
We begin the evaluation by comparing the performance of \emph{DiffATD} with the baselines in terms of SR. For this comparison, we consider various targets (e.g., truck, plane) from the DOTA dataset~\citep{xia2018dota} 
and present the results in Table~\ref{tab: dota}. 
The empirical outcome indicates that \emph{DiffATD} significantly outperforms all baselines, with improvements ranging from $\textbf{7.01\%}$ to $\textbf{17.23\%}$ over the most competitive method across all measurement budgets. We also present visualizations of \emph{DiffATD}'s exploration strategy in Fig.~\ref{fig:vis2}, demonstrating its effectiveness in ATD. As depicted in Fig.~\ref{fig:vis2}, \emph{DiffATD} efficiently explores the search space, uncovering distinct target clusters within the measurement budget, highlighting its effectiveness in balancing exploration and exploitation. Similar visualizations with different target categories are in the Appendix.
\vspace{-8pt}
\begin{center}
\begin{minipage}{0.48\textwidth}
  \centering
  \includegraphics[width=\linewidth]{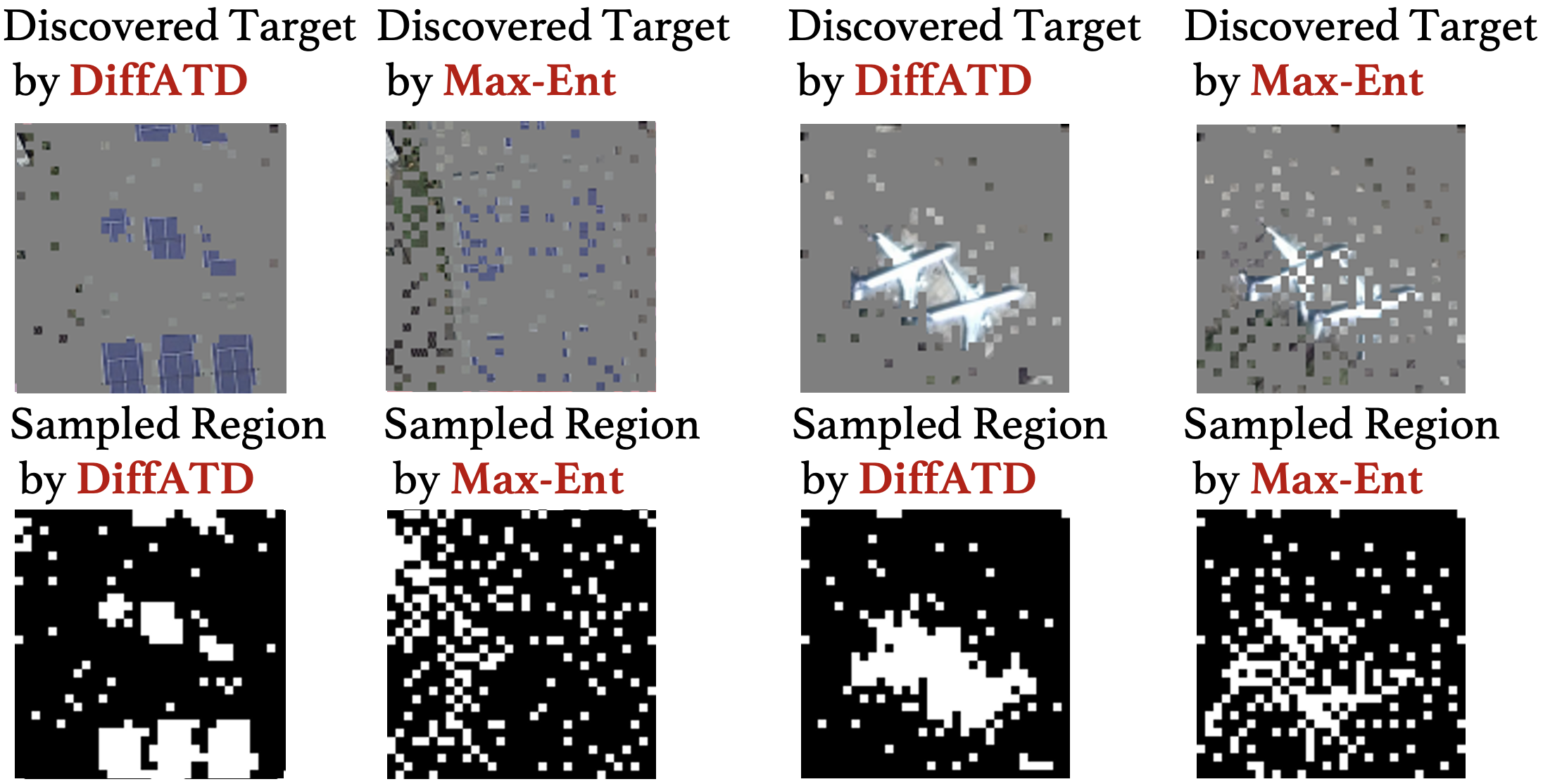}
  \captionof{figure}{\small{Active Discovery of Plane, Playground.}}
  \label{fig:vis2}
\end{minipage}
\hfill
\begin{minipage}{0.50\textwidth}
  \centering
  \footnotesize
  \captionof{table}{\emph{SR} Comparison with Baseline Approaches}
  \vspace{-2pt}
  \begin{tabular}{p{1.33cm}p{1.33cm}p{1.33cm}p{1.33cm}}
    \toprule
    \multicolumn{4}{c}{Active Discovery with targets, e.g., plane, truck} \\
    \midrule
    Method & $\mathcal{B}=200$ & $\mathcal{B}=250$ & $\mathcal{B}=300$ \\
    \midrule
    RS & 0.1990 & 0.2487 & 0.2919 \\
    Max-Ent & 0.4625 & 0.5524 & 0.6091 \\
    GA & 0.4586 & 0.5961 & 0.6550 \\
    \hline 
    \textbf{\emph{DiffATD}} & \textbf{0.5422} & \textbf{0.6379} & \textbf{0.7309} \\ 
    \bottomrule
  \end{tabular}
  \label{tab: dota}
\end{minipage}
\end{center}
\vspace{-11pt}
\paragraph{Active Discovery of Species}
Next, we evaluate the efficacy of \emph{DiffATD} in uncovering an unknown species from iNaturalist~\citep{inaturalist}. The target species distribution is formed by dividing a large geospatial region into equal-sized grids and counting target species occurrences on each grid, as detailed in Appendix. We compare \emph{DiffATD} with the baselines in terms of \emph{SR}, and across varying $\mathcal{B}$. The results are presented in Table~\ref{tab: species}. We observe significant improvements in the performance of the proposed \emph{DiffATD} approach compared to all baselines in each measurement budget setting, ranging from $~\textbf{8.48\%}$ to $~\textbf{13.86\%}$ improvement relative to the most competitive method. In Fig.~\ref{fig:visspe}, we present a qualitative comparison of the exploration strategies of \emph{DiffATD} and GA, illustrating that \emph{DiffATD} uncovers more target clusters compared to the GA approach within a fixed sampling budget.  
\vspace{-6pt}
\begin{center}
\begin{minipage}{0.48\textwidth}
  \vspace{-4pt}
  \centering
  \includegraphics[width=\linewidth]{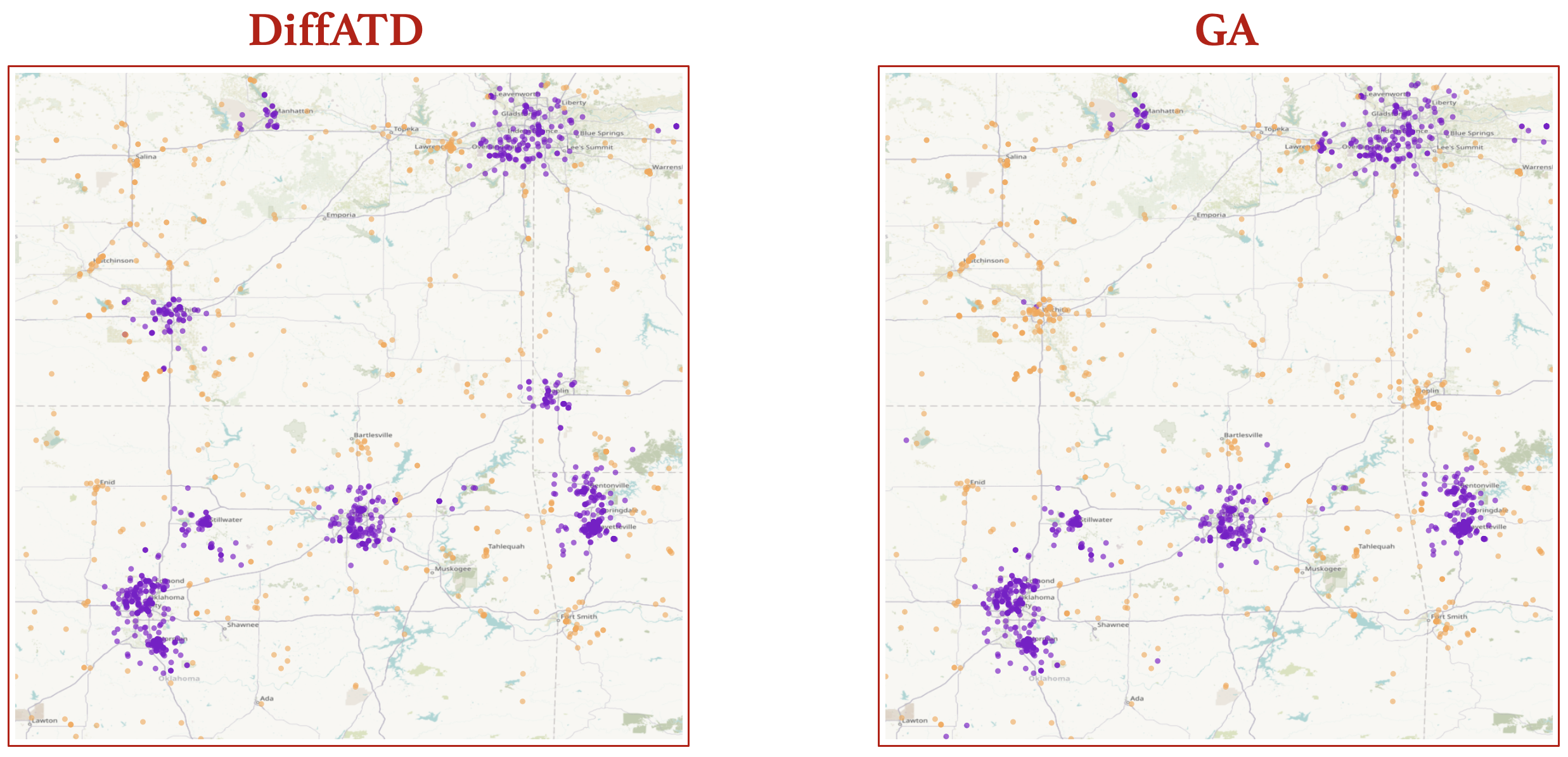}
  \captionof{figure}{\small{ATD of Species (purple $\rightarrow$ targets).}}
  \label{fig:visspe}
\end{minipage}
\hfill
\begin{minipage}{0.50\textwidth}
  \centering
  \footnotesize
  \captionof{table}{\emph{SR} Comparison with Baseline Approaches}
  \vspace{-2pt}
  \begin{tabular}{p{1.33cm}p{1.33cm}p{1.33cm}p{1.33cm}}
    \toprule
    \multicolumn{4}{c}{Active Discovery of Coccinella Septempunctata} \\
    \midrule
    Method & $\mathcal{B}=100$ & $\mathcal{B}=150$ & $\mathcal{B}=200$ \\
    \midrule
    RS & 0.1241 & 0.1371 & 0.1932 \\
    Max-Ent & 0.4110 & 0.5044 & 0.5589 \\
    GA & 0.4345 & 0.5284 & 0.5826 \\
    \textbf{\emph{DiffATD}} & \textbf{0.4947} & \textbf{0.5732} & \textbf{0.6401} \\
    \bottomrule
  \end{tabular}
  \label{tab: species}
\end{minipage}
\end{center}
\vspace{-9pt}
\paragraph{Active Discovery of Skin Disease}
Next, we test the impact of \emph{DiffATD} in the medical imaging setting. To this end, we compare the performance of \emph{DiffATD} with the baselines in terms of \emph{SR} using target classes from the skin imaging dataset~\cite{rotemberg2021patient}. We compare the performance across varying measurement budgets $\mathcal{B}$. The results are presented in Table~\ref{tab: skin}. We observe significant improvements in the performance of the proposed \emph{DiffATD} approach compared to all baselines in each measurement budget setting, ranging from $~\textbf{7.92\%}$ to $~\textbf{10.18\%}$ improvement relative to the most competitive method. In Fig.~\ref{fig:vis1}, we present a qualitative comparison of the exploration strategies of \emph{DiffATD} and Max-Ent. 
As shown in Fig.~\ref{fig:vis1}, \emph{DiffATD} efficiently explores the search space, uncovering two distinct target clusters within the measurement budget, highlighting its effectiveness in balancing exploration and exploitation. Several such visualizations and the results with other datasets are in Appendix.
\begin{center}
\begin{minipage}{0.48\textwidth}
  \centering
  \includegraphics[width=\linewidth]{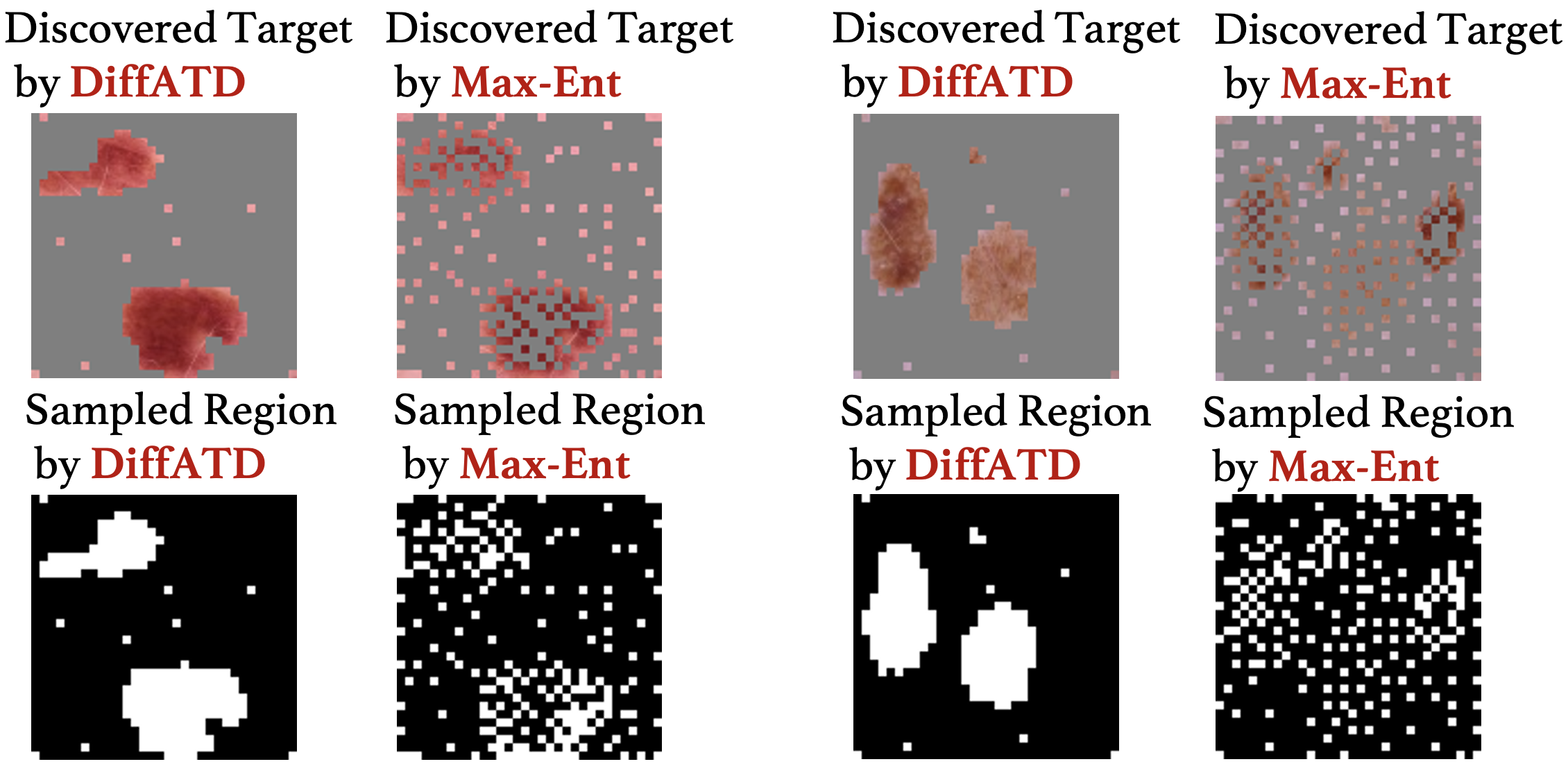}
  \captionof{figure}{\small{Active Discovery of Skin Disease.}}
  \label{fig:vis1}
\end{minipage}
\hfill
\begin{minipage}{0.50\textwidth}
  \centering
  \footnotesize
  \captionof{table}{\emph{SR} Comparison with Baselines.}
  \vspace{-2pt}
  \begin{tabular}{p{1.33cm}p{1.33cm}p{1.33cm}p{1.33cm}}
    \toprule
    \multicolumn{4}{c}{Active Discovery of Malignant Skin Diseases} \\
    \midrule
    Method & $\mathcal{B}=150$ & $\mathcal{B}=200$ & $\mathcal{B}=250$ \\
    \midrule
    RS & 0.2777 & 0.2661 & 0.2695 \\
    Max-Ent & 0.5981 & 0.5816 & 0.5717 \\
    GA & 0.8396 & 0.8261 & 0.7943 \\
    \textbf{\emph{DiffATD}} & \textbf{0.9061} & \textbf{0.8974} & \textbf{0.8752} \\
    \bottomrule
  \end{tabular}
  \label{tab: skin}
\end{minipage}
\end{center}

\paragraph{Active Discovery of Bone Suppression}\label{app:chest}
Next, we evaluate \emph{DiffATD} on the Chest X-Ray dataset~\cite{van2006segmentation}, with results shown in Table~\ref{tab: Chest}. The findings follow a similar trend to previous datasets, with \emph{DiffATD} achieving notable performance improvements of $~\textbf{41.08\%}$ to $~\textbf{59.22\%}$ across all measurement budgets. These empirical outcomes further reinforce the efficacy of \emph{DiffATD} in active target discovery. Visualizations of \emph{DiffATD}'s exploration strategy are provided in Fig.~\ref{fig:vis4}, with additional examples are presented in the Appendix.
\vspace{-7pt}
\begin{center}
\begin{minipage}{0.48\textwidth}
  \centering
  \includegraphics[width=\linewidth]{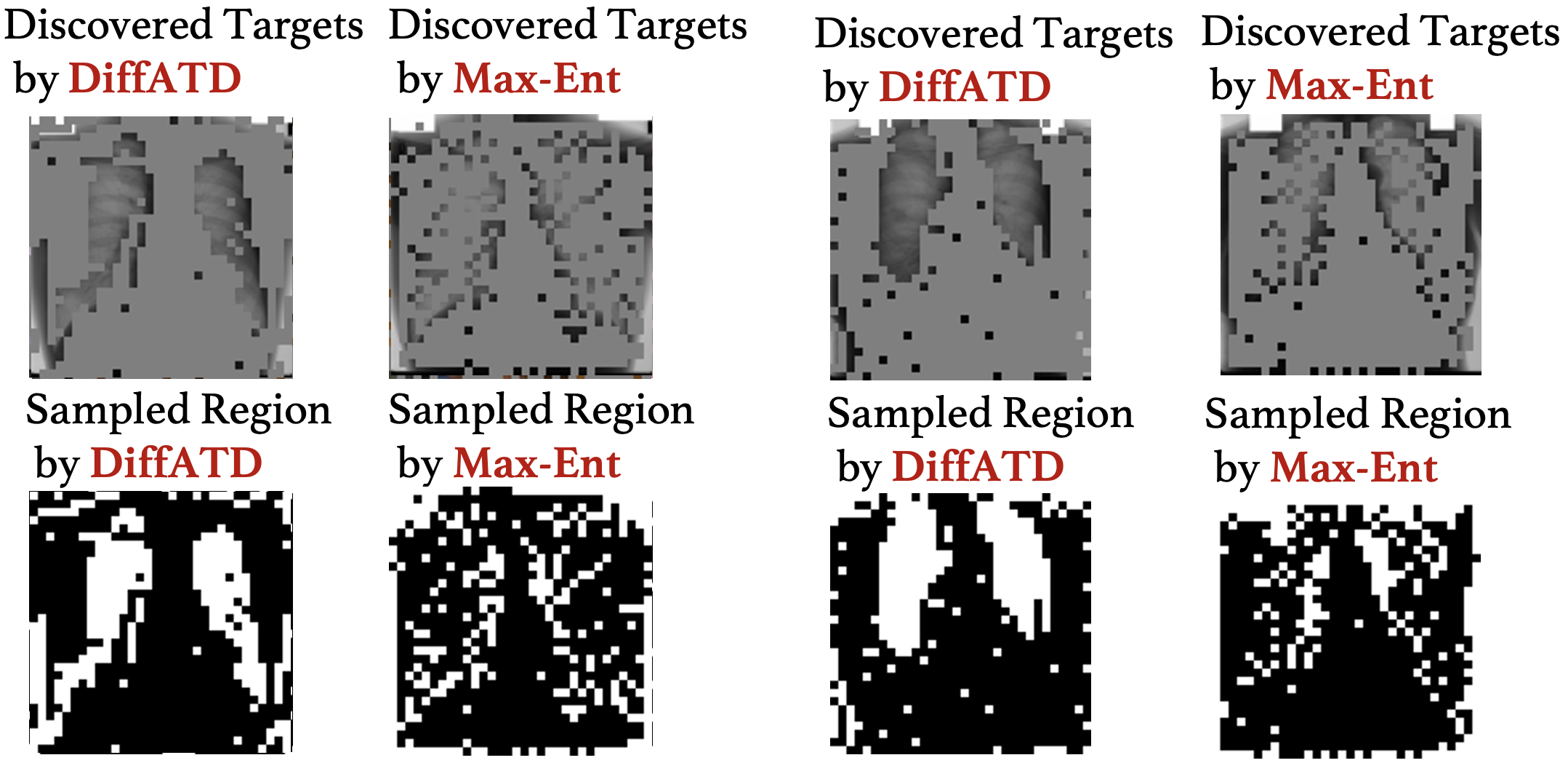}
  \captionof{figure}{\small{Active Discovery of Bone Suppression.}}
  \label{fig:vis4}
\end{minipage}
\hfill
\begin{minipage}{0.50\textwidth}
  \centering
  \footnotesize
  \captionof{table}{\emph{SR} Comparison with Chest X-Ray Dataset}
  \vspace{-2pt}
  \begin{tabular}{p{1.33cm}p{1.33cm}p{1.33cm}p{1.33cm}}
    \toprule
    \multicolumn{4}{c}{Active Discovery of Bone Suppression} \\
    \midrule
    Method & $\mathcal{B}=200$ & $\mathcal{B}=250$ & $\mathcal{B}=300$ \\
    \midrule
    RS & 0.1925 & 0.2501 & 0.2936  \\
    Max-Ent & 0.1643 & 0.2194 & 0.2616 \\
    GA & 0.1607 & 0.2010 & 0.2211  \\
    \hline 
    \textbf{\emph{DiffATD}} & \textbf{0.3065} & \textbf{0.3733} & \textbf{0.4142}  \\ 
    \bottomrule
  \end{tabular}
  \label{tab: Chest}
\end{minipage}
\end{center}

\vspace{-8pt}
\paragraph{Qualitative Comparison with GA Approach}
We compare the exploration strategy of \emph{DiffATD} 
\begin{wrapfigure}{r}{0.5\textwidth}
    \centering
    \includegraphics[width=0.5\textwidth]{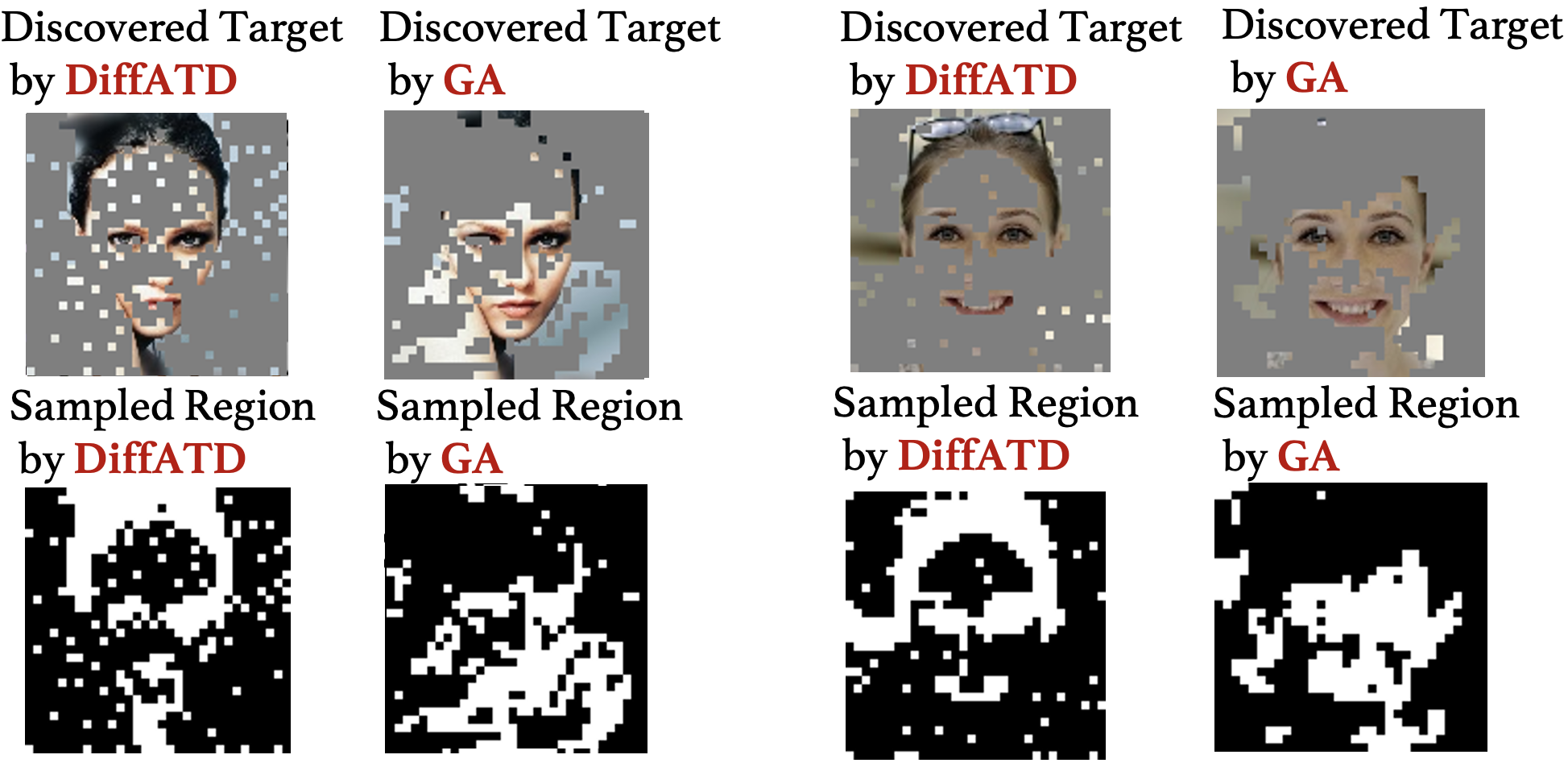}
    \caption{Active Discovery of Eye, hair, nose, lip.}
    \label{fig:GA_main}
    \vspace{-10pt}
\end{wrapfigure}
with that of \emph{Greedy-Adaptive} (GA) using CelebA samples, as shown in Fig.~\ref{fig:GA_main}. \emph{DiffATD} achieves a better balance between exploration and exploitation, leading to higher success rates and more efficient target discovery within a fixed budget. In contrast, GA focuses solely on exploitation, relying on an incrementally learned reward model $r_{\phi}$. 
These visual comparisons highlight the limitations of purely greedy approaches, such as GA, especially when targets have complex or irregular structures with spatially isolated regions. \textbf{GA often struggles early in the search, when training samples are limited, the reward model tends to overfit noise, misguiding the discovery process.} In contrast, \emph{DiffATD} minimizes early reliance on reward-based exploitation, gradually increasing it as more measurements are collected and the reward model strengthens. This adaptive process makes \emph{DiffATD} more resilient and reliable than GA. More qualitative results are in the Appendix.
\vspace{-4pt}
\paragraph{Effect of $\kappa(\mathcal{B}$) }\label{par:kappa}
We conduct experiments to assess the impact of $\kappa(\mathcal{B})$ on \emph{DiffATD}'s active discovery performance. Specifically, we investigate how amplifying the exploration weight,
\begin{wraptable}{r}{0.4\textwidth}
  \vspace{-10pt} 
  \scriptsize
  \caption{Effect of $\kappa(\mathcal{B})$}
  \label{tab: balance}
  \begin{tabular}{p{1cm}p{1cm}p{1cm}p{1cm}}
    \toprule
    \multicolumn{4}{c}{Performance across varying $\alpha$ with $\mathcal{B}=200$} \\
    \midrule
    Dataset & $\alpha=0.2$ & $\alpha=1.0$ & $\alpha=5.0$ \\
    \midrule
    DOTA & 0.5052 & \textbf{0.5422} & 0.4823 \\ 
    Skin & 0.8465 & \textbf{0.8974} & 0.8782 \\
    \bottomrule
  \end{tabular}
  \vspace{-10pt} 
\end{wraptable}
 by setting $\kappa(\mathcal{B}) = \max{\{0, \kappa(\alpha \cdot \mathcal{B})\}}$ with $\alpha > 1$, and enhancing the exploitation weight by setting $\alpha < 1$, influence the overall effectiveness of the approach. 
We present results for $\alpha \in \{ 0.2, 1, 5\}$ using the DOTA and Skin datasets, as shown in Table~\ref{tab: balance}. The best performance is achieved with $\alpha = 1$, and the results suggest that \textbf{extreme values of $\alpha$ (either too low or too high) hinder performance}, as both exploration and exploitation are equally crucial to achieve superior performance.
\vspace{-8pt}
\paragraph{Comparison with Supervised and Fully Observable Approach}
To evaluate the efficacy of the proposed \emph{DiffATD} framework, we compare its performance to a fully supervised SOTA
\begin{wraptable}{l}{0.27\textwidth}
    \centering
    \footnotesize
    \begin{tabular}{p{1.1cm}p{0.8cm}p{0.8cm}}
        \toprule
        \multicolumn{3}{c}{\small{Active Discovery of Targets}} \\
        \midrule
        Method & \footnotesize{Skin} & \footnotesize{DOTA} \\
        \midrule
        \emph{FullSEG} & 0.6221  & \textbf{0.8731}  \\ 
        \emph{DiffATD} & \textbf{0.8642}  & 0.7309  \\ 
        \bottomrule
    \end{tabular}
    \caption{\small{SR Comparison.}}
    \label{tab:full_obs}
    \vspace{-3pt}
\end{wraptable} 
semantic segmentation approach, SAM~\cite{kirillov2023segment}, that operates under full observability of the search space, referred to as \emph{FullSEG}.
Note that during inference, \emph{FullSEG} selects the top $\mathcal{B}$ most probable target regions for measurement in a one-shot manner. We present the result for diverse target categories with $\mathcal{B} = 300$ in Table~\ref{tab:full_obs}.
 We observe that \textbf{\emph{DiffATD} outperforms SAM for rare targets}, such as skin diseases, while performing comparably to \emph{FullSEG} for non-rare targets like vehicles, planes, etc., despite never observing the full search area and being trained entirely in an unsupervised manner, unlike \emph{FullSEG}. This further showcases the strength of \emph{DiffATD} in active target discovery in a partially observable environment.
\vspace{-3pt}
\paragraph{Visualizing \emph{DiffATD}'s Exploration Behavior at Different Stage}\label{sec:B}
We illustrate the behavior of \emph{DiffATD} with an example. As shown in Fig.~\ref{fig:explain}, during the initial search phase, \emph{DiffATD} prioritizes
\begin{figure}
    \centering
    \includegraphics[width=0.9\textwidth]{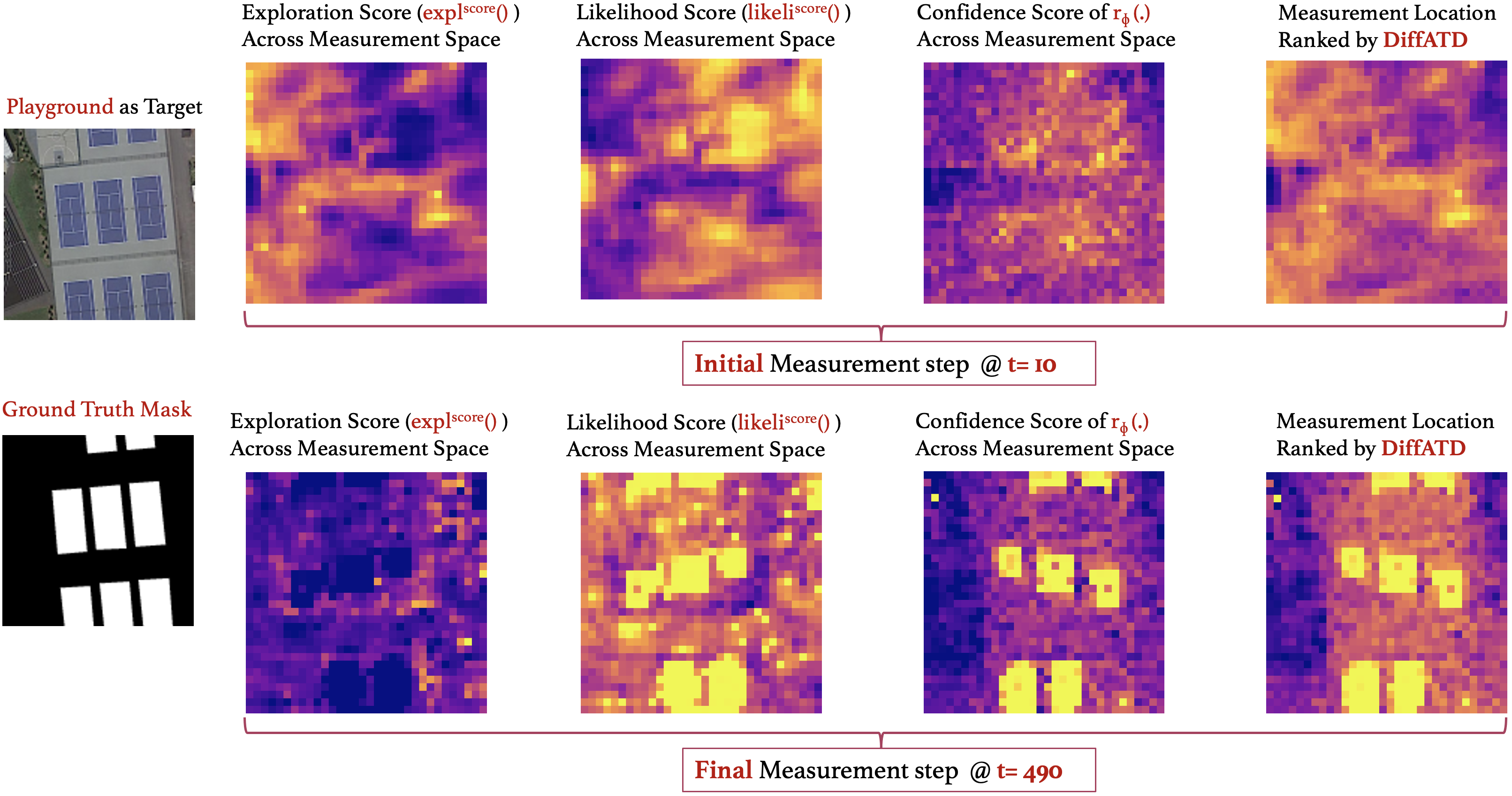}
    \caption{\small{Explanation of \emph{DiffATD}. Brighter indicates a higher value.}}
    \label{fig:explain}
    \vspace{-10pt}
\end{figure}
the exploration score when ranking measurement locations.
However, as the search nears its final stages, rankings are increasingly driven by the exploitation score, as defined in Eqn.~\ref{eq:exploit-score}. \textbf{As the search progresses, the confidence of $r_{\phi}$ and $\mathrm{likeli}^{\mathrm{score}}(.)$ become more accurate, making the $\mathrm{exploit}^{\mathrm{score}}(.)$ more reliable, which justifies \emph{DiffATD}'s increasing reliance on it in the later phase.} More such visualizations are presented in the Appendix.
\vspace{-4pt}

\paragraph{Comparison With Vision-Language Model}
We conduct comparisons between DiffATD and two State-of-the-art VLMs: GPT-4o and Gemini, on the DOTA dataset. The corresponding empirical results are summarized in the following Table~\ref{tab: VLM1}. We observe that DiffATD consistently outperforms VLM baselines across different measurement budgets. \textbf{These findings underscore DiffATD’s advantage over alternatives like VLMs in the ATD setting, driven by its principled balance of exploration and exploitation based on the maximum entropy framework.} Additionally, in Figure~\ref{fig:VLM1}, we present comparative exploration strategies for different targets from DOTA to provide further intuition. 

\begin{center}
\begin{minipage}{0.48\textwidth}
  \centering
  \includegraphics[width=\linewidth]{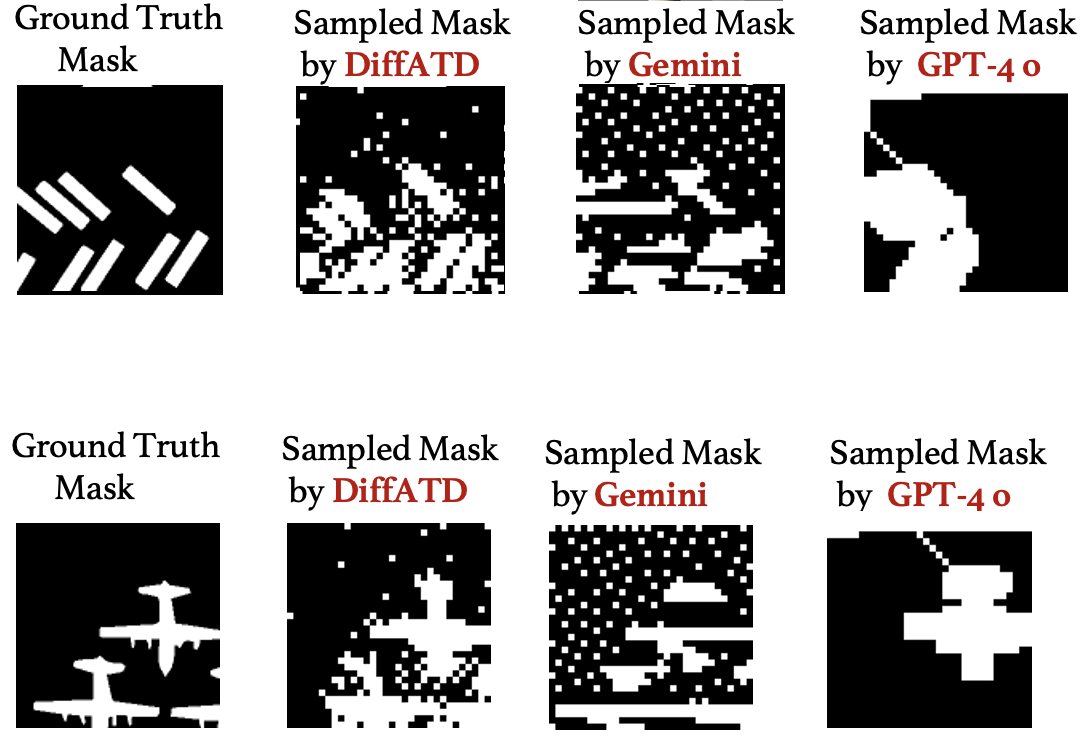}
  \captionof{figure}{\small{Active Discovery of Plane, Truck.}}
  \label{fig:VLM1}
\end{minipage}
\hfill
\begin{minipage}{0.50\textwidth}
  \centering
  \captionof{table}{Quantitative Comparison with VLM using DOTA Overhead Imagery Dataset}
  \vspace{-2pt}
  \begin{tabular}{p{1.33cm}p{1.33cm}p{1.33cm}p{1.33cm}}
    \toprule
    \multicolumn{4}{c}{Active Discovery with targets, e.g., plane, truck} \\
    \midrule
    Method & $\mathcal{B}=150$ & $\mathcal{B}=180$ & $\mathcal{B}=300$ \\
    \midrule
    GPT4-o &  0.3383 &   0.3949 & 0.5678 \\ [0.5em]
    Gemini & 0.4027  &   0.4608   & 0.6453 \\ [0.5em]
    \hline 
    \textbf{\emph{DiffATD}} & \textbf{0.4537} & \textbf{0.5092} & \textbf{0.7309} \\ 
    \bottomrule
  \end{tabular}
  \label{tab: VLM1}
\end{minipage}
\end{center}

\section{Conclusions}
\vspace{-7pt}
We present Diffusion-guided Active Target Discovery (\emph{DiffATD}), a novel approach that harnesses diffusion dynamics for efficient target discovery in partially observable environments within a fixed sampling budget. \emph{DiffATD} eliminates the need for supervised training, addressing a key limitation of prior approaches. Additionally, \emph{DiffATD} offers interpretability, in contrast to existing black-box policies. Our extensive experiments across diverse domains demonstrate its efficacy. We believe our work will inspire further exploration of this impactful problem and encourage the adoption of \emph{DiffATD} in real-world applications across diverse fields.

\paragraph{Acknowledgement} This research was partially supported by the National Science Foundation (CCF-2403758, IIS-2214141), Army Research Office (W911NF2510059), Office of Naval Research (N000142412663), Amazon.
\bibliographystyle{unsrt}
\bibliography{reference}

\begin{thebibliography}{10}

\bibitem{jayaraman2016look}
Dinesh Jayaraman and Kristen Grauman.
\newblock Look-ahead before you leap: end-to-end active recognition by forecasting the effect of motion.
\newblock In {\em Computer Vision--ECCV 2016: 14th European Conference, Amsterdam, The Netherlands, October 11-14, 2016, Proceedings, Part V 14}, pages 489--505. Springer, 2016.

\bibitem{jayaraman2018learning}
Dinesh Jayaraman and Kristen Grauman.
\newblock Learning to look around: Intelligently exploring unseen environments for unknown tasks.
\newblock In {\em Proceedings of the IEEE conference on computer vision and pattern recognition}, pages 1238--1247, 2018.

\bibitem{xiong2018snap}
Bo~Xiong and Kristen Grauman.
\newblock Snap angle prediction for 360 panoramas.
\newblock In {\em Proceedings of the European Conference on Computer Vision (ECCV)}, pages 3--18, 2018.

\bibitem{huijben2020deep}
Iris Huijben, Bastiaan~S Veeling, and Ruud~JG van Sloun.
\newblock Deep probabilistic subsampling for task-adaptive compressed sensing.
\newblock In {\em 8th International Conference on Learning Representations, ICLR 2020}, 2020.

\bibitem{bahadir2020deep}
Cagla~D Bahadir, Alan~Q Wang, Adrian~V Dalca, and Mert~R Sabuncu.
\newblock Deep-learning-based optimization of the under-sampling pattern in mri.
\newblock {\em IEEE Transactions on Computational Imaging}, 6:1139--1152, 2020.

\bibitem{van2021active}
Hans Van~Gorp, Iris Huijben, Bastiaan~S Veeling, Nicola Pezzotti, and Ruud~JG Van~Sloun.
\newblock Active deep probabilistic subsampling.
\newblock In {\em International Conference on Machine Learning}, pages 10509--10518. PMLR, 2021.

\bibitem{bakker2020experimental}
Tim Bakker, Herke van Hoof, and Max Welling.
\newblock Experimental design for mri by greedy policy search.
\newblock {\em Advances in Neural Information Processing Systems}, 33:18954--18966, 2020.

\bibitem{yin2021end}
Tianwei Yin, Zihui Wu, He~Sun, Adrian~V Dalca, Yisong Yue, and Katherine~L Bouman.
\newblock End-to-end sequential sampling and reconstruction for mri.
\newblock {\em arXiv preprint arXiv:2105.06460}, 2021.

\bibitem{stevens2022accelerated}
Tristan~SW Stevens, Nishith Chennakeshava, Frederik~J de~Bruijn, Martin Peka{\v{r}}, and Ruud~JG van Sloun.
\newblock Accelerated intravascular ultrasound imaging using deep reinforcement learning.
\newblock In {\em ICASSP 2022-2022 IEEE International Conference on Acoustics, Speech and Signal Processing (ICASSP)}, pages 1216--1220. IEEE, 2022.

\bibitem{sanchez2020closed}
Thomas Sanchez, Igor Krawczuk, Zhaodong Sun, and Volkan Cevher.
\newblock Closed loop deep bayesian inversion: Uncertainty driven acquisition for fast mri.
\newblock 2020.

\bibitem{nolan2024active}
Oisin Nolan, Tristan~SW Stevens, Wessel~L van Nierop, and Ruud~JG van Sloun.
\newblock Active diffusion subsampling.
\newblock {\em arXiv preprint arXiv:2406.14388}, 2024.

\bibitem{uzkent2020learning}
Burak Uzkent and Stefano Ermon.
\newblock Learning when and where to zoom with deep reinforcement learning.
\newblock In {\em Proceedings of the IEEE/CVF conference on computer vision and pattern recognition}, pages 12345--12354, 2020.

\bibitem{sarkar2024visual}
Anindya Sarkar, Michael Lanier, Scott Alfeld, Jiarui Feng, Roman Garnett, Nathan Jacobs, and Yevgeniy Vorobeychik.
\newblock A visual active search framework for geospatial exploration.
\newblock In {\em Proceedings of the IEEE/CVF Winter Conference on Applications of Computer Vision}, pages 8316--8325, 2024.

\bibitem{sarkar2023partially}
Anindya Sarkar, Nathan Jacobs, and Yevgeniy Vorobeychik.
\newblock A partially-supervised reinforcement learning framework for visual active search.
\newblock {\em Advances in Neural Information Processing Systems}, 36:12245--12270, 2023.

\bibitem{nguyen2024amortized}
Quan Nguyen, Anindya Sarkar, and Roman Garnett.
\newblock Amortized nonmyopic active search via deep imitation learning.
\newblock {\em arXiv preprint arXiv:2405.15031}, 2024.

\bibitem{garnett2012bayesian}
Roman Garnett, Yamuna Krishnamurthy, Xuehan Xiong, Jeff Schneider, and Richard Mann.
\newblock Bayesian optimal active search and surveying.
\newblock {\em arXiv preprint arXiv:1206.6406}, 2012.

\bibitem{jiang2017efficient}
Shali Jiang, Gustavo Malkomes, Geoff Converse, Alyssa Shofner, Benjamin Moseley, and Roman Garnett.
\newblock Efficient nonmyopic active search.
\newblock In {\em International Conference on Machine Learning}, pages 1714--1723. PMLR, 2017.

\bibitem{jiang2019cost}
Shali Jiang, Roman Garnett, and Benjamin Moseley.
\newblock Cost effective active search.
\newblock {\em Advances in Neural Information Processing Systems}, 32, 2019.

\bibitem{rangrej2022consistency}
Samrudhdhi~B Rangrej, Chetan~L Srinidhi, and James~J Clark.
\newblock Consistency driven sequential transformers attention model for partially observable scenes.
\newblock {\em arXiv preprint arXiv:2204.00656}, 2022.

\bibitem{pirinen2022aerial}
Aleksis Pirinen, Anton Samuelsson, John Backsund, and Kalle Astr{\"o}m.
\newblock Aerial view goal localization with reinforcement learning.
\newblock {\em arXiv preprint arXiv:2209.03694}, 2022.

\bibitem{sarkar2024gomaa}
Anindya Sarkar, Srikumar Sastry, Aleksis Pirinen, Chongjie Zhang, Nathan Jacobs, and Yevgeniy Vorobeychik.
\newblock Gomaa-geo: Goal modality agnostic active geo-localization.
\newblock {\em arXiv preprint arXiv:2406.01917}, 2024.

\bibitem{song2020score}
Yang Song, Jascha Sohl-Dickstein, Diederik~P Kingma, Abhishek Kumar, Stefano Ermon, and Ben Poole.
\newblock Score-based generative modeling through stochastic differential equations.
\newblock {\em arXiv preprint arXiv:2011.13456}, 2020.

\bibitem{anderson1982reverse}
Brian~DO Anderson.
\newblock Reverse-time diffusion equation models.
\newblock {\em Stochastic Processes and their Applications}, 12(3):313--326, 1982.

\bibitem{ho2020denoising}
Jonathan Ho, Ajay Jain, and Pieter Abbeel.
\newblock Denoising diffusion probabilistic models.
\newblock {\em Advances in neural information processing systems}, 33:6840--6851, 2020.

\bibitem{chung2022diffusion}
Hyungjin Chung, Jeongsol Kim, Michael~T Mccann, Marc~L Klasky, and Jong~Chul Ye.
\newblock Diffusion posterior sampling for general noisy inverse problems.
\newblock {\em arXiv preprint arXiv:2209.14687}, 2022.

\bibitem{song2023pseudoinverse}
Jiaming Song, Arash Vahdat, Morteza Mardani, and Jan Kautz.
\newblock Pseudoinverse-guided diffusion models for inverse problems.
\newblock In {\em International Conference on Learning Representations}, 2023.

\bibitem{rout2024solving}
Litu Rout, Negin Raoof, Giannis Daras, Constantine Caramanis, Alex Dimakis, and Sanjay Shakkottai.
\newblock Solving linear inverse problems provably via posterior sampling with latent diffusion models.
\newblock {\em Advances in Neural Information Processing Systems}, 36, 2024.

\bibitem{hershey2007approximating}
John~R Hershey and Peder~A Olsen.
\newblock Approximating the kullback leibler divergence between gaussian mixture models.
\newblock In {\em 2007 IEEE International Conference on Acoustics, Speech and Signal Processing-ICASSP'07}, volume~4, pages IV--317. IEEE, 2007.

\bibitem{xia2018dota}
Gui-Song Xia, Xiang Bai, Jian Ding, Zhen Zhu, Serge Belongie, Jiebo Luo, Mihai Datcu, Marcello Pelillo, and Liangpei Zhang.
\newblock Dota: A large-scale dataset for object detection in aerial images.
\newblock In {\em Proceedings of the IEEE conference on computer vision and pattern recognition}, pages 3974--3983, 2018.

\bibitem{inaturalist}
{\em iNaturalist}.
\newblock https://www.inaturalist.org/.

\bibitem{rotemberg2021patient}
Veronica Rotemberg, Nicholas Kurtansky, Brigid Betz-Stablein, Liam Caffery, Emmanouil Chousakos, Noel Codella, Marc Combalia, Stephen Dusza, Pascale Guitera, David Gutman, et~al.
\newblock A patient-centric dataset of images and metadata for identifying melanomas using clinical context.
\newblock {\em Scientific data}, 8(1):34, 2021.

\bibitem{van2006segmentation}
Bram Van~Ginneken, Mikkel~B Stegmann, and Marco Loog.
\newblock Segmentation of anatomical structures in chest radiographs using supervised methods: a comparative study on a public database.
\newblock {\em Medical image analysis}, 10(1):19--40, 2006.

\bibitem{kirillov2023segment}
Alexander Kirillov, Eric Mintun, Nikhila Ravi, Hanzi Mao, Chloe Rolland, Laura Gustafson, Tete Xiao, Spencer Whitehead, Alexander~C Berg, Wan-Yen Lo, et~al.
\newblock Segment anything.
\newblock In {\em Proceedings of the IEEE/CVF International Conference on Computer Vision}, pages 4015--4026, 2023.

\bibitem{lee2020maskgan}
Cheng-Han Lee, Ziwei Liu, Lingyun Wu, and Ping Luo.
\newblock Maskgan: Towards diverse and interactive facial image manipulation.
\newblock In {\em Proceedings of the IEEE/CVF conference on computer vision and pattern recognition}, pages 5549--5558, 2020.

\bibitem{song2020denoising}
Jiaming Song, Chenlin Meng, and Stefano Ermon.
\newblock Denoising diffusion implicit models.
\newblock {\em arXiv preprint arXiv:2010.02502}, 2020.

\end{thebibliography}

\newpage

\newpage
\clearpage
\newpage

\clearpage
\appendix
\appendix

\newcommand{\ldotsfill}{\leavevmode\leaders\hbox to .5em{\hss.\hss}\hfill\kern0pt}

\section*{Appendix: Online Feedback Efficient Active Target Discovery in Partially Observable Environments}
\vspace{3pt}

\section*{Overview of the Contents}

\noindent
\begin{tabularx}{\textwidth}{@{}Xr@{}}
\textbf{A:} \textbf{Proofs of Theoretical Results}  \ldotsfill & 22 \\ \\
\quad A.1 Proof of Proposition 1 \ldotsfill & 22 \\ [0.5em]
\quad A.2 Proof of Theorem 1 \ldotsfill & 22-23 \\ [0.5em]
\quad A.3 Proof of Theorem 2 \ldotsfill & 23-24 \\[0.5em]
\quad A.4 Proof of Theorem 3 \ldotsfill & 23 \\ \\ [0.5em] 

\textbf{B: Empirical Analysis of Active Target Discovery across Diverse Domains} \ldotsfill & 24 \\ \\
\quad B.1 Active Target Discovery of Different Parts of Human Face \ldotsfill & 24 \\ [0.5em]
\quad B.3 Active Target Discovery of Handwritten Digits \ldotsfill & 25 \\ \\ [0.5em]

\textbf{C: Qualitative Comparisons with Greedy Adaptive Approach} \ldotsfill & 25 \\ \\
\quad C.1 Qualitative Comparisons using Lung Disease, Skin Disease, DOTA, CelebA Images \ldotsfill & 25-27 \\ \\ [0.5em]

\textbf{D: Additional Visualization of Exploration Strategy of DiffATD} \ldotsfill & 27 \\ \\
\quad D.1  Comparative Visualizations using Lung Disease, Skin Disease, DOTA, CelebA Images \ldotsfill & 27-29 \\ \\ [0.5em]

\textbf{E: Details of Training and Inference} \ldotsfill & 30 \\ \\
\quad E.1  Details of Computing Resources, Training, and Inference Hyperparameters \ldotsfill & 30 \\  [0.5em]
\quad E.2  Details of Reward Model $r_{\phi}$ \ldotsfill & 30 \\ \\ [0.5em]

\textbf{F: Challenges in Active Target Discovery for Rare Categories} \ldotsfill & 30 -31 \\ [0.5em]

\textbf{G: Scalability of DiffATD on Larger Search Spaces} \ldotsfill & 31 \\ [0.5em]

\textbf{H: Additional Results to Access the Effect of $\kappa(\beta)$} \ldotsfill & 31 \\ [0.5em]

\textbf{I: Rationale Behind Uniform Measurement Schedule Over the Number of Reverse Diffusion Steps} \ldotsfill & 31-32 \\ [0.5em]

\textbf{J: Performance of DiffATD Under Noisy Observations} \ldotsfill & 32 \\ [0.5em]

\textbf{K: Species Distribution Modelling as Active Target Discovery Problem} \ldotsfill & 33 \\ [0.5em]

\textbf{L: Statitical Significance Results of DiffATD} \ldotsfill & 33 \\ [0.5em]

\textbf{M: Comparison With Multi-Armed Bandit Based Method} \ldotsfill & 33-34 \\ [0.5em]

\textbf{N: Segmentation and Active Target Discovery are Fundamentally Different
Task} \ldotsfill & 34 \\ [0.5em]

\textbf{O: More Details on Computational Cost across Search Space} \ldotsfill & 34 \\ [0.5em]

\textbf{P: More Visualizations on DiffATD’s Exploration vs Exploitation strategy at
Different Stage} \ldotsfill & 34-37 \\  [0.5em]

\textbf{Q: Importance of Utilizing a Strong Prior in Tackling Active Target Discovery} \ldotsfill & 37 \\ [0.5em]

\textbf{R: Limitations and Future Work} \ldotsfill & 37 \\ [0.5em]

\textbf{S: Notation Table} \ldotsfill & 37-38 \\ [0.5em]


\end{tabularx}

\newpage

\section{ Proof of Theoretical Results } 
\subsection{Proof of Proposition 1}\label{pr:p1}
\begin{proof}
\[
q^{\text{exp}}_t = \arg\max_{q_t} \left[ I(\hat{x}_t; x \mid Q_t, \tilde{x}_{t-1}) \right]
\]
\[
= \arg\max_{q_t} \left[ \mathbb{E}_{p(\hat{x}_t \mid x, Q_t)p(x \mid \tilde{x}_{t-1})} \left[ \log p(\hat{x}_t \mid Q_t, \tilde{x}_{t-1}) - \log p(\hat{x}_t \mid x, Q_t, \tilde{x}_{t-1}) \right] \right]
\]
\[
= \arg\max_{q_t} \left[ H(\hat{x}_t \mid Q_t, \tilde{x}_{t-1}) - \underbrace{H(\hat{x}_t \mid x, Q_t, \tilde{x}_{t-1})}_{\textit{remains independent of the $q_t$. Therefore, this term can be disregarded when optimizing for $q_t$}} \right].
\]
\[
= \arg\max_{q_t} \left[ H(\hat{x}_t \mid Q_t, \tilde{x}_{t-1}) \right]
\]
\end{proof}

\subsection{Proof of Theorem 1}\label{pr:p2}
\begin{proof}
We demonstrate that maximizing the marginal entropy of the belief distribution as defined in Equation~\ref{eq:mar_ent} does not require computing a separate set of particles for every possible choice of measurement location $q_t$ when the action corresponds to selecting a measurement location. Since the measurement locations $Q_t = Q_{t-1} \cup q_t$ only differ in the newly selected indices $q_t$ within the \text{arg max},  the elements of each particle $x^{(i)}_t$ remain unchanged across all possible $Q_t$,  except at the indices specified by $q_t$. Exploiting this structure, we decompose the squared $L_2$ norm into two parts: one over the indices in $q_t$ and the other over those in $Q_{t-1}$. The term associated with $Q_{t-1}$ becomes a constant in the \text{arg max} and can be disregarded. Consequently, the formulation reduces the computing of the squared $L_2$ norms exclusively for the elements corresponding to $q_t$. We denote $k$ as the set of possible measurement locations at step $t$. Utilizing Equation~\ref{eq:mar_ent}, we can write,

\[
q^{\text{exp}}_t = \arg\max_{q_t} \sum^{N_B}_{i=0} \alpha_i \text{log} \sum^{N_B}_{j=0} \alpha_j \text{exp} \left\{ {\frac{||\hat{x}^{(i)}_t - \hat{x}^{(j)}_t||^2_2}{2\sigma^2_x}} \right\}
\]

\[
q^{\text{exp}}_t = \arg\max_{q_t} \sum_{i,j} \log \left( \exp \left( \frac{\| \hat{x}_t^{(i)} - \hat{x}_t^{(j)} \|^2}{2\sigma_x^2} \right) \right)  (\text{By assuming}, \alpha_i = \alpha_j, \, \forall i, j )
\]

\[
q^{\text{exp}}_t = \arg\max_{q_t} \sum_{i,j} \log \left( \exp \left( \frac{\sum_{a \in Q_t} ([\hat{x}^{(i)}_t]_a - [\hat{x}^{(j)}_t]_a)^2}{2\sigma_x^2} \right) \right)
\]

\[
q^{\text{exp}}_t = \arg\max_{q_t} \sum_{i,j} \log \left( \exp \left( \frac{\sum_{q_t \in k} ([\hat{x}^{(i)}_t]_{q_t} - [\hat{x}^{(j)}_t]_{q_t})^2 + \sum_{r \in Q_{t-1}} ([\hat{x}^{(i)}_t]_r - [\hat{x}^{(j)}_t]_r)^2}{2\sigma_x^2} \right) \right)
\]

\[
q^{\text{exp}}_t = \arg\max_{q_t} \sum_{i,j} \log \left( \prod_{q_t \in k} \exp \left( \frac{([\hat{x}^{(i)}_t]_{q_t} - [\hat{x}^{(j)}_t]_{q_t})^2}{2\sigma_x^2} \right) \prod_{r \in Q_{t-1}} \exp \left( \frac{([\hat{x}^{(i)}_t]_r - [\hat{x}^{(j)}_t]_r)^2}{2\sigma_x^2} \right) \right)
\]

\[
q^{\text{exp}}_t \propto \arg\max_{q_t} \sum_{i,j} \left( \sum_{q_t \in k} \frac{([\hat{x}^{(i)}_t]_{q_t} - [\hat{x}^{(j)}_t]_{q_t})^2}{2\sigma_x^2} + \underbrace{\sum_{r \in Q_{t-1}} \frac{([\hat{x}^{(i)}_t]_r - [\hat{x}^{(j)}_t]_r)^2}{2\sigma_x^2}}_\textit{We can ignore as it doesn't depend on the choice of $q_t$.} \right)
\]

\[
q^{\text{exp}}_t \propto \arg\max_{q_t} \sum_{i,j} \sum_{q_t \in k} \frac{([\hat{x}^{(i)}_t]_{q_t} - [\hat{x}^{(j)}_t]_{q_t})^2}{2\sigma_x^2}.
\]

\[
q^{\text{exp}}_t = \arg\max_{q_t} \left[ \sum_{i=0}^{N_B} \log \sum_{j=0}^{N_B} \exp \left( \frac{ \sum_{q_t \in k}^{ }([\hat{x}^{(i)}_t]_{q_t} - [\hat{x}^{(j)}_t]_{q_t})^2 }{2\sigma_x^2} \right) \right]
\]
\end{proof}

\subsection{Proof of Theorem 3}\label{pr:p4}
A pure greedy strategy selects the measurement location that maximizes the expected reward at the current time step, as defined below:

\[
= \arg\max_{q_t} \mathbb{E}_{\hat{x}_t \sim p(\hat{x}_t | Q_t, \tilde{x}_{t-1})} [r_{\phi}[\hat{x}_t]_{q_t}]
\]

\[
= \arg\max_{q_t} \underbrace{\int p(\hat{x}_t | Q_t, \tilde{x}_{t-1}) r_{\phi}[\hat{x}_t]_{q_t} d\hat{x}_t}_\textit{J($q_t$)}
\]

In order to maximize J($q_t$) w.r.t $q_t$, we do the following:
\[
\nabla_{q_t} J(q_t) = \arg\max_{q_t} \int \nabla_{q_t} p(\hat{x}_t | Q_t, \tilde{x}_{t-1}) r_{\phi}[\hat{x}_t]_{q_t} d\hat{x}_t
\]

\[
= \arg\max_{q_t} \int p(\hat{x}_t | Q_t, \tilde{x}_{t-1}) \nabla_{q_t} \log p(\hat{x}_t | Q_t, \tilde{x}_{t-1}) r_{\phi}[\hat{x}_t]_{q_t} d\hat{x}_t
\]
We obtain the last equality using the following identity:
\[
p_{\theta}(x) \nabla_{\theta} \log p_{\theta}(x) = \nabla_{\theta} p_{\theta}(x)
\]
We simplify the expression using the definition of expectation, as shown below:
\[
= \arg\max_{q_t} \mathbb{E}_{\hat{x}_t \sim p(\hat{x}_t | Q_t, \tilde{x}_{t-1})} [\underbrace{\nabla_{q_t} \log p(\hat{x}_t | Q_t, \tilde{x}_{t-1})}_\textit{Maximizing expected log-likelihood score, i.e., $\mathrm{likeli}^{\mathrm{score}}(.)$.} \underbrace{r_{\phi}[\hat{x}_t]_{q_t}}_\textit{  Reward at current time.}]
\]

By definition, it is equivalent to the following
\[
= \arg\max_{q_t} \left[ \mathrm{exploit}^{\mathrm{score}}(q_t) \right] \>\>\>\>\>\> \text{(as defined in Equation~\ref{eq:exploit-score}.)}
\]


\subsection{Proof of Theorem 2}\label{pr:p3}
\begin{proof}
We start with the definition of entropy $H$:
\[
\mathbb{E}_{\hat{x}_t}[\log p(\hat{x}_t | Q_t, \tilde{x}_{t-1})]=-H(\hat{x}_t | Q_t, \tilde{x}_{t-1})
\]
Substituting the expression of $H(\hat{x}_t | Q_t, \tilde{x}_{t-1})$ as defined in Equation~\ref{eq:mar_ent}, and by setting \( \alpha_i = \alpha_j = 1 \), we obtain:
\[
\mathbb{E}_{\hat{x}_t}[\log p(\hat{x}_t | Q_t, \tilde{x}_{t-1})] \propto - \sum^{N_B}_{i=0} \log \sum^{N_B}_{j=0} \exp \left\{ {\frac{||\hat{x}^{(i)}_t - \hat{x}^{(j)}_t||^2_2}{2\sigma^2_x}} \right\}
\]

\[
\mathbb{E}_{\hat{x}_t}[\log p(\hat{x}_t | Q_t, \tilde{x}_{t-1})] \propto -  \sum_{i,j} \log \left( \exp \left( \frac{\sum_{a \in Q_t} ([\hat{x}^{(i)}_t]_a - [\hat{x}^{(j)}_t]_a)^2}{2\sigma_x^2} \right) \right)
\]

Assuming $k$ is the set of potential measurement locations at time step t, and $Q_t = Q_{t-1} \cup q_t$, where $q_t \in k$.
\[
\mathbb{E}_{\hat{x}_t}[\log p(\hat{x}_t | Q_t, \tilde{x}_{t-1})] \propto - \sum_{i,j} \log \left( \exp \left( \frac{\sum_{q_t \in k} ([\hat{x}^{(i)}_t]_{q_t} - [\hat{x}^{(j)}_t]_{q_t})^2 + \sum_{r \in Q_{t-1}} ([\hat{x}^{(i)}_t]_r - [\hat{x}^{(j)}_t]_r)^2}{2\sigma_x^2} \right) \right)
\]

\[
\mathbb{E}_{\hat{x}_t}[\log p(\hat{x}_t | Q_t, \tilde{x}_{t-1})] \propto - \sum_{i,j} \log \left( \prod_{q_t \in k} \exp \left( \frac{([\hat{x}^{(i)}_t]_{q_t} - [\hat{x}^{(j)}_t]_{q_t})^2}{2\sigma_x^2} \right) \prod_{r \in Q_{t-1}} \exp \left( \frac{([\hat{x}^{(i)}_t]_r - [\hat{x}^{(j)}_t]_r)^2}{2\sigma_x^2} \right) \right)
\]

\[
\mathbb{E}_{\hat{x}_t}[\log p(\hat{x}_t | Q_t, \tilde{x}_{t-1})] \propto - \sum_{i,j} \left( \sum_{q_t \in k} \frac{([\hat{x}^{(i)}_t]_{q_t} - [\hat{x}^{(j)}_t]_{q_t})^2}{2\sigma_x^2} + \sum_{r \in Q_{t-1}} \frac{([\hat{x}^{(i)}_t]_r - [\hat{x}^{(j)}_t]_r)^2}{2\sigma_x^2} \right)
\]

Next, we compute the expected log-likelihood at a specified measurement location $q_t$. Accordingly, we ignore all the terms that do not depend on $q_t$. This simple observation helps us to simplify the above expression as follows:

\[
\underbrace{\mathbb{E}_{\hat{x}_t}[\log p(\hat{x}_t | Q_t, \tilde{x}_{t-1})] \biggr\rvert_{q_t}}_\textit{The expected log-likelihood at a measurement location $q_t$} \propto  \sum_{i,j} \left( -\frac{([\hat{x}^{(i)}_t]_{q_t} - [\hat{x}^{(j)}_t]_{q_t})^2}{2\sigma_x^2} \right)
\]

Equivalently, we can write the above expression as:

\[
\mathbb{E}_{\hat{x}_t}[\log p(\hat{x}_t | Q_t, \tilde{x}_{t-1})] \biggr\rvert_{q_t} \propto  \left( \underbrace{\sum_{i=0}^{N_B} \sum_{j=0}^{N_B} \exp \left\{ - \frac{([\hat{x}^{(i)}_t]_{q_t} - [\hat{x}^{(j)}_t]_{q_t})^2}{2\sigma_x^2} \right\}}_{\mathrm{likeli}^{\mathrm{score}}(q_t)} \right)
\]

By definition, the L.H.S of the above expression is the same as $\mathrm{likeli}^{\mathrm{score}}(q_t)$. 
\end{proof}

\section{ Empirical Analysis of Active Target Discovery across Diverse Domains}
\subsection{Active Discovery of Different Parts of Human Face}\label{app:face} 
We compare \emph{DiffATD} with the baselines with the human face as the target from the CelebA dataset~\cite{lee2020maskgan} and report the findings in Table~\ref{tab: celeba}.
These results reveal a similar trend to that observed with the previous datasets. We observe significant performance gains with \emph{DiffATD} over all baselines, with improvements of $~42.60\%$ to $~60.79\%$ across all measurement budgets, demonstrating its mastery in balancing exploration and exploitation for effective active target discovery. We present visualizations of \emph{DiffATD}'s exploration strategy in Fig.\ref{fig:vis3}, with additional visualizations are in appendix~\ref{sec:EXP_A}.
\vspace{-6pt}
\begin{center}
\begin{minipage}{0.48\textwidth}
  \centering
  \includegraphics[width=\linewidth]{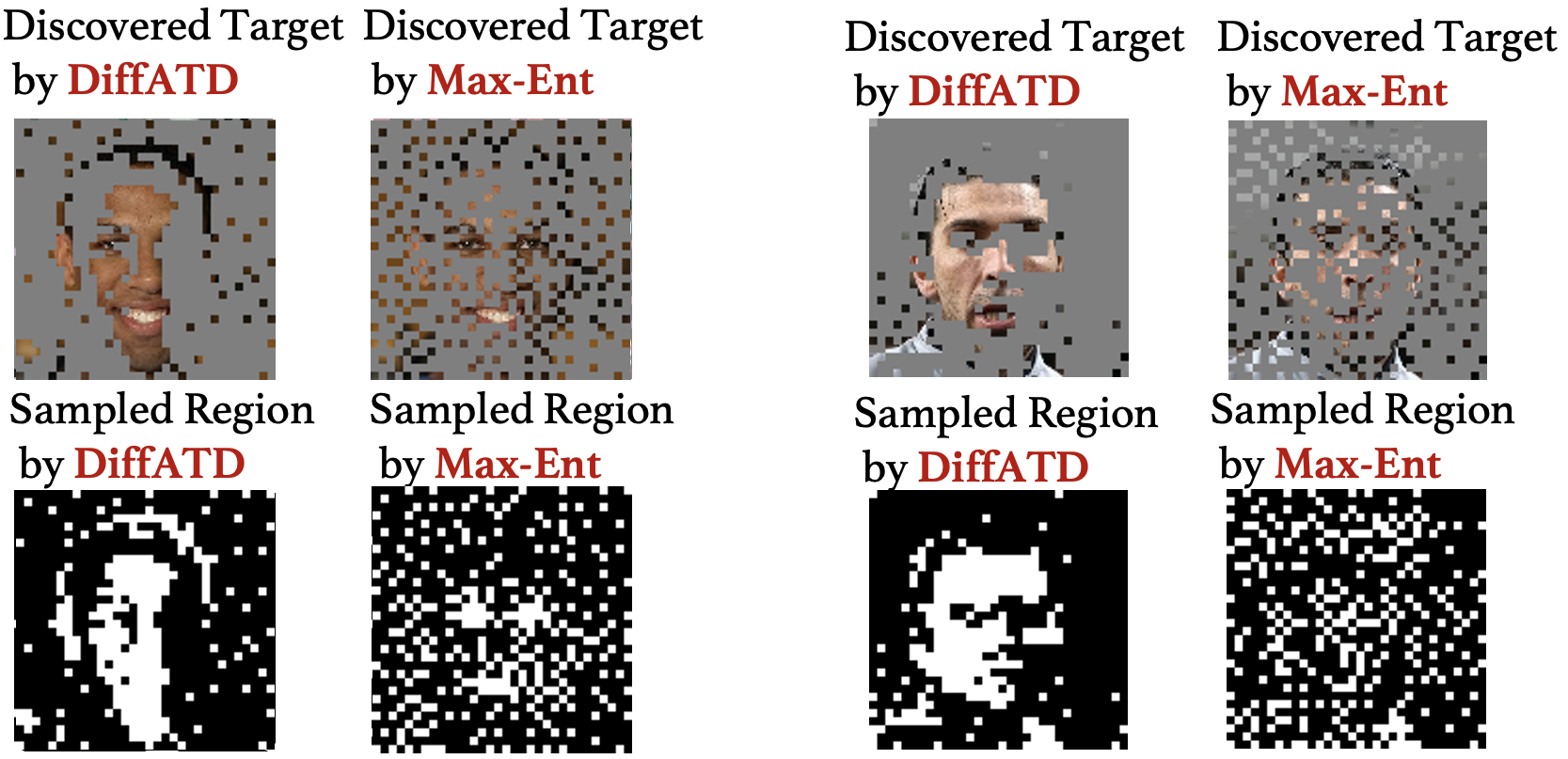}
  \captionof{figure}{\small{Active Discovery of regions with Face.}}
  \label{fig:vis3}
\end{minipage}
\hfill
\begin{minipage}{0.50\textwidth}
  \centering
  \footnotesize
  \captionof{table}{\emph{SR} Comparison with CelebA Dataset}
  \vspace{-2pt}
  \begin{tabular}{p{1.33cm}p{1.33cm}p{1.33cm}p{1.33cm}}
    \toprule
    \multicolumn{4}{c}{Active Discovery with face as the target} \\
    \midrule
    Method & $\mathcal{B}=200$ & $\mathcal{B}=250$ & $\mathcal{B}=300$ \\
    \midrule
    RS & 0.1938 & 0.2441 & 0.2953  \\
    Max-Ent & 0.2399 & 0.3510 & 0.4498 \\
    GA & 0.2839 & 0.3516 & 0.4294  \\
    \hline 
    \textbf{\emph{DiffATD}} & \textbf{0.4565} & \textbf{0.5646} & \textbf{0.6414}  \\ 
    \bottomrule
  \end{tabular}
  \label{tab: celeba}
\end{minipage}
\end{center}

\subsection{Active Discovery of Hand-written Digits}
\begin{figure}[!h] 
    \centering
    \includegraphics[width=0.88\textwidth]{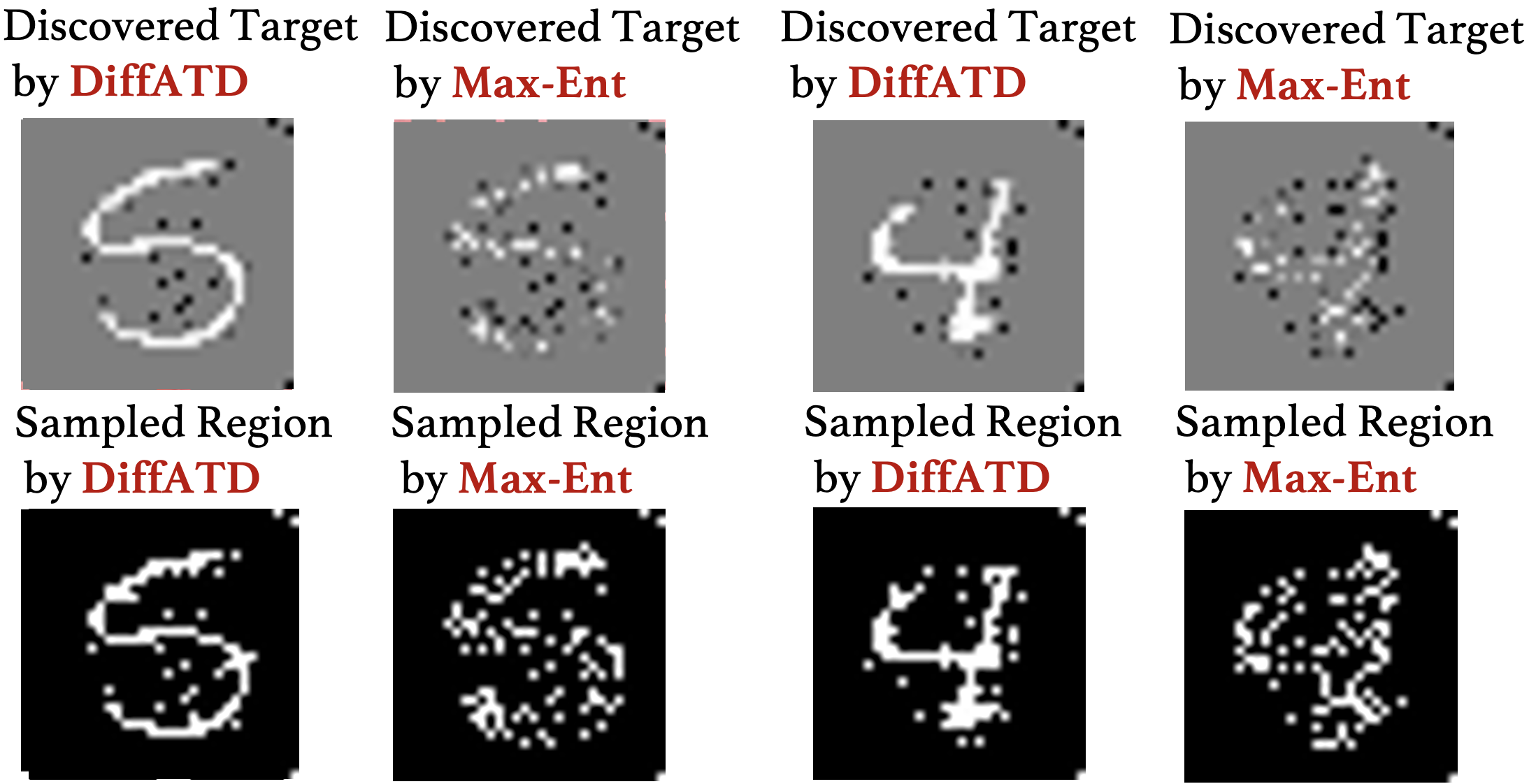}
    \caption{Visualization of Active Discovery of Handwritten Digits.}
    \label{fig:mnist11}
\end{figure}

Here, we evaluate performance by comparing \emph{DiffATD} with the baselines using the \emph{SR} metric. We compare the performance across varying measurement budgets $\mathcal{B}$. The results are presented in Table~\ref{tab: digit}. We observe significant improvements in the performance of the proposed \emph{DiffATD} approach compared to all baselines in each measurement budget setting, ranging from $~16.30\%$ to $~45.23\%$ improvement relative to the most competitive method. These empirical results are consistent with those from other datasets we explored, such as DOTA, and CelebA. We present a qualitative comparison of the exploration strategies of \emph{DiffATD} and Max-Ent through visual illustration. As shown in Fig.~\ref{fig:mnist11}, \emph{DiffATD} efficiently explores the search space, identifying the underlying true handwritten digit within the measurement budget, highlighting its effectiveness in balancing exploration and exploitation. We provide several such visualizations in the following section. 
\begin{table}[H]
    \centering
    \footnotesize
    \caption{\emph{SR} comparison with MNIST Dataset}
    \begin{tabular}{p{1.5cm}p{1.5cm}p{1.5cm}p{1.5cm}}
        \toprule
        \multicolumn{4}{c}{Active Discovery of Handwritten Digits} \\
        \midrule
        Method & $\mathcal{B}=100$ & $\mathcal{B}=150$ & $\mathcal{B}=200$  \\
        \midrule
        RS & 0.1270 & 0.1518 & 0.1943   \\
        Max-Ent & 0.5043 & 0.6285 & 0.8325  \\
        GA & 0.0922 & 0.4133 & 0.5820  \\
        \hline 
        \textbf{\emph{DiffATD}} & \textbf{0.7324} & \textbf{0.8447} & \textbf{0.9682}   \\ 
        \bottomrule
    \end{tabular}
    \label{tab: digit}
\end{table}

\section{ Qualitative Comparison with Greedy Adaptive Approach}\label{sec:GA_A}
\subsection{Qualitative Comparisons using Lung Disease, Skin Disease, DOTA, and CelebA Images}
\begin{figure}[!h] 
    \centering
    \includegraphics[width=0.92\textwidth]{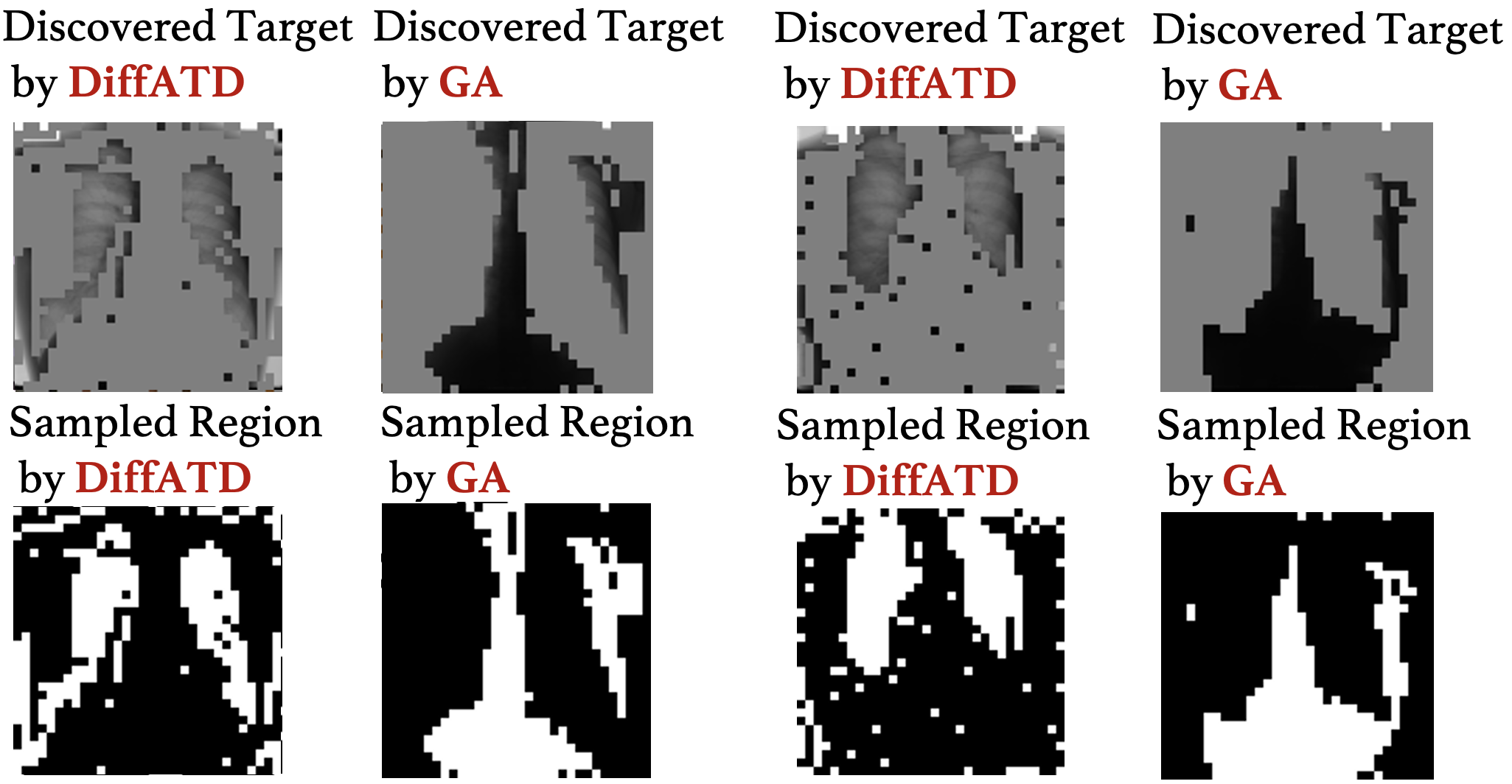}
    \caption{Visualization of Active Discovery of Lung Disease.}
    \label{fig:GA1}
\end{figure}

\begin{figure}[!h] 
    \centering
    \includegraphics[width=0.92\textwidth]{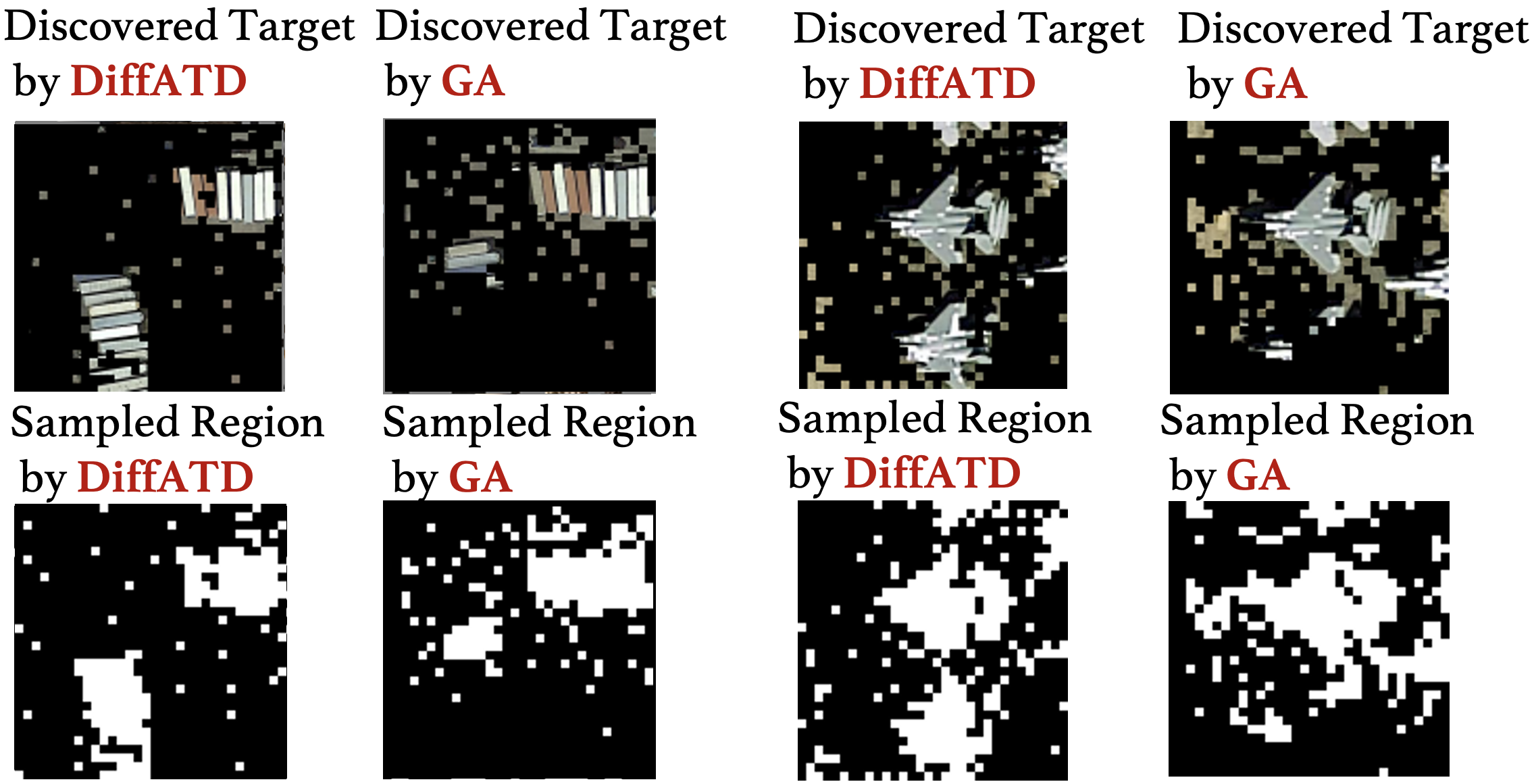}
    \caption{Visualization of Active Discovery of (left) Truck and (right) Plane}
    \label{fig:GA2}
\end{figure}

\begin{figure}[!h] 
    \centering
    \includegraphics[width=0.92\textwidth]{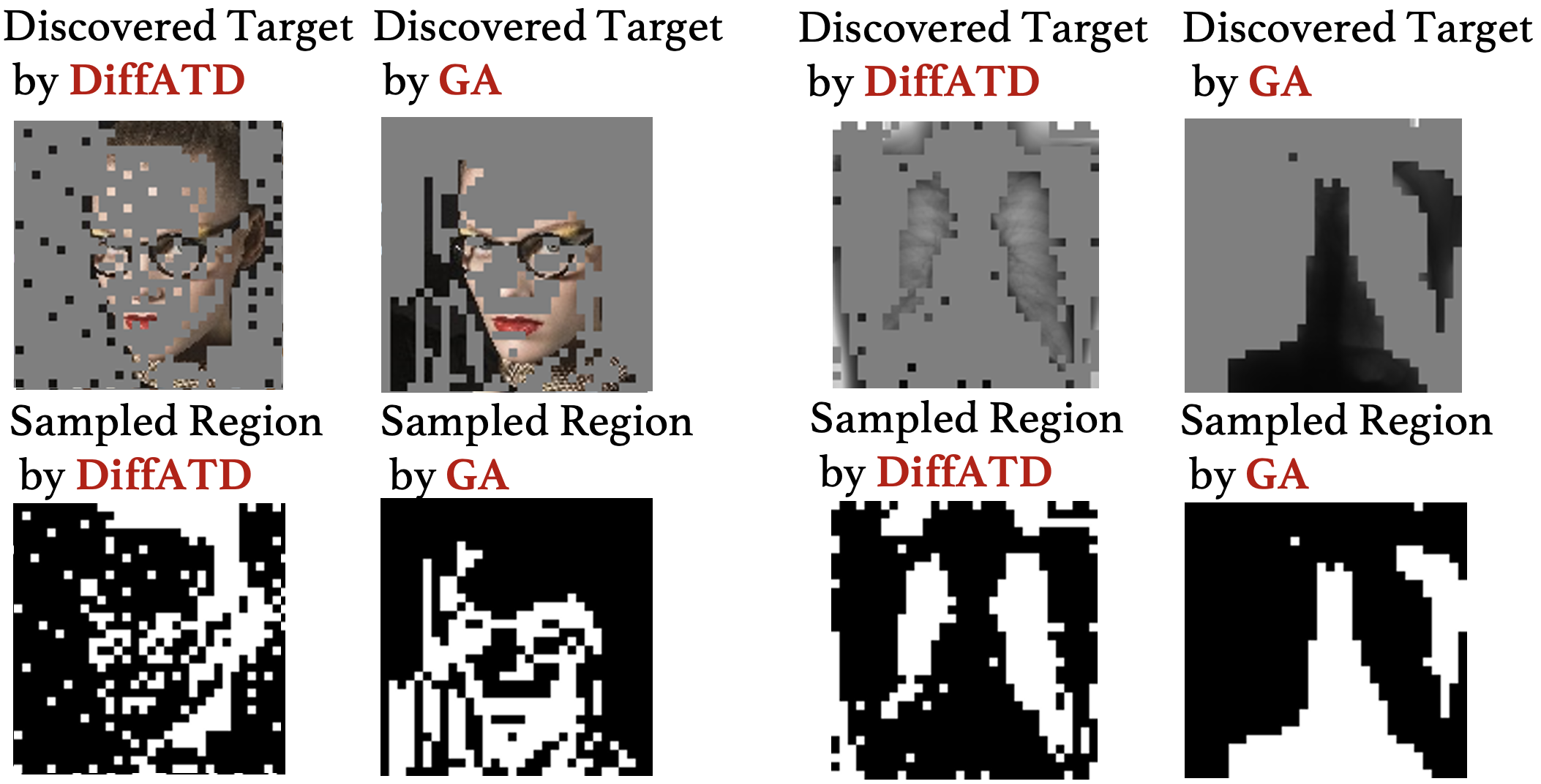}
    \caption{Visualization of Active Discovery of Eye, hair, nose, and lip.}
    \label{fig:GA3}
\end{figure}

\begin{figure}[!h] 
    \centering
    \includegraphics[width=0.92\textwidth]{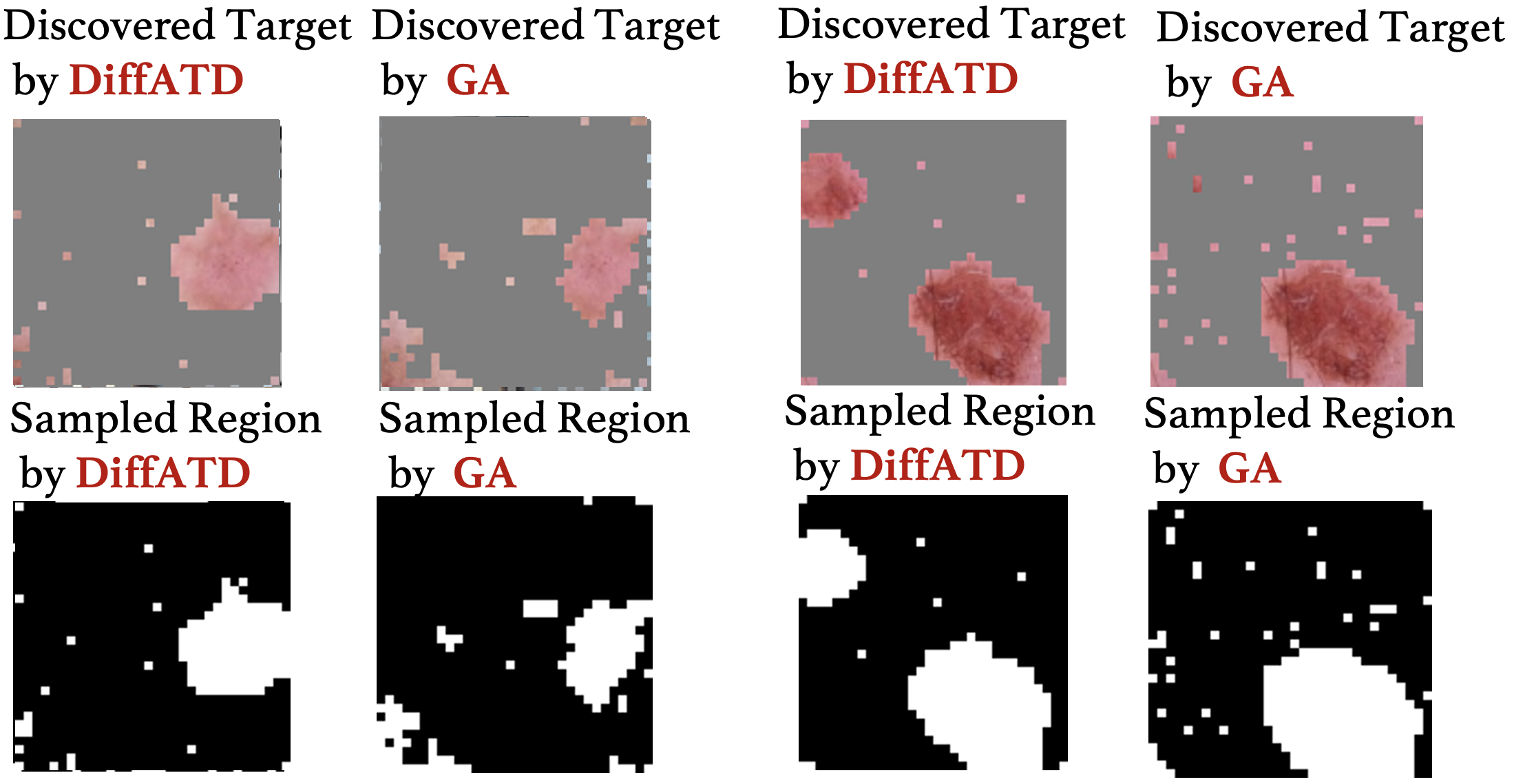}
    \caption{Visualization of Active Discovery of Skin Disease.}
    \label{fig:GA4}
\end{figure}

In this section, we present additional comparative visualizations of \emph{DiffATD}'s exploration strategy against \emph{Greedy-Adaptive} (GA), using samples from diverse datasets, including DOTA, CelebA, and Lung images. These visualizations are shown in Figures~\ref{fig:GA1},\ref{fig:GA2},\ref{fig:GA3},~\ref{fig:GA4}. In each case, \emph{DiffATD} strikes a strategic balance between exploration and exploitation, leading to a notably higher success rate and more efficient target discovery within a fixed budget, compared to the greedy strategy, which focuses solely on exploitation based on an incrementally learned reward model $r_{\phi}$. 
These qualitative observations emphasize key challenges in the active target discovery problem that are difficult to address with entirely greedy strategies like GA. For instance, the greedy approach struggles when the target has a complex and irregular structure with spatially isolated regions. Due to its nature, once it identifies a target in one region, it begins exploiting neighboring areas, guided by the reward model that assigns higher scores to similar regions. As a result, such methods struggle to discover spatially disjoint targets, as shown in these visualizations. 
Interestingly, the greedy approach also falters in noisy measurement environments due to its over-reliance on the reward model. Early in the search process, when training samples are limited, the model tends to overfit the noise, misguiding the discovery process. In contrast, our approach minimizes reliance on reward-based exploitation in the early stages. As more measurements are collected, and the reward model becomes incrementally more robust, our method begins to leverage reward-driven exploitation more effectively. This gradual adaptation makes our approach significantly more resilient and reliable, particularly in noisy environments, compared to the greedy strategy.

\vspace{62pt}

\vspace{62pt}

\vspace{52pt}

\vspace{2pt}


\section{ Additional Visualizations of the Exploration Strategy of \emph{DiffATD}}\label{sec:EXP_A}
\subsection{Comparative Visualizations using Lung Disease, Skin Disease, DOTA, CelebA Images}
\begin{figure}[!h] 
    \centering
    \includegraphics[width=0.88\textwidth]{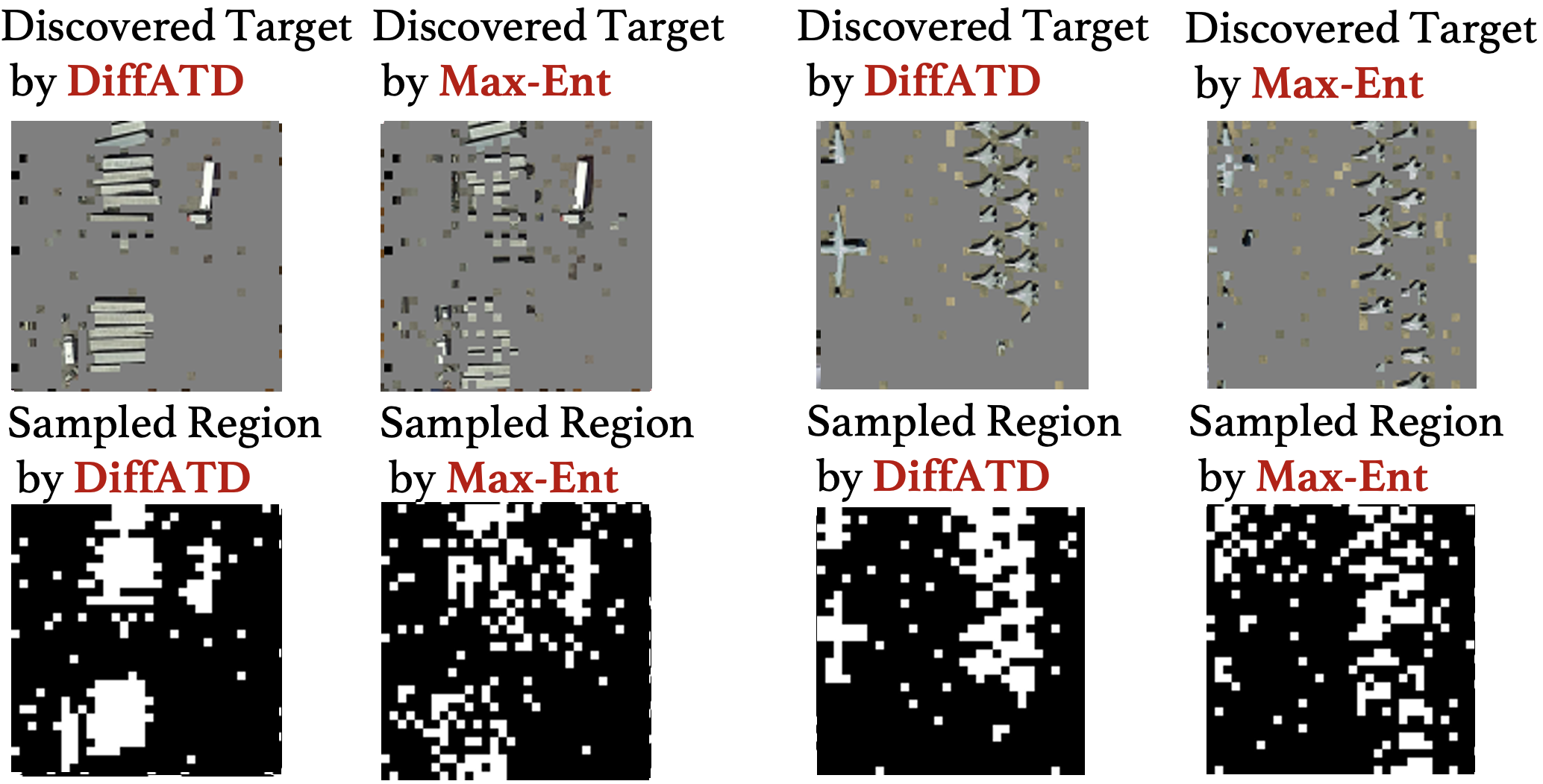}
    \vspace{9pt}
    \caption{\small{Visualizations of Active Discovery of (left) \emph{large-vehicle}, and (right) \emph{Plane}}.}
    \label{fig:explain_s_1}
\end{figure}

\begin{figure}[!h] 
    \centering
    \includegraphics[width=0.88\textwidth]{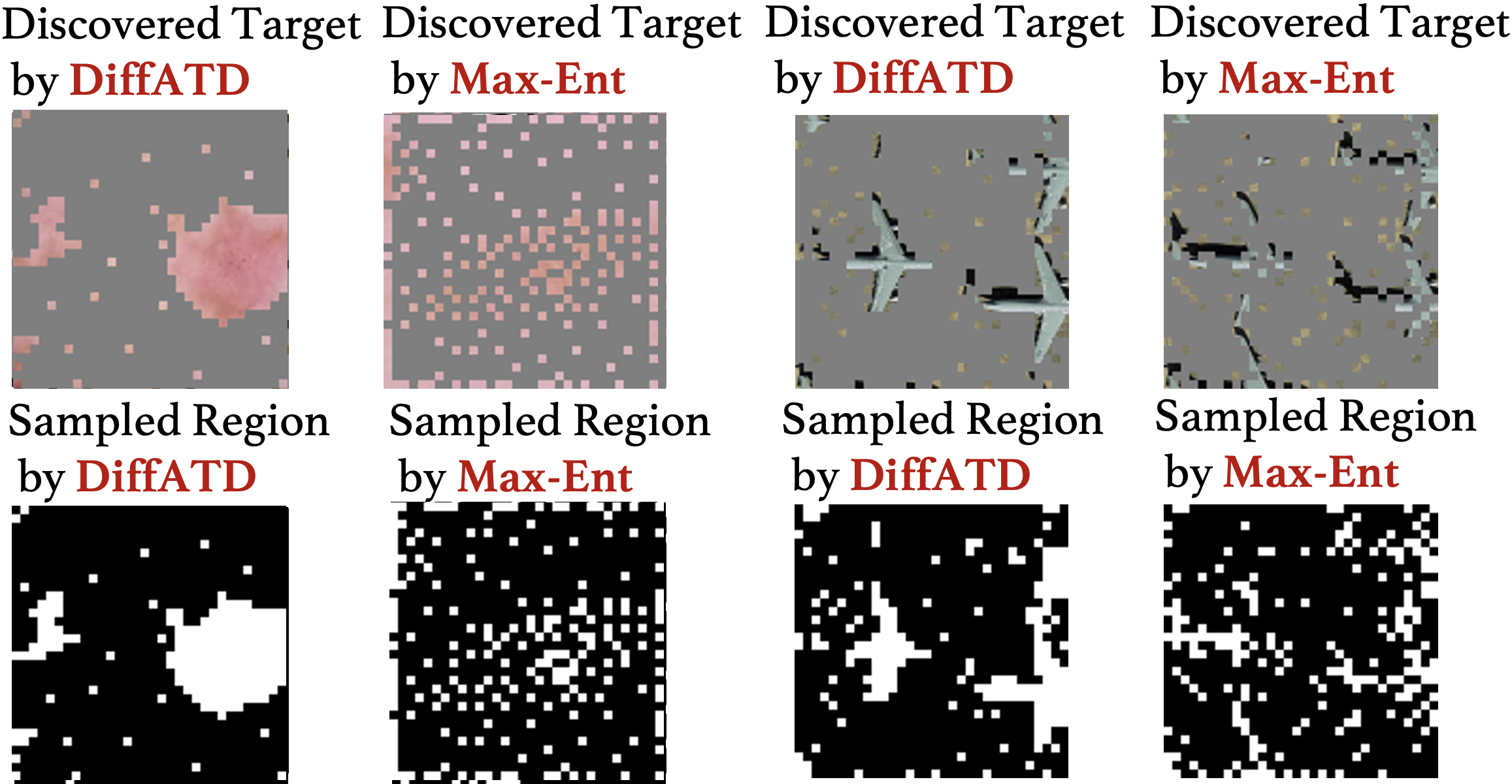}
    \caption{\small{Visualizations of Active Discovery of (left) \emph{Skin disease}, and (right) \emph{Plane}.}}
    \label{fig:explain_s_2}
\end{figure}

\begin{figure}[!h] 
    \centering
    \includegraphics[width=0.88\textwidth]{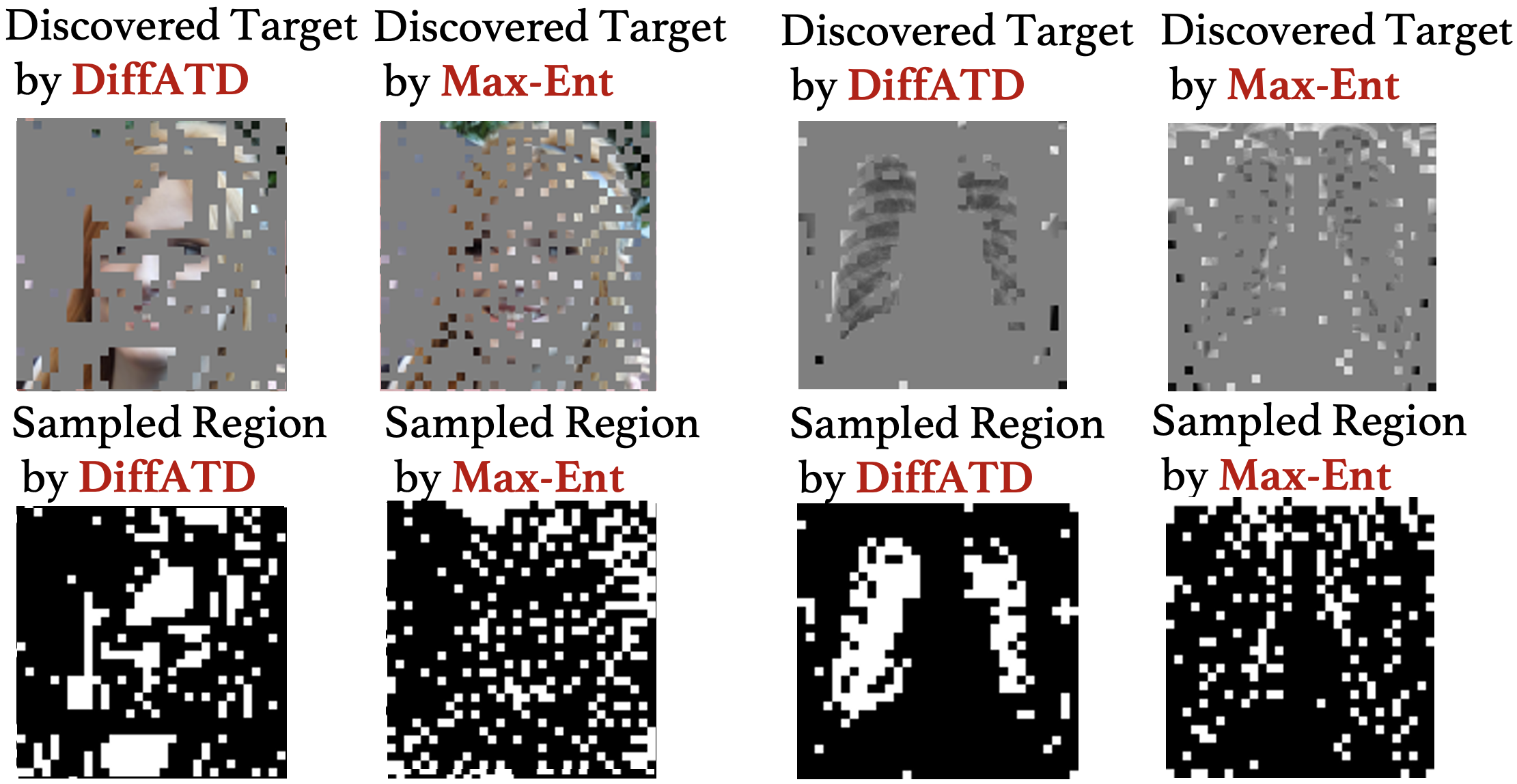}
    \caption{\small{Visualizations of Active Discovery of (left) \emph{human face}, and (right) \emph{lung disease, such as TB}.}}
    \label{fig:explain_s_3}
\end{figure}

\begin{figure}[!h] 
    \centering
    \includegraphics[width=0.88\textwidth]{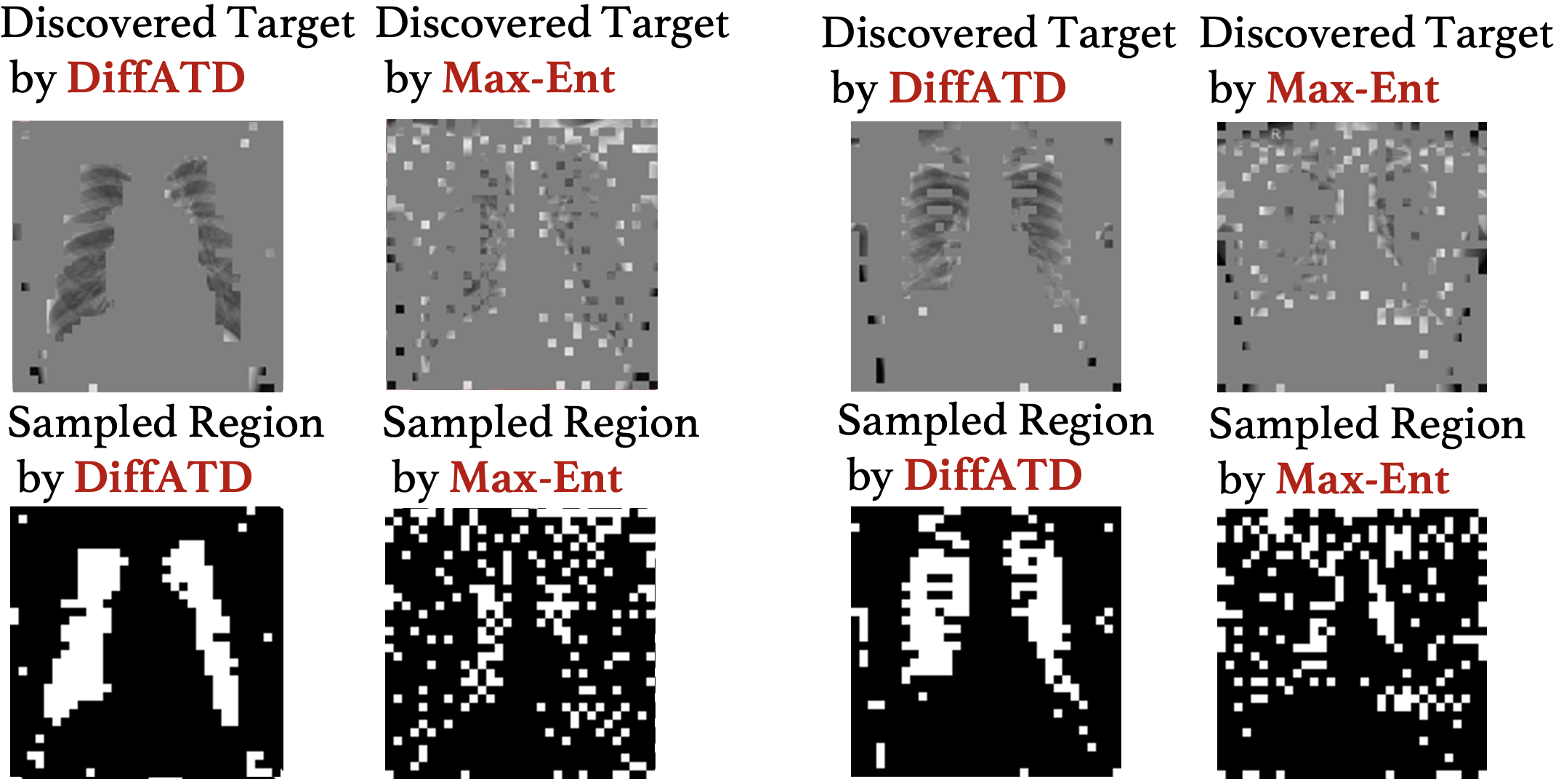}
    \caption{Visualization of Active Discovery of Lung Disease.}
    \label{fig:explain_s_4}
\end{figure}

\begin{figure}[!h] 
    \centering
    \includegraphics[width=0.88\textwidth]{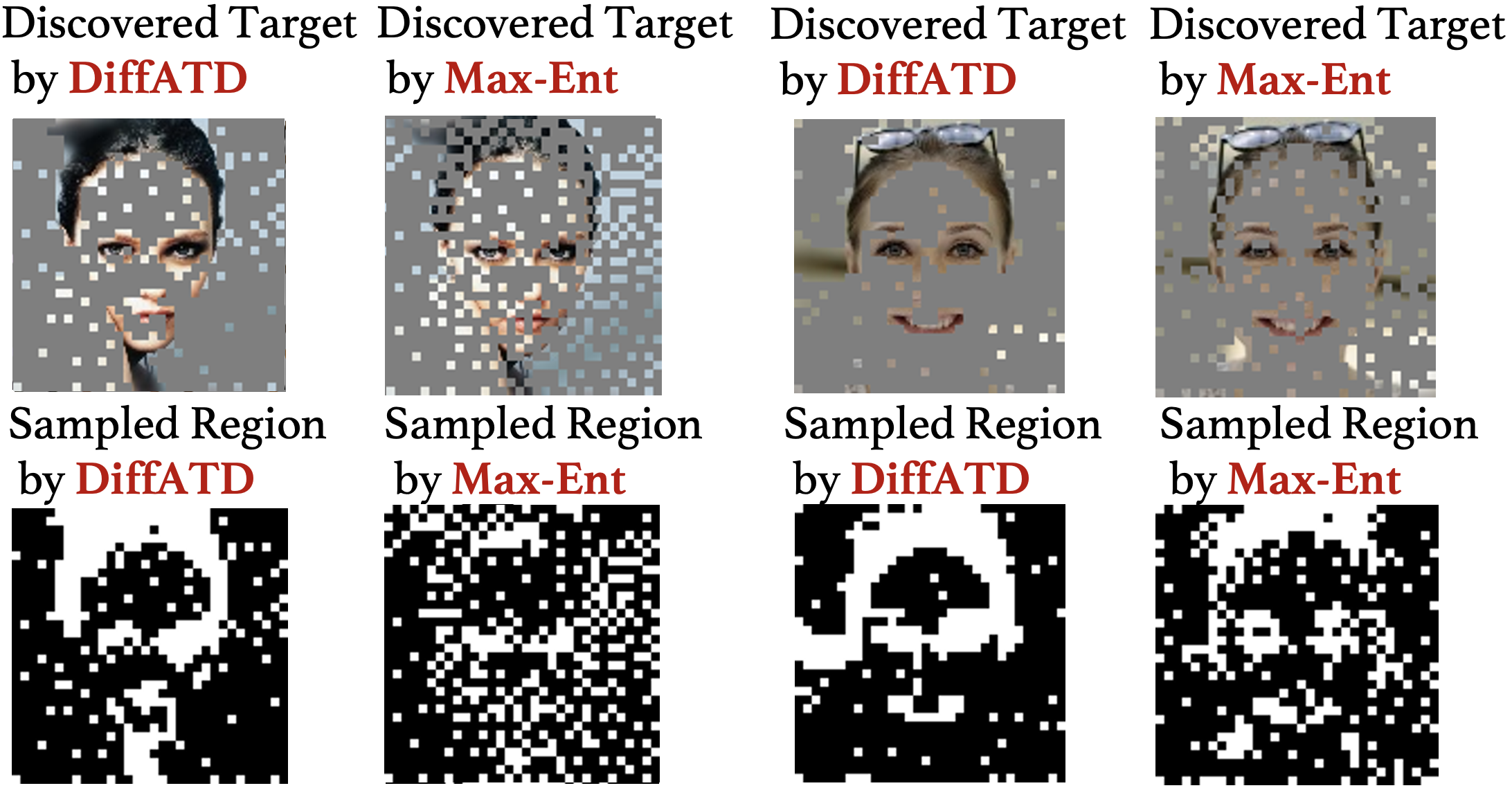}
    \caption{Visualization of Active Discovery of hair, eye, nose, and lip.}
    \label{fig:explain_s_5}
\end{figure}

\begin{figure}[!h] 
    \centering
    \includegraphics[width=0.88\textwidth]{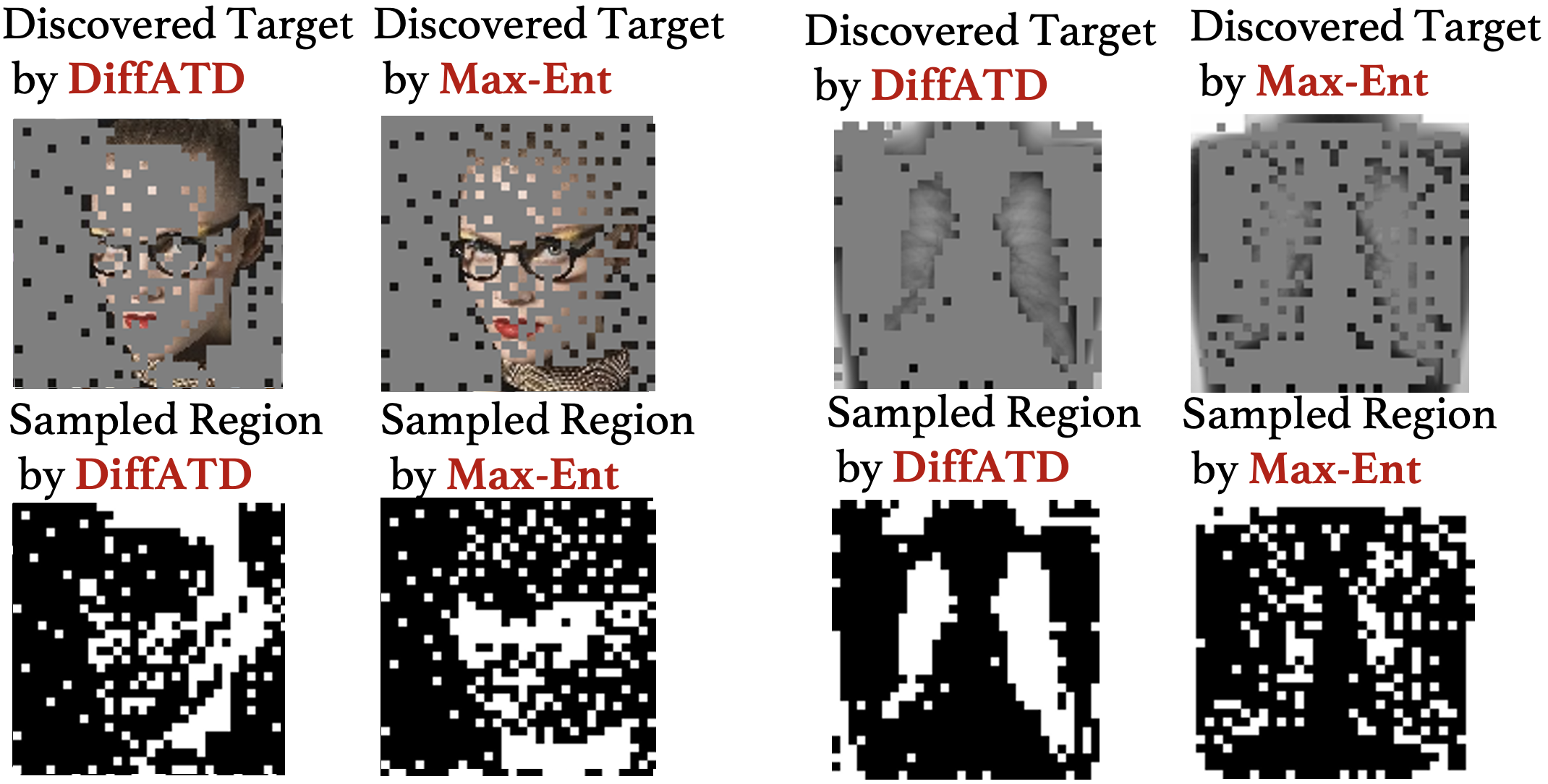}
    \vspace{10pt}
    \caption{Viz. of Active Discovery of Lung Disease (right) and hair, eye, nose, and lip (left).}
    \label{fig:explain_s_6}
    \vspace{-10pt}
\end{figure}

In this section, we provide further comparative visualizations of \emph{DiffATD}'s exploration strategy versus \emph{Max-Ent}, using samples from a variety of datasets, including DOTA, CelebA, Skin, and Lung images. We present the visualization in figures~\ref{fig:explain_s_1},~\ref{fig:explain_s_2},~\ref{fig:explain_s_3},~\ref{fig:explain_s_4},~\ref{fig:explain_s_5},~\ref{fig:explain_s_6}. In each example, \emph{DiffATD} demonstrates a strategic balance between exploration and exploitation, resulting in a significantly higher success rate and more effective target discovery within a predefined budget compared to strategies focused solely on maximizing information gain. These visualizations further reinforce the effectiveness of \emph{DiffATD} in active target discovery within partially observable environments.
\vspace{-15pt}
\vspace{-15pt}

\section{ Details of Training and Inference}
\subsection{Details of Computing Resources, Training, and Inference Hyperparameters}
This section provides the training and inference hyperparameters for each dataset used in our experiments. We use DDIM~\cite{song2020denoising} as the diffusion model across datasets. The diffusion models used in different experiments are based on widely adopted U-Net-style architecture. For the MNIST dataset, we use 32-dimensional diffusion time-step embeddings, with the diffusion model consisting of 2 residual blocks. We select the time-step embedding vector dimension to match the input feature size, ensuring the diffusion model can process it efficiently. The block widths are set to $[32, 64, 128]$, and training involves 30 diffusion steps. DOTA, CelebA, and Skin imaging datasets share the same input feature size of $[128, 128, 3]$ and architecture, featuring 128-dimensional time-step embeddings and a diffusion model with 2 residual blocks of width $[64, 128, 256, 256, 512]$. For these datasets, we perform training with 100 diffusion steps. We use 128-dimensional time-step embeddings for the Bone dataset and a diffusion model with 2 residual blocks (each block width: [64, 128, 256, 256]). We use 100 diffusion steps during training. We set the learning rate and weight decay factor to $1e^{-4}$ for all experimental settings. We set a measurement schedule ($M$) of 100 for a measurement budget ($\mathcal{B}$) of 200, ensuring that $\mathcal{B} \approx \frac{T}{M}$, where $T$ is the total number of reverse diffusion steps, set to approximately 2000 for the DOTA, CelebA, and Skin datasets. Finally, all experiments are implemented in Tensorflow and conducted on NVIDIA A100 40G GPUs. Our training and inference code will be made public.



\subsection{Details of Reward Model $r_{\phi}$}
Our proposed method, DiffATD, utilizes a parameterized reward model, 
$r_{\phi}$, to steer the exploitation process. To this end, we employ a neural network consisting of two fully connected layers, with non-linear ReLU activations as the reward model ($r_{\phi}$). The reward model's goal is to predict a score ranging from 0 to 1, where a higher score indicates a higher likelihood that the measurement location corresponds to the target, based on its semantic features. 
Note that the size of the input semantic feature map for a given measurement location can vary depending on the downstream task. For instance, when working with the MNIST dataset, we use a $1 \times 1$ pixel as the input feature, while for other datasets like CelebA, DOTA, Bone, and Skin imaging, we use an $4 \times 4$ patch as the input feature size.
After each measurement step, we update the model parameters ($\phi$) using the objective function outlined in Equation~\ref{eq:reward}. Additionally, the training dataset is updated with the newly observed data point, refining the model’s predictions over time. Naturally, as the search advances, the reward model refines its predictions, accurately identifying target-rich regions, which makes it progressively more dependable for informed decision-making. 
The reward model architecture is consistent across datasets, including Chest X-ray, Skin, CelebA, and DOTA. It consists of 1 convolutional layer with a $3 \times 3$ kernel, followed by 5 fully connected (FC) layers, each with its own weights and biases. The first FC layer maps an input of size $\frac{(input\>size)^2}{4}$ to an output of size 4 with weights and biases of size $[\frac{(input\>size)^2}{4}, 4]$ and $[4]$ respectively. The second FC layer transforms an input of dimension 4 to an output of size 32 with a 2-dimensional weight of size $[4, 32]$ and a bias of size $[32]$. The third FC layer maps 32 inputs to 16 outputs via a weight matrix of shape $[32, 16]$ and a bias vector of size $[16]$. The pre-final FC layer transforms inputs of size 16 to outputs of size 8  with $[16,8]$ weights, and a bias of shape $[8]$. The final FC layer produces an output of size 2, with weights of size $[8, 2]$ and a bias of size $[2]$, representing the target and non-target scores. The reward model uses the leaky ReLU activation function after each layer. We update the reward model parameters after each measurement step based on the objective in Equation~\ref{eq:reward}. The reward model is trained incrementally for 3 epochs after each measurement step using the gathered supervised dataset resulting from sequential observation, with a learning rate of $0.01$.
\vspace{-3pt}
\section{ Challenges in Active Target Discovery for Rare Categories}
\begin{table}[!h]
    \centering
    \footnotesize
    \caption{Comparison with supervised and fully observable method}
    \begin{tabular}{p{1.5cm}p{1.5cm}p{1.5cm}p{1.5cm}}
        \toprule
        \multicolumn{4}{c}{Active Discovery of Targets on Skin Images} \\
        \midrule
        Method & $\mathcal{B}=150$ & $\mathcal{B}=200$ & $\mathcal{B}=250$ \\
        \midrule
        \cmidrule(r){1-4}  
        \emph{FullSEG} & 0.7304 & 0.6623 & 0.6146  \\ 
        \emph{DiffATD} & \textbf{0.9061} & \textbf{0.8974} & \textbf{0.8752}  \\ 
        \bottomrule
    \end{tabular}
    \label{tab: full_obs1}
    \vspace{-3pt}
\end{table}
To highlight the challenges in Active Target Discovery for rare categories, such as skin disease, we conduct an experiment comparing the performance of \emph{DiffATD} with a fully supervised state-of-the-art semantic segmentation model, SAM, which operates under full observability of the search space (referred to as \emph{FullSEG}). During inference, \emph{FullSEG} selects the top $\mathcal{B}$ most probable target regions for measurement in a single pass. We present the results for Skin Disease as the target, with varying budgets $\mathcal{B}$, in Table~\ref{tab: full_obs1}. A significant performance gap is observed across different measurement budgets. These results underscore the challenges in Active Target Discovery for rare categories, as state-of-the-art segmentation models like SAM struggle to efficiently discover rare targets, like skin disease, further demonstrating the effectiveness of \emph{DiffATD}.

\section{ Scalability of DiffATD on Larger Search Spaces}
To empirically validate DiffATD, we conduct experiments across diverse domains, including remote sensing (DOTA) and medical imaging (Lung Disease dataset), and explore targets of varying complexity, from structured MNIST digits to spatially disjoint human face parts. These cases require strategic exploration, highlighting DiffATD’s adaptability. To assess scalability, we compare DiffATD and the baseline approaches with a larger search space size of 256 $\times$ 256, and present the result in Table~\ref{tab: scale}. Our findings reinforce DiffATD’s effectiveness in complex tasks. 

\begin{table}[H]
    \centering
    \footnotesize
    \caption{Results With Larger Search Space (256 $\times$ 256) using DOTA Dataset}
    \begin{tabular}{p{1.5cm}p{1.5cm}p{1.5cm}p{1.5cm}}
        \toprule
        \multicolumn{4}{c}{Active Discovery of Objects, e.g. Plane, Truck, etc.} \\
        \midrule
        Method & $\mathcal{B}=100$ & $\mathcal{B}=150$ & $\mathcal{B}=200$  \\
        \midrule
        RS & 0.1990 & 0.2487 & 0.2919   \\
        Max-Ent & 0.3909 & 0.4666 & 0.5759  \\
        GA & 0.4780 & 0.5632 & 0.6070  \\
        \hline 
        \textbf{\emph{DiffATD}} & \textbf{0.5251} & \textbf{0.6106} & \textbf{0.7576}   \\ 
        \bottomrule
    \end{tabular}
    \label{tab: scale}
\end{table}

\section{ Additional Results to Access the Effect of $\kappa{(\beta})$}
In the main paper, we analyze how the exploration-exploitation tradeoff function impacts search performance, using both the DOTA and Skin imaging datasets (see Table~\ref{tab: kappa} for the result). Additionally, in this section, we have included sensitivity analysis results on the exploration-exploitation tradeoff function for the CelebA and datasets in the following Table~\ref{tab: kappa}. The observed trends are consistent with those reported in Table 5 of the main paper. These additional findings further strengthen our hypothesis regarding the role of the exploration-exploitation tradeoff function.

\begin{table}[H]
    \centering
    \footnotesize
    \caption{\emph{DiffATD's} Performance Across Varying $\alpha$ with $\mathcal{B}=200$}
    \begin{tabular}{p{1.5cm}p{1.5cm}p{1.5cm}p{1.5cm}}
        \toprule
        \multicolumn{4}{c}{Active Discovery of Handwritten Digits} \\
        \midrule
        Dataset & $\alpha =0.2$ & $\alpha =1.0$ & $\alpha =5.0$  \\
        \midrule
        CelebA & 0.4193 & \textbf{0.4565} & 0.4258   \\
        \hline 
        MNIST & 0.9227 & \textbf{0.9682} & 0.9459   \\ 
        \bottomrule
    \end{tabular}
    \label{tab: kappa}
\end{table}

\section{ Rationale Behind Uniform Measurement Schedule Over the Number of Reverse Diffusion Steps}
We adopt a uniform measurement schedule across all denoising steps, and this design choice is deliberate. A non-uniform schedule — with fewer measurements at higher noise levels and more frequent measurements at lower noise levels — might seem intuitive, but is less effective. By the time the noise level is low, the image structure is largely formed, and overly frequent measurements at that stage do not provide additional value, as they limit the diffusion model’s ability to adapt meaningfully based on the measurements. Instead, a uniform schedule ensures that measurements are distributed evenly, giving the diffusion model adequate time to adapt and refine its predictions based on each measurement. This balance ultimately leads to more stable and effective reconstructions across diverse settings. On the other hand, a non-uniform schedule — with more frequent measurements at higher noise levels and fewer at lower noise levels — can be counterproductive. At higher noise levels, the reconstruction is predominantly governed by random noise, making frequent measurements less informative and potentially wasteful.

\section{ Performance of DiffATD Under Noisy Observations}
\begin{figure}[!h] 
    \centering
    \includegraphics[width=0.50\textwidth]{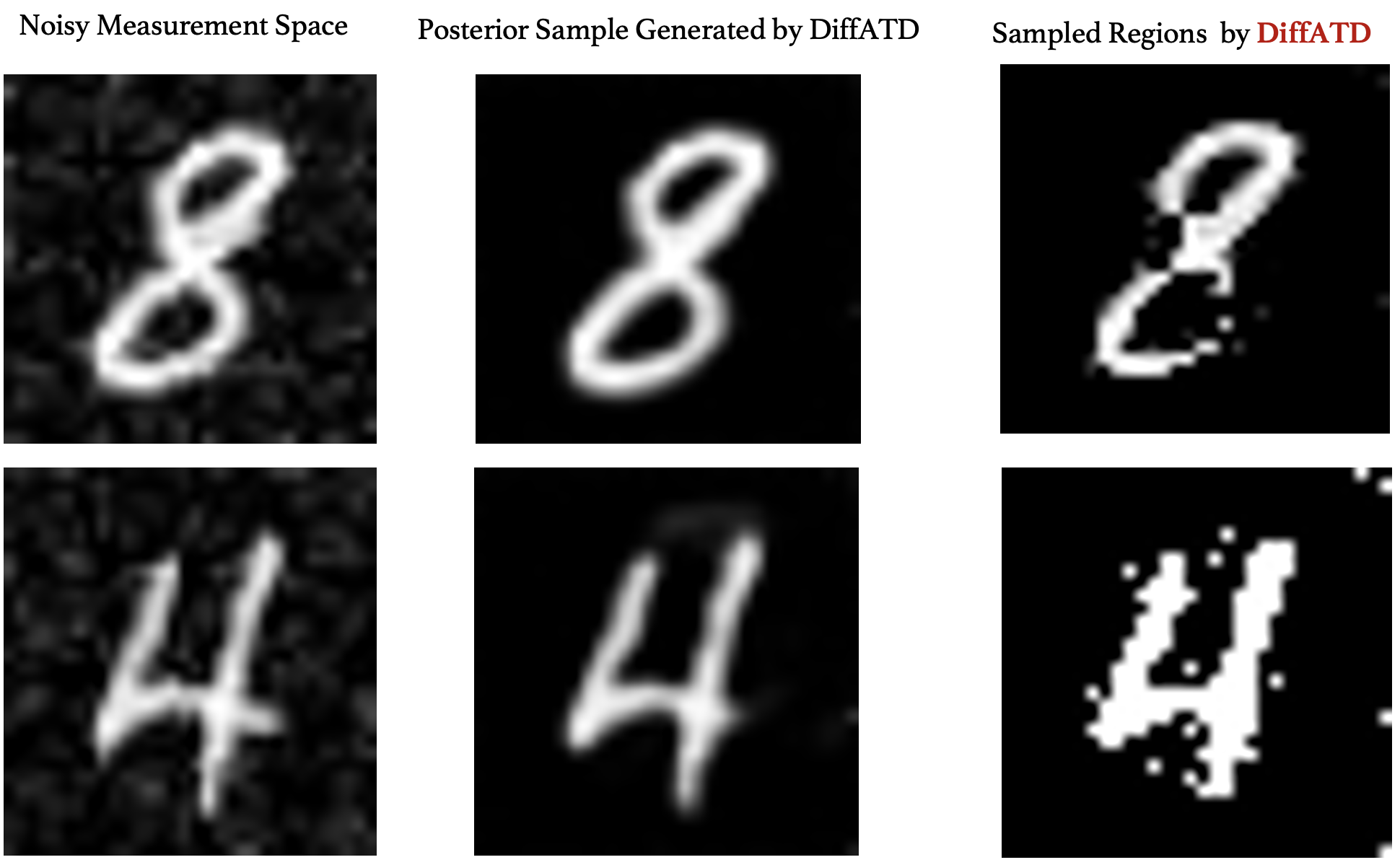}
    \vspace{10pt}
    \caption{Visualizations of Reconstructed Search Space from Noisy Observations.}
    \label{fig:noise_recon}
\end{figure}

We evaluate the robustness of DiffATD in the presence of noisy observations by introducing varying levels of noise into the observation space. As shown in Tables~\ref{tab:noise_mnist} and \ref{tab:noise_dota}, DiffATD consistently maintains strong performance across noise levels across different settings, demonstrating its resilience and reliability under imperfect data conditions. To further understand DiffATD's robustness, we visualize posterior reconstructions of the search space from sparse, partially observable, and noisy inputs (Figure~\ref{fig:noise_recon}). Despite the noise, DiffATD effectively reconstructs a clean representation of the underlying space, explaining its consistent performance across varying noise levels in the active target discovery task.

\begin{table}[H]
    \centering
    \footnotesize
    \caption{DiffATD's Performance with MNIST dataset Under Noisy Observation.}
    \begin{tabular}{p{2.2cm}p{1.5cm}p{1.5cm}p{1.5cm}}
        \toprule
        \multicolumn{4}{c}{Active Discovery of Handwritten Digits} \\
        \midrule
        Noise Level & $\mathcal{B}=100$ & $\mathcal{B}=150$ & $\mathcal{B}=200$  \\
        \midrule
        $\mu$=0 and $\sigma$ = 20 & 0.7318 & 0.8420 & 0.9674   \\
        $\mu$=0 and $\sigma$ = 30 & 0.7307 & 0.8401 & 0.9641  \\
        $\mu$=10 and $\sigma$ = 30 & 0.7298 & 0.8393 & 0.9624  \\
        \hline 
        No Noise & \textbf{0.7324} & \textbf{0.8447} & \textbf{0.9682}   \\ 
        \bottomrule
    \end{tabular}
    \label{tab:noise_mnist}
\end{table}

\begin{table}[H]
    \centering
    \footnotesize
    \caption{DiffATD's Performance with DOTA dataset Under Noisy Observation.}
    \begin{tabular}{p{3.2cm}p{1.5cm}p{1.5cm}p{1.5cm}}
        \toprule
        \multicolumn{4}{c}{Active Discovery of objects, e.g, plane, truck, etc.} \\
        \midrule
        White Noise Level & $\mathcal{B}=200$ & $\mathcal{B}=250$ & $\mathcal{B}=300$  \\
        \midrule
        $\mu$=0 and $\sigma$ = 20 & 0.5410 & 0.6150 & 0.7297   \\
        $\mu$=0 and $\sigma$ = 30 & 0.5139 & 0.6335 & 0.7384  \\
        $\mu$=10 and $\sigma$ = 30 & 0.5292 & 0.6319 & 0.7269  \\
        \hline 
        No Noise & \textbf{0.5422} & \textbf{0.6379} & \textbf{0.7309}   \\ 
        \bottomrule
    \end{tabular}
    \label{tab:noise_dota}
\end{table}

\section{ Species Distribution Modelling as Active Target Discovery Problem}\label{app:sde}
We constructed our species distribution experiment using observation data of lady beetle species from iNaturalist. Center points were randomly sampled within North America (latitude 25.6°N to 55.0°N, longitude 123.1°W to 75.0°W). Around each center, we defined a square region approximately 480 km × 480 km in size (roughly 5 degrees in both latitude and longitude).  Each retained region was discretized into a 64×64 grid, where the value of each cell represents the number of observed lady beetles. To simulate the querying process, each 2×2 block of grid cells was treated as a query. We evaluated our method on real Coccinella Septempunctata observation data without subsampling. Our goal is to find as many Coccinella Septempunctata as possible within a region.

\section{ Statistical Significance Results of DiffATD}
In order to strengthen our claim on DiffATD's superiority over the baseline methods, we have included the statistical significance results with the DOTA and CelebA datasets, and present the results in Tables~\ref{tab: sg_dota},~\ref{tab: sg_celeba}. These results are based on 5 independent trials and further strengthen our empirical findings, reinforcing the effectiveness of DiffATD across diverse domains. 
\begin{table}[H]
    \centering
    \footnotesize
    \caption{Statistical Significance Results with DOTA Dataset}
    \begin{tabular}{p{1.5cm}p{2.5cm}p{2.5cm}p{2.5cm}}
        \toprule
        \multicolumn{4}{c}{Active Discovery of Objects like Plane, Truck, etc.} \\
        \midrule
        Method & $\mathcal{B}=100$ & $\mathcal{B}=150$ & $\mathcal{B}=200$  \\
        \midrule
        RS & 0.1990 $\pm$ 0.0046 & 0.2487 $\pm$ 0.0032 & 0.2919 $\pm$ 0.0036   \\
        Max-Ent & 0.4625 $\pm$ 0.0150 & 0.5524 $\pm$ 0.0131 & 0.6091 $\pm$ 0.0188  \\
        GA & 0.4586 $\pm$ 0.0167 & 0.5961 $\pm$ 0.0119 & 0.6550 $\pm$ 0.0158  \\
        \hline 
        \textbf{\emph{DiffATD}} & \textbf{0.5422 $\pm$ 0.0141} & \textbf{0.6379 $\pm$ 0.0115} & \textbf{0.7309 $\pm$ 0.0107}   \\ 
        \bottomrule
    \end{tabular}
    \label{tab: sg_dota}
\end{table}

\begin{table}[H]
    \centering
    \footnotesize
    \caption{Statistical Significance Results with CelebA Dataset}
    \begin{tabular}{p{1.5cm}p{2.5cm}p{2.5cm}p{2.5cm}}
        \toprule
        \multicolumn{4}{c}{Active Discovery of Different Parts of Human Faces.} \\
        \midrule
        Method & $\mathcal{B}=200$ & $\mathcal{B}=250$ & $\mathcal{B}=300$  \\
        \midrule
        RS & 0.1938 $\pm$ 0.0017 & 0.2441 $\pm$ 0.0021 & 0.2953 $\pm$ 0.0002   \\
        Max-Ent & 0.2399 $\pm$ 0.0044 & 0.3510 $\pm$ 0.0060 & 0.4498 $\pm$ 0.0118  \\
        GA & 0.2839 $\pm$ 0.0106 & 0.3516 $\pm$ 0.0110 & 0.4294 $\pm$ 0.0121  \\
        \hline 
        \textbf{\emph{DiffATD}} & \textbf{0.4565 $\pm$ 0.0089} & \textbf{0.5646 $\pm$ 0.0086} & \textbf{0.6414 $\pm$ 0.0081}   \\ 
        \bottomrule
    \end{tabular}
    \label{tab: sg_celeba}
\end{table}

\section{ Comparison With Multi-Armed Bandit Based Method}
We compare the performance with MAB-based methods across varying measurement budgets, with results summarized in the Table below~\ref{tab: MAB},~\ref{tab: MAB_dota}. These analyses are conducted in the context of active target discovery on the MNIST digits and DOTA overhead images, respectively. \textbf{These MAB-based methods struggle in the active target discovery setting due to their lack of prior knowledge and structured guidance compared to the Diffusion model, resulting in significantly weaker performance than DiffATD.}

\begin{table}[H]
    \centering
    \caption{\emph{SR} comparison with MNIST Dataset}
    \begin{tabular}{p{1.5cm}p{1.5cm}p{1.5cm}p{1.5cm}}
        \toprule
        \multicolumn{4}{c}{Active Discovery of Handwritten Digits} \\
        \midrule
        Method & $\mathcal{B}=100$ & $\mathcal{B}=150$ & $\mathcal{B}=200$  \\
        \midrule
        UCB & 0.0026 & 0.0194 & 0.0923   \\
        $\epsilon$-greedy & 0.0151 & 0.0394 & 0.0943  \\
        \hline 
        \textbf{\emph{DiffATD}} & \textbf{0.7324} & \textbf{0.8447} & \textbf{0.9682}   \\ 
        \bottomrule
    \end{tabular}
    \label{tab: MAB}
\end{table}
\begin{table}[H]
    \centering
    \caption{\emph{SR} comparison with DOTA Dataset}
    \begin{tabular}{p{1.5cm}p{1.5cm}p{1.5cm}p{1.5cm}}
        \toprule
        \multicolumn{4}{c}{Active Discovery of Objects in Overhead Images} \\
        \midrule
        Method & $\mathcal{B}=200$ & $\mathcal{B}=250$ & $\mathcal{B}=300$  \\
        \midrule
        UCB & 0.1132 & 0.2487 & 0.2146   \\
        \hline 
        \textbf{\emph{DiffATD}} & \textbf{0.5422} & \textbf{0.6379} & \textbf{0.7309}   \\ 
        \bottomrule
    \end{tabular}
    \label{tab: MAB_dota}
\end{table}

\section{ Segmentation and Active Target Discovery are Fundamentally Different Tasks}
\begin{figure}[!h] 
    \centering
    \includegraphics[width=0.78\textwidth]{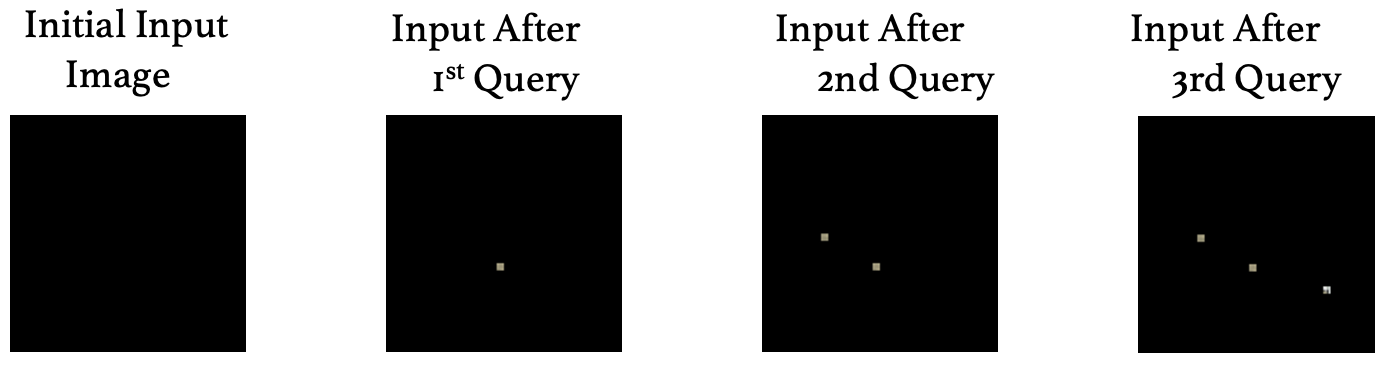}
    \vspace{10pt}
    \caption{Visualization of Search Space at initial Active Target Discovery Phase.}
    \label{fig:no_seg}
\end{figure}

Segmentation and Active Target Discovery (ATD) serve fundamentally different purposes. Segmentation techniques operate under the assumption that the full or partial image is already available, focusing on labeling observed regions. In contrast, ATD is centered around reasoning about unobserved regions using only sparse, partial observations. This distinction is crucial—when only a few pixels are known, as in our setting (An illustrative example of this scenario is shown in Figure~\ref{fig:no_seg}), segmentation models offer little value, since the primary challenge lies not in interpreting the visible content, but in strategically exploring and inferring the hidden parts of the search space to maximize the target object discovery.

\section{ More Details on Computational Cost across Search Space}
We have conducted a detailed evaluation of sampling time and computational requirements of \emph{DiffATD} across various search space sizes. We present the results in~\ref{tab: COMP}. Our results show that \emph{DiffATD} remains efficient even as the search space scales, with sampling time per observation ranging from 0.41 to 3.26 seconds, which is well within practical limits for most downstream applications. This further reinforces DiffATD's scalability and real-world applicability.

\begin{table}[H]
    \centering
    \footnotesize
    \caption{Details of Computation and Sampling Cost Across Varying Search Space Sizes}
    \begin{tabular}{p{2.5cm}p{3.5cm}p{4.5cm}}
        \toprule
        \multicolumn{3}{c}{Active Discovery of Handwritten Digits} \\
        \midrule
        Search Space & Computation Cost & Sampling Time (Seconds)  \\
        \midrule
        28 $\times$ 28 & 726.9 MB & 0.41    \\
        128 $\times$ 128 & 1.35 GB & 1.48  \\
        \hline 
        256 $\times$ 256 & 3.02 GB & 3.26   \\ 
        \bottomrule
    \end{tabular}
    \label{tab: COMP}
\end{table}

\section{ More Visualizations on \emph{DiffATD}'s Exploration vs Exploitation strategy at Different Stage}\label{sec:EXP_B}
\begin{figure}[!h] 
    \centering
    \includegraphics[width=0.98\textwidth]{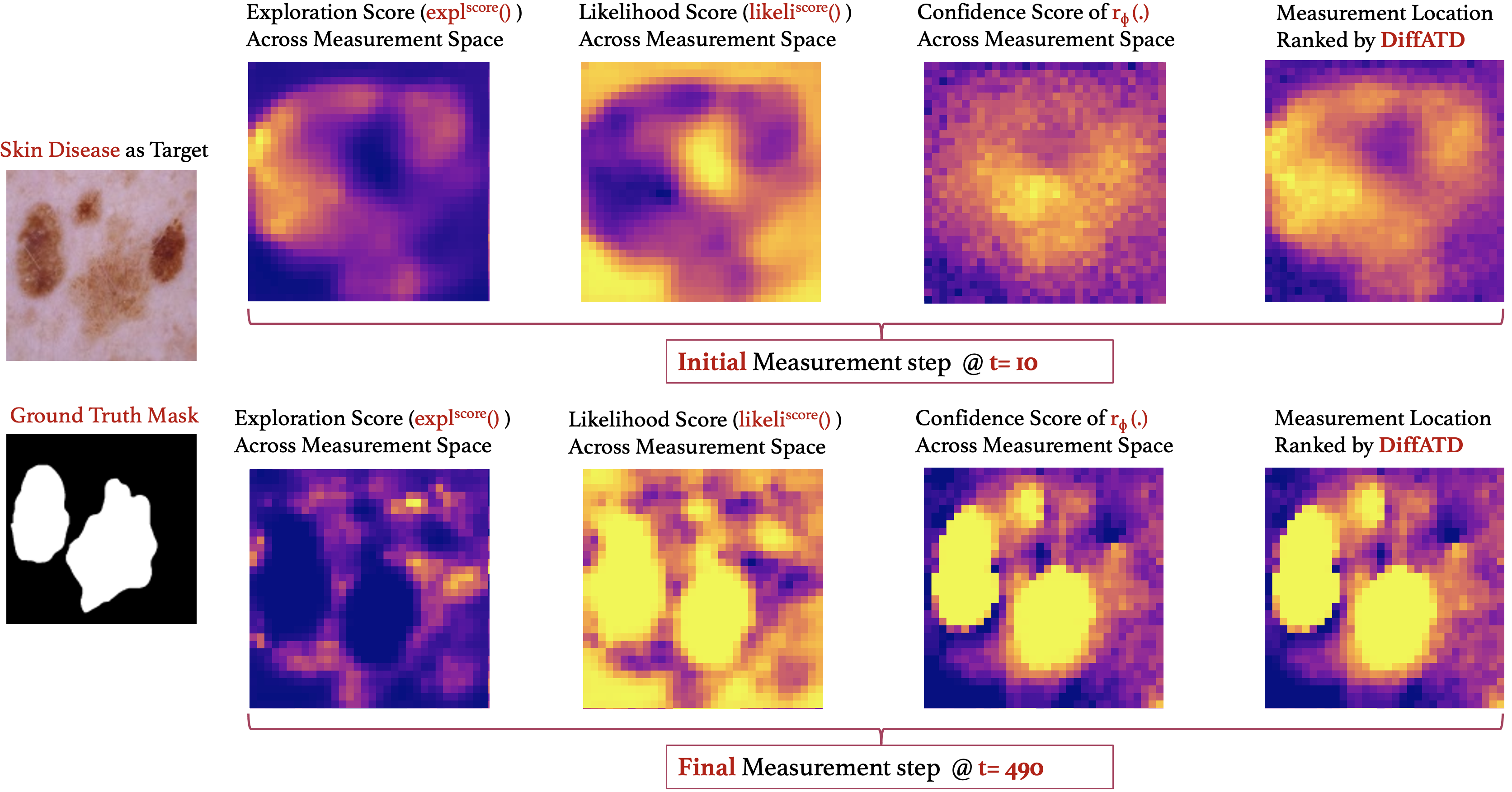}
    \vspace{7pt}
    \caption{\small{Explanation of \emph{DiffATD}. Brighter indicates a higher value and higher rank.}}
    \label{fig:explain_1}
    \vspace{2pt}
\end{figure}

\begin{figure}[!h] 
    \centering
    \includegraphics[width=0.98\textwidth]{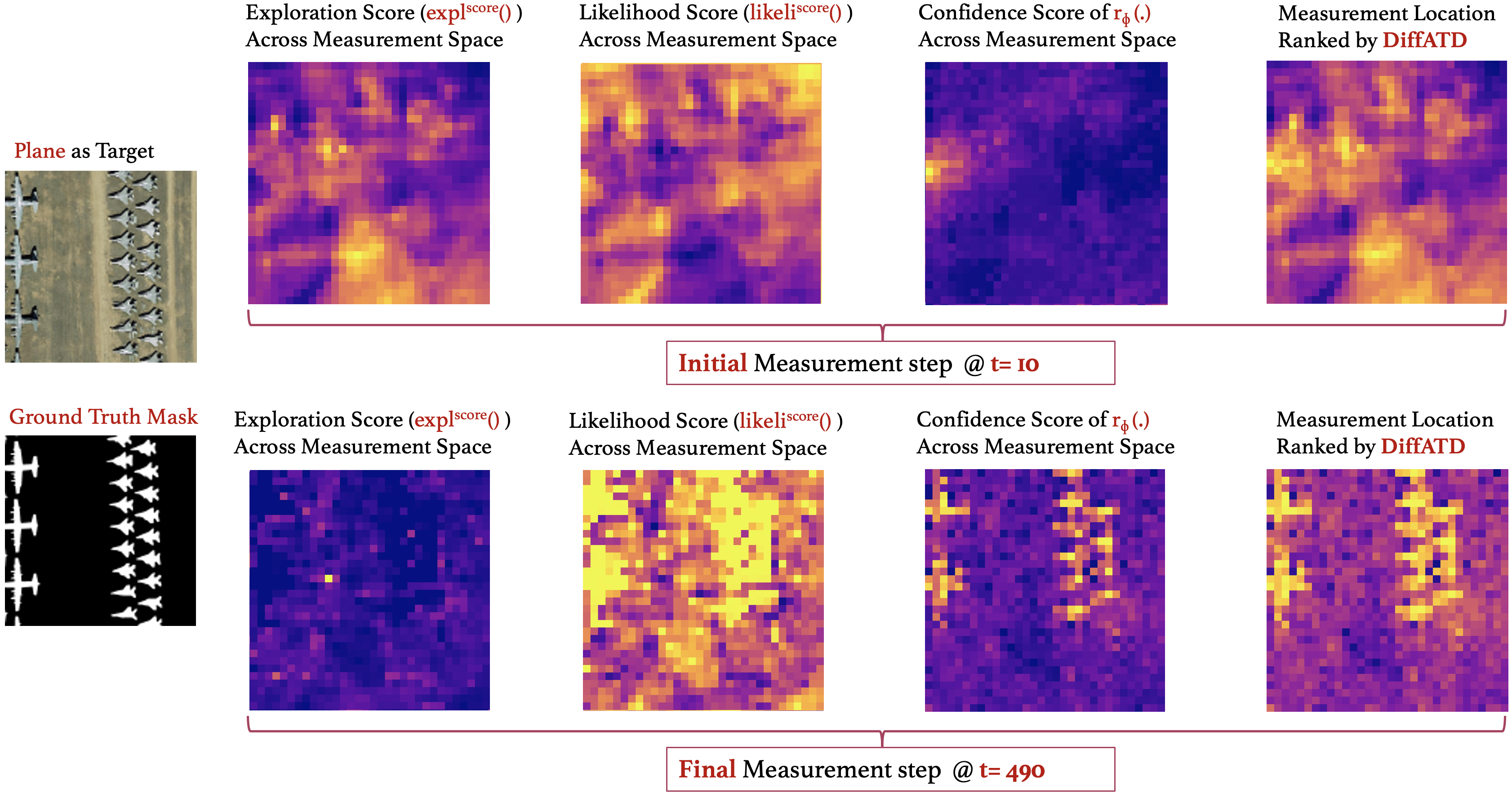}
    \caption{\small{Explanation of \emph{DiffATD}. Brighter indicates a higher value and higher rank.}}
    \label{fig:explain_2}
\end{figure}

\begin{figure}[!h] 
    \centering
    \includegraphics[width=0.98\textwidth]{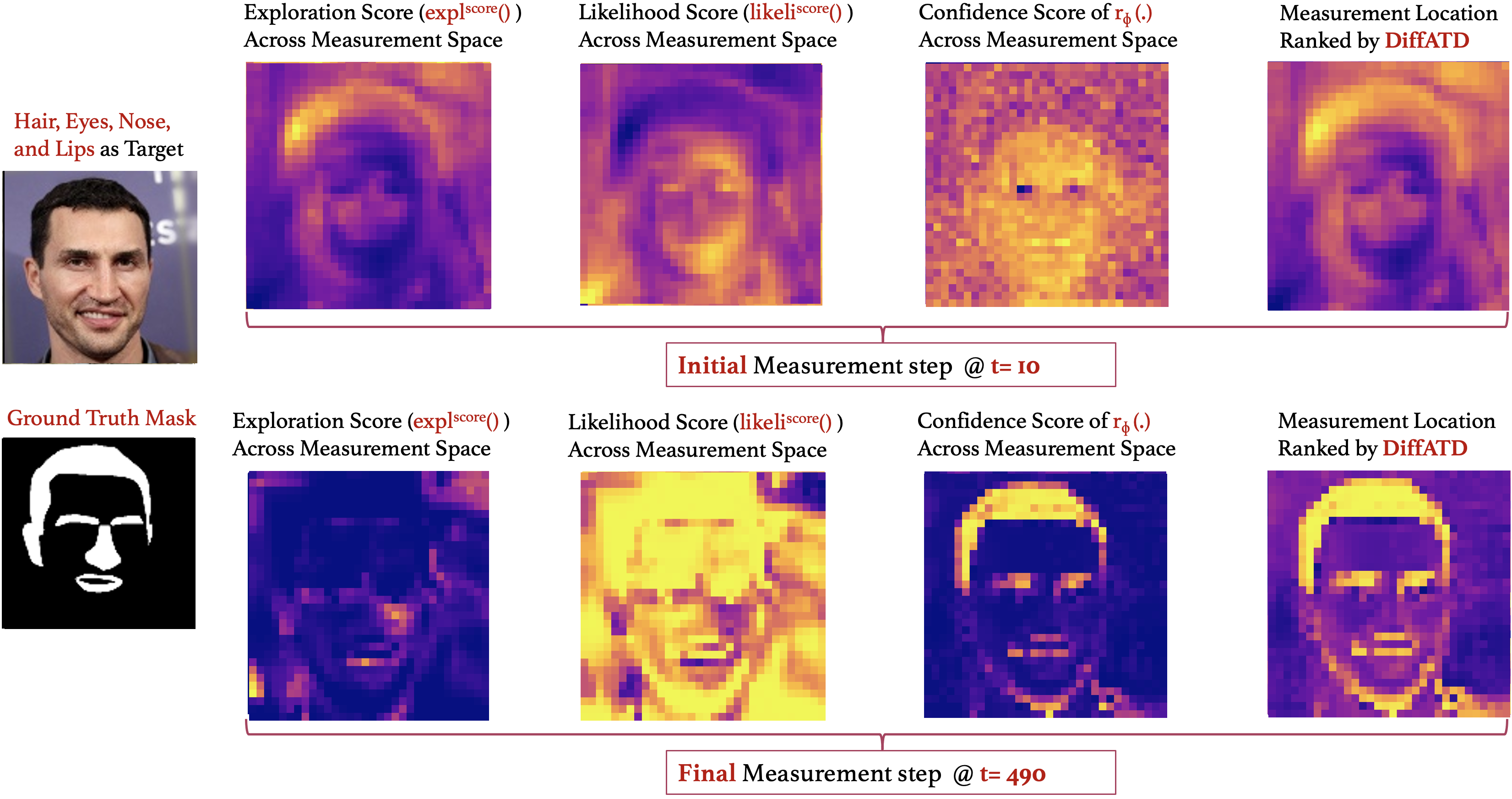}
    \caption{\small{Explanation of \emph{DiffATD}. Brighter indicates a higher value, and higher rank.}}
    \label{fig:explain_3}
\end{figure}

\begin{figure}[!h] 
    \centering
    \includegraphics[width=0.98\textwidth]{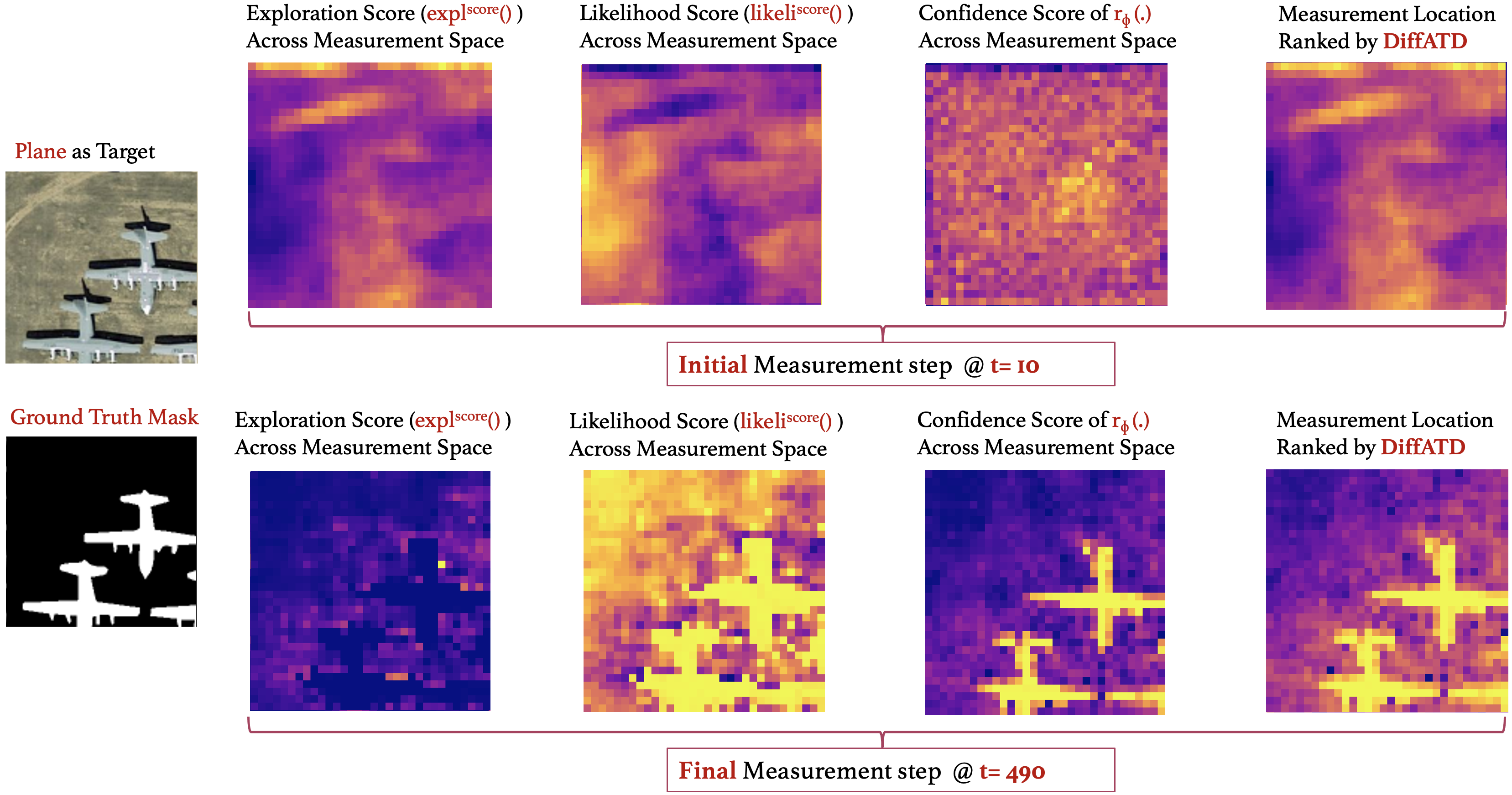}
    \caption{\small{Explanation of \emph{DiffATD}. Brighter indicates a higher value, and higher rank.}}
    \label{fig:explain_5}
\end{figure}

\begin{figure}[!h] 
    \centering
    \includegraphics[width=0.98\textwidth]{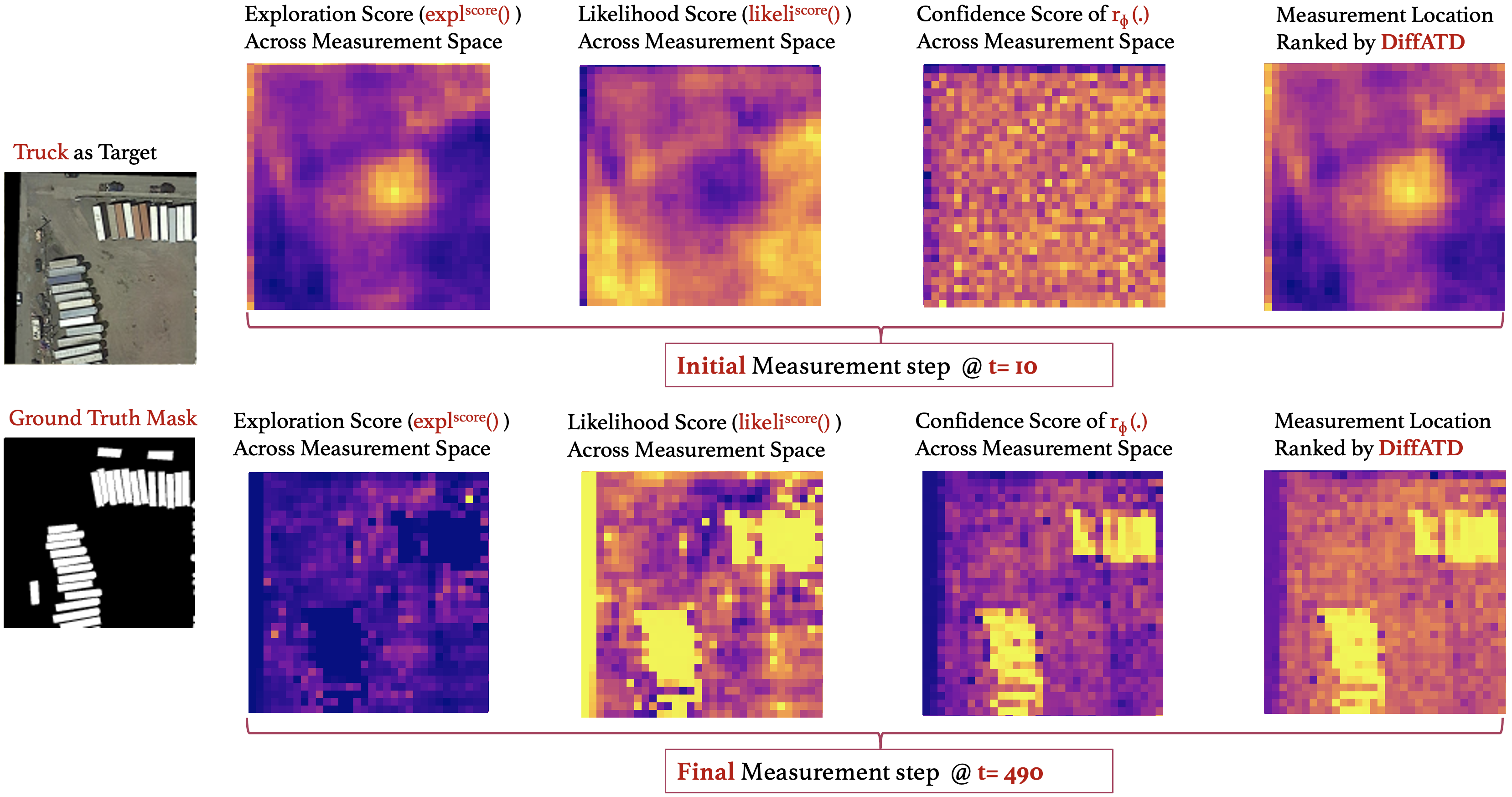}
    \caption{\small{Explanation of \emph{DiffATD}. Brighter indicates a higher value, and higher rank.}}
    \label{fig:explain_6}
\end{figure}

In this section, we provide additional visualizations that illustrate how \emph{DiffATD} balances exploration and exploitation at various stages of the active discovery process. These visualizations are shown in figures~\ref{fig:explain_1},~\ref{fig:explain_2},~\ref{fig:explain_3},~\ref{fig:explain_5},~\ref{fig:explain_6}. In each example, we show the exploration score ($\mathrm{expl}^{\mathrm{score}} ()$), likelihood score ($\mathrm{likeli}^{\mathrm{score}} ()$), and the confidence score of the reward model ($r_{\phi}$) across measurement space at two distinct stages of the active discovery process. The top row represents the initial phase (measurement step $10$), while the bottom row corresponds to a near-final stage (measurement step $490$). We observe a similar trend across all examples, i.e., \emph{DiffATD} prioritizes the exploration score when selecting measurement locations during the early phase of active discovery. Thus, \emph{DiffATD} aims to maximize information gain during the initial search stage. As the search approaches its final stages, the rankings of the measurement locations shift to being primarily driven by the exploitation score, as defined in Equation~\ref{eq:exploit-score}. As the search progresses, the confidence score of $r_{\phi}$ becomes more accurate, making the $\mathrm{expl}^{\mathrm{score}}$() more reliable. This explains \emph{DiffATD}'s growing reliance on the exploitation score in the later phases of the process. These additional visualizations illustrate \emph{DiffATD}'s exploration strategy, particularly how it dynamically balances exploration and exploitation. These visualizations also underscore the effectiveness of \emph{DiffATD} in addressing active target discovery within partially observable environments.
\vspace{26pt}

\vspace{-6pt}
\newpage
\section{ Importance of Utilizing a Strong Prior in Tackling Active Target Discovery}
The purpose of the prior belief model (e.g., a pretrained Diffusion model) in DiffATD is to construct a dynamically updated, uncertainty-aware belief distribution over unobserved regions—critical for guiding the exploration-exploitation tradeoff in target discovery. Thus, it is important to utilize a belief model that is capable of understanding the hidden structure of the underlying search space. In order to validate this, we conduct an experiment with a different choice of the prior belief model that is trained on a different domain (i.e., ImageNet), and compare its performance with DiffATD where the prior belief model is trained on the same domain (i.e., DOTA). We present the result in the following Table~\ref{tab: prior1}. Our empirical observation further reinforces the fact that leveraging a strong prior belief model is crucial for efficiently tackling active target discovery.
\vspace{-8pt}
\begin{table}[H]
    \vspace{-3pt}
    \centering
    \caption{\emph{SR} comparison with DOTA Dataset}
    \begin{tabular}{p{5.5cm}p{1.5cm}p{1.5cm}}
        \toprule
        \multicolumn{3}{c}{Active Discovery of Objects from Overhead Imagery} \\
        \midrule
        Method & $\mathcal{B}=250$ & $\mathcal{B}=300$  \\
        \midrule
        DiffATD with a weak prior Model  & 0.5143 & 0.6391   \\
        \hline 
        \textbf{\emph{DiffATD}} & \textbf{0.6379} & \textbf{0.7309}   \\ 
        \bottomrule
    \end{tabular}
    \label{tab: prior1}
    \vspace{-3pt}
\end{table}
\vspace{-9pt}
\section{ Notation Table}
To prevent ambiguity, we include below a compact notation table summarizing all symbols and variables used in describing the DiffATD methodology.
\begin{table}[h!]
\centering
\caption{Notation summary for the DiffATD framework.}
\vspace{0.5em}
\begin{tabular}{ll}
\hline
\textbf{Notation} & \textbf{Description} \\
\hline
$\hat{x}_t$ & Predicted or estimated search space $x$ at the $t$-th observation step. \\
$\tilde{x}_t$ & Set of previously observed partial glimpses of the search space up to step $t$. \\
$\hat{x}^{(i)}_t$ & $i$-th element in the batch of particles (superscript $(i)$ denotes particle index). \\
$q_t$ & Measurement location at the $t$-th time step. \\
$Q_t$ & Set of measurement locations up to time step $t$. \\
$\tau$ & Time step index in the reverse diffusion process. \\
\hline
\end{tabular}
\label{tab:notations}
\end{table}

\section{Limitations and Future Work}
While DiffATD demonstrates strong efficiency in addressing active target discovery tasks across diverse settings, it currently depends on the availability of a robust domain-specific prior model to perform effectively. This requirement can be restrictive in real-world applications such as rare species identification or rare disease discovery, where data scarcity makes it difficult to establish a reliable prior. Consequently, the method’s success is inherently linked to the quality and expressiveness of the prior model. In future work, we aim to extend DiffATD toward settings with weak or no prior knowledge, enabling active target discovery under an uninformative prior.
\vspace{-4pt}


\end{document}